\documentclass[10pt,journal,compsoc]{IEEEtran}

\ifCLASSOPTIONcompsoc
  \usepackage[nocompress]{cite}
\else
  \usepackage{cite}
\fi

\usepackage{booktabs} 
\usepackage[ruled]{algorithm2e} 

\usepackage{epsfig}
\usepackage{graphicx}
\usepackage{amsmath}
\usepackage{amssymb}
\usepackage{subcaption}
\usepackage{ifthen}

\usepackage[colorlinks,urlcolor=blue]{hyperref}
\usepackage{array}
\usepackage{color}
\usepackage{amsmath}
\usepackage{float}
\usepackage{subcaption}
\usepackage{soul}

\soulregister\cite7 
\soulregister\citep7 
\soulregister\citet7 
\soulregister\ref7 
\soulregister\pageref7 

\usepackage{cuted}
\usepackage{capt-of}

\usepackage{wrapfig}

\begin{document}

\title{Exemplar-Based 3D Portrait Stylization}

\author{Fangzhou~Han$^1$,
        Shuquan~Ye$^1$,
        Mingming~He,
        Menglei~Chai,
        and~Jing~Liao$^*$
\IEEEcompsocitemizethanks{
\IEEEcompsocthanksitem $^1$: equal contribution. $^*$: corresponding author.
\IEEEcompsocthanksitem F. Han, S. Ye and J. Liao are with Department of Computer Science, City University of Hong Kong. E-mail: han.fangzhou@my.cityu.edu.hk, shuquanye2-c@my.cityu.edu.hk, jingliao@cityu.edu.hk.
\IEEEcompsocthanksitem M. He is with the Institute for Creative Technologies, University of Southern California. E-mail: hmm.lillian@gmail.com.
\IEEEcompsocthanksitem M. Chai is with the Creative Vision team, Snap Inc.. E-mail: cmlatsim@gmail.com.
\IEEEcompsocthanksitem This work has been submitted to the IEEE for possible publication. Copyright may be transferred without notice, after which this version may no longer be accessible.}
}

\IEEEtitleabstractindextext{%
\begin{abstract}
Exemplar-based portrait stylization is widely attractive and highly desired. Despite recent successes, it remains challenging, especially when considering both texture and geometric styles. In this paper, we present the first framework for one-shot 3D portrait style transfer, which can generate 3D face models with both the geometry exaggerated and the texture stylized while preserving the identity from the original content. It requires only one arbitrary style image instead of a large set of training examples for a particular style, provides geometry and texture outputs that are fully parameterized and disentangled, and enables further graphics applications with the 3D representations. The framework consists of two stages. In the first geometric style transfer stage, we use facial landmark translation to capture the coarse geometry style and guide the deformation of the dense 3D face geometry. In the second texture style transfer stage, we focus on performing style transfer on the canonical texture by adopting a differentiable renderer to optimize the texture in a multi-view framework. Experiments show that our method achieves robustly good results on different artistic styles and outperforms existing methods. We also demonstrate the advantages of our method via various 2D and 3D graphics applications. Project page is: \url{https://halfjoe.github.io/projs/3DPS/index.html}.
\end{abstract}

}

\maketitle

\IEEEpeerreviewmaketitle

\IEEEraisesectionheading{\section{Introduction}\label{sec:introduction}}

\begin{figure*}[ht]
    \centering
    \setlength{\tabcolsep}{0mm}{
\begin{tabular}{ccccccccc}
  \includegraphics[width=0.11\linewidth]{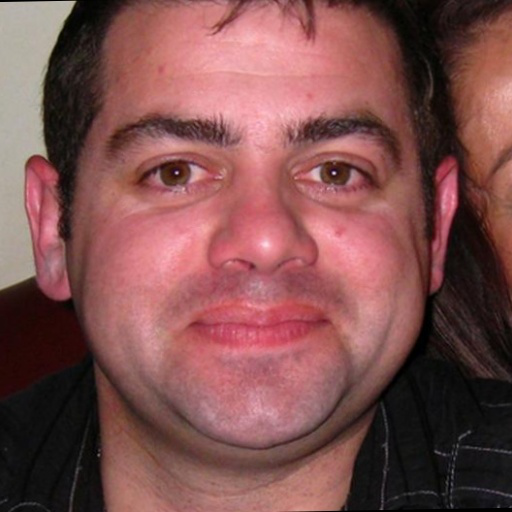} &
  \includegraphics[width=0.11\linewidth]{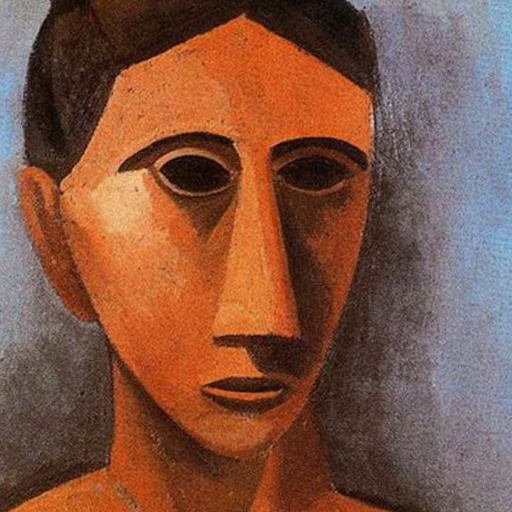} &
  \includegraphics[width=0.11\linewidth]{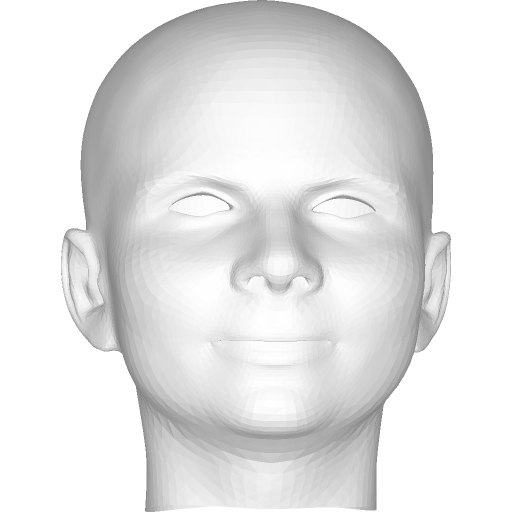} &
  \includegraphics[width=0.11\linewidth]{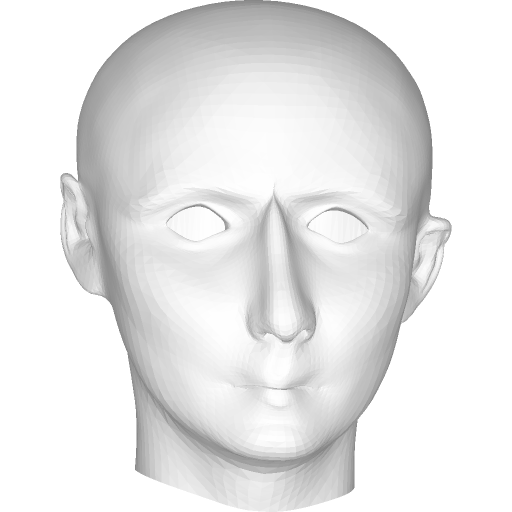}&
  \includegraphics[width=0.11\linewidth]{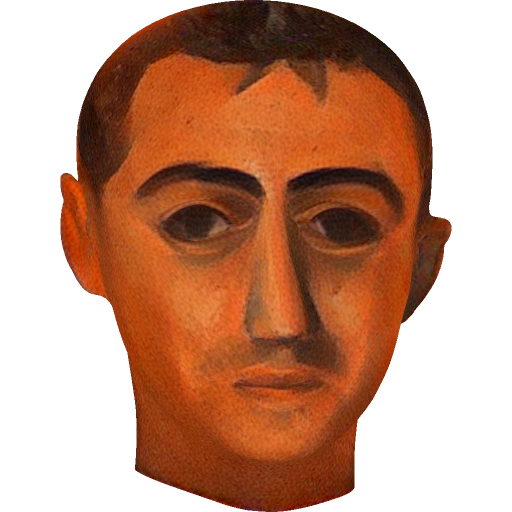} &
  \includegraphics[width=0.11\linewidth]{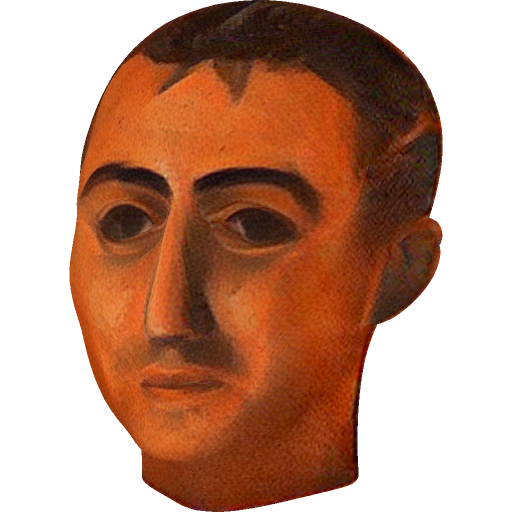} &
  \includegraphics[width=0.11\linewidth]{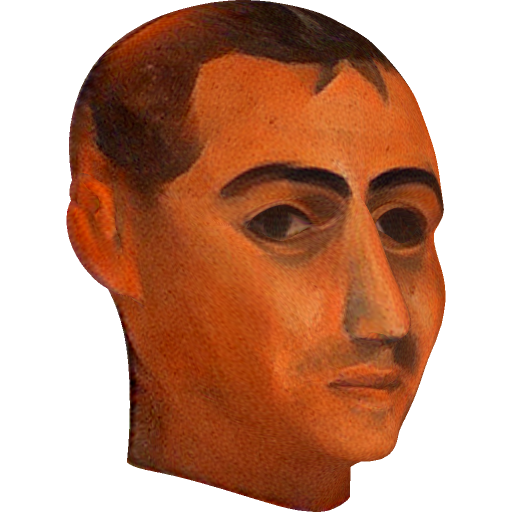} &
  \includegraphics[width=0.11\linewidth]{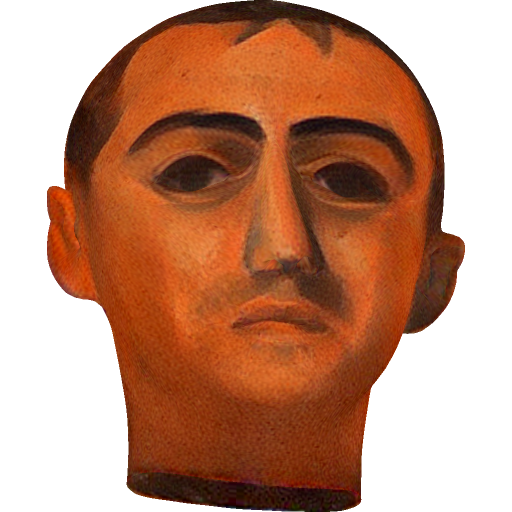} &
  \includegraphics[width=0.11\linewidth]{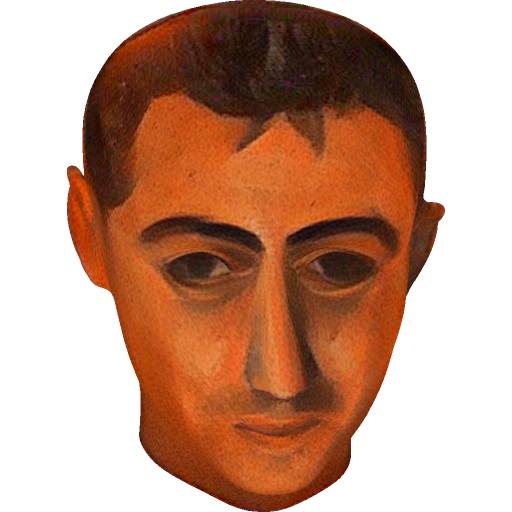} \\
  &
  \includegraphics[width=0.11\linewidth]{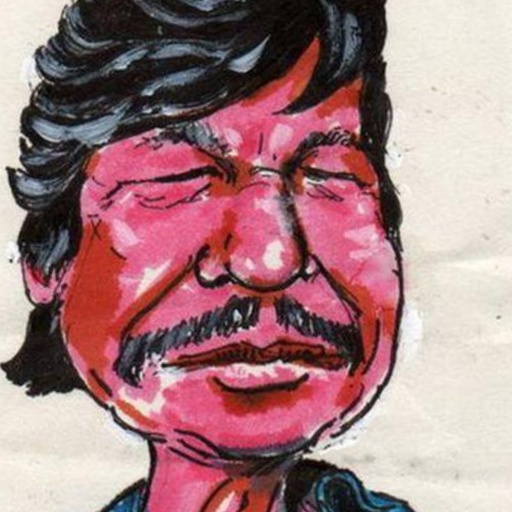} &
  \includegraphics[width=0.11\linewidth]{figures/teaser/t1224-3/res_not_00.png} &
  \includegraphics[width=0.11\linewidth]{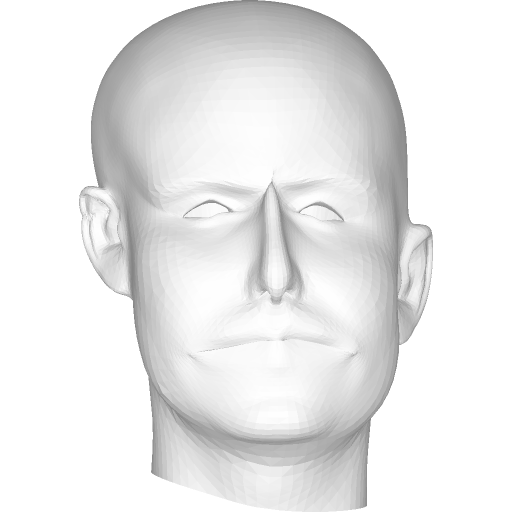}&
  \includegraphics[width=0.11\linewidth]{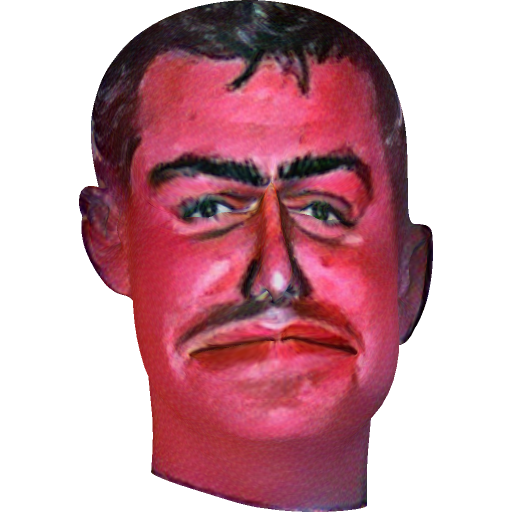} &
  \includegraphics[width=0.11\linewidth]{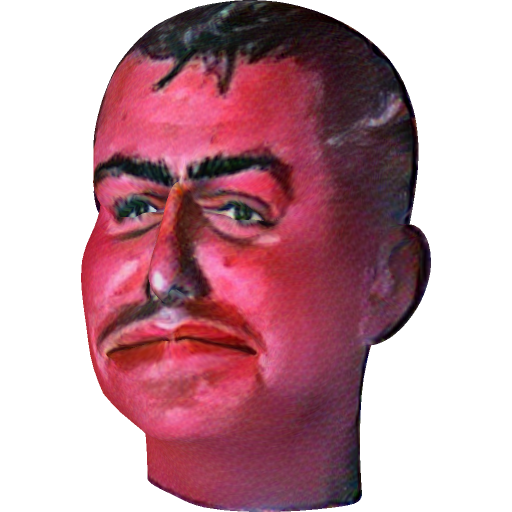} &
  \includegraphics[width=0.11\linewidth]{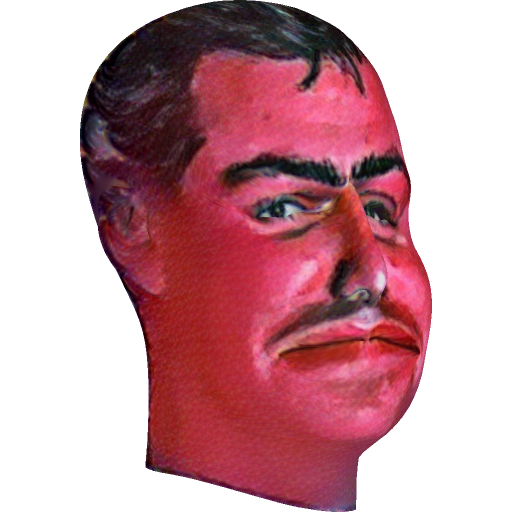} &
  \includegraphics[width=0.11\linewidth]{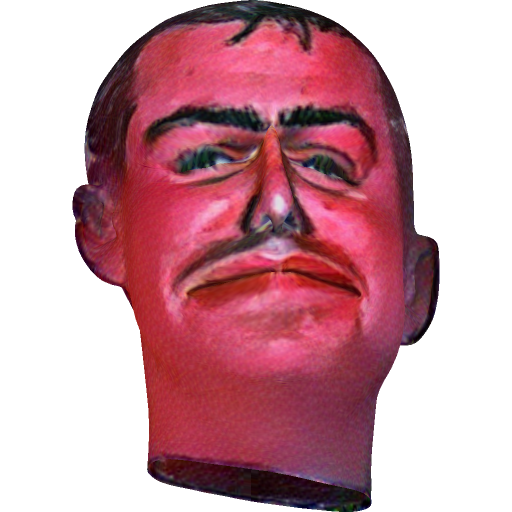} &
  \includegraphics[width=0.11\linewidth]{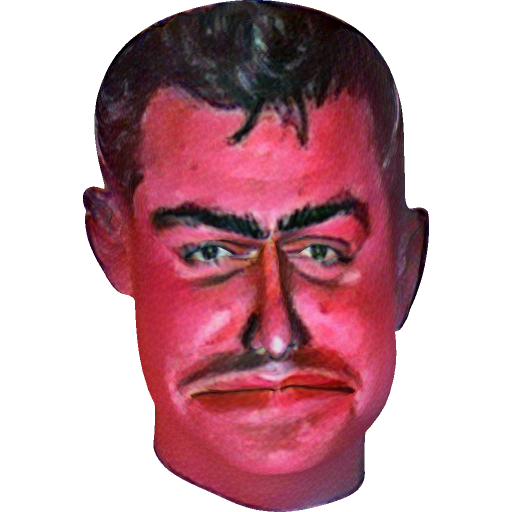} \\\hline
  & \\[\dimexpr-\normalbaselineskip+3pt]
  \includegraphics[width=0.11\linewidth]{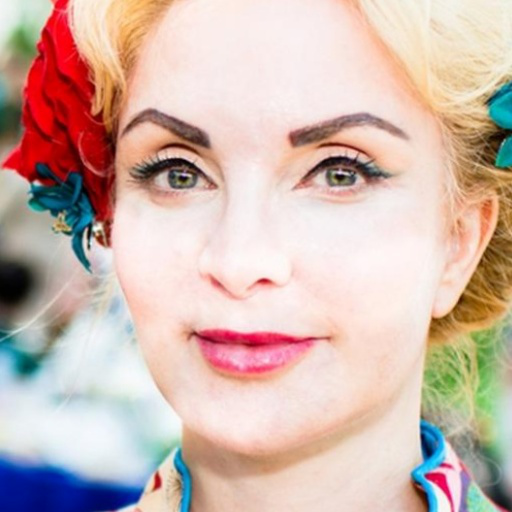} &
  \includegraphics[width=0.11\linewidth]{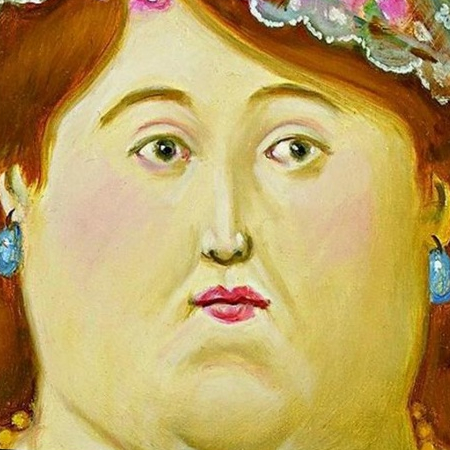} &
  \includegraphics[width=0.11\linewidth]{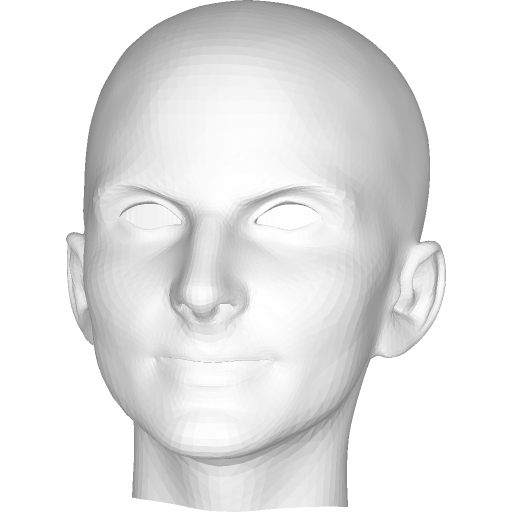} &
  \includegraphics[width=0.11\linewidth]{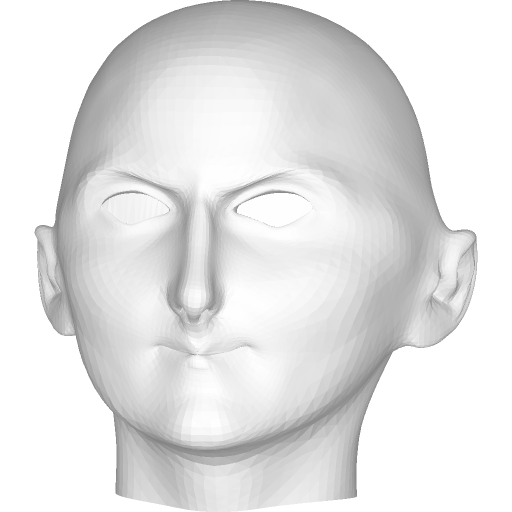}&
  \includegraphics[width=0.11\linewidth]{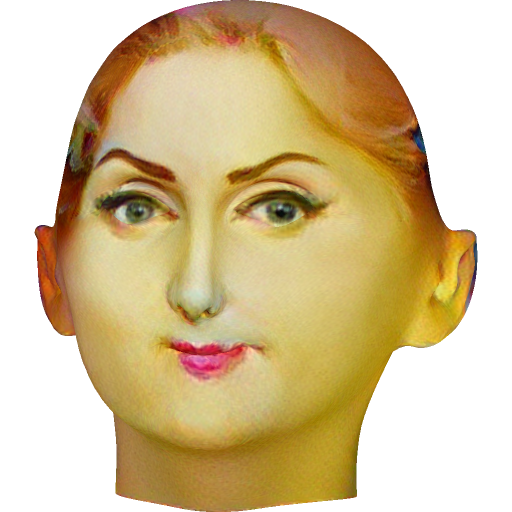} &
  \includegraphics[width=0.11\linewidth]{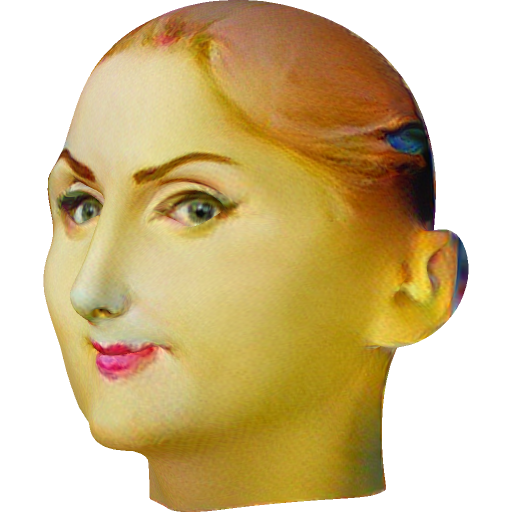} &
  \includegraphics[width=0.11\linewidth]{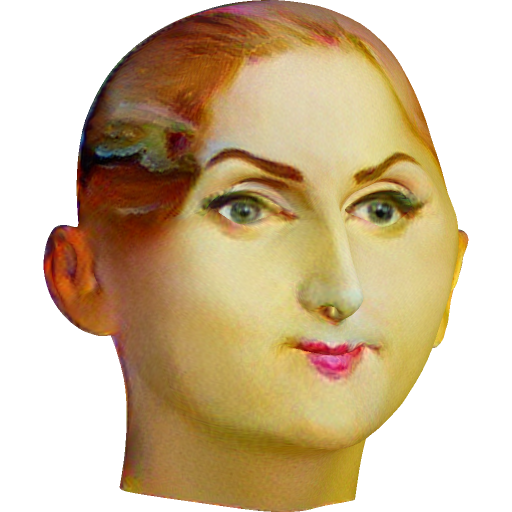} &
  \includegraphics[width=0.11\linewidth]{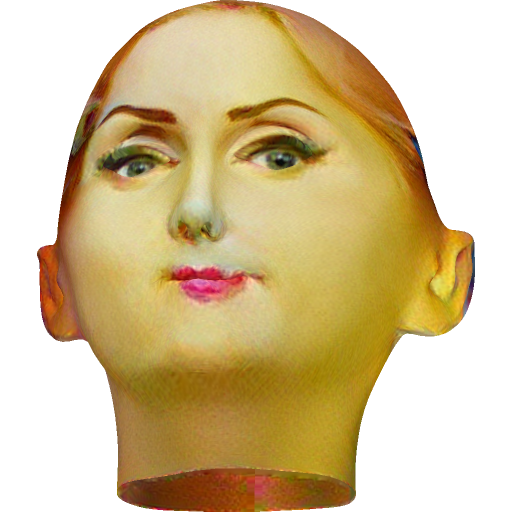} &
  \includegraphics[width=0.11\linewidth]{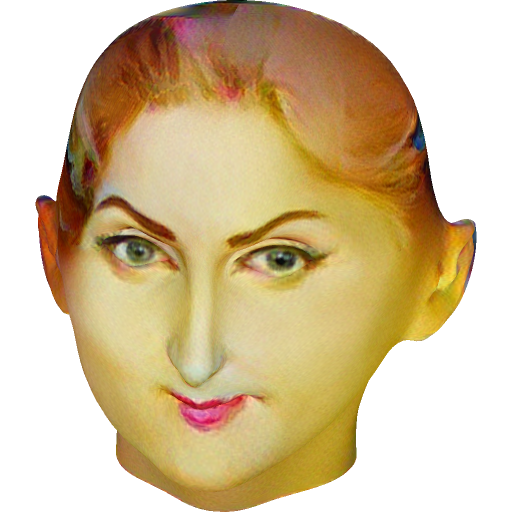} \\
  &
  \includegraphics[width=0.11\linewidth]{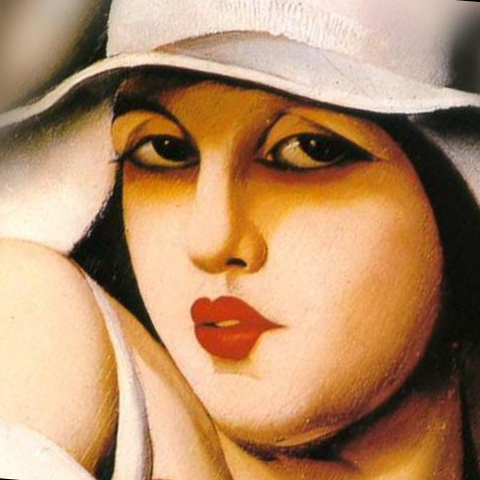} &
  \includegraphics[width=0.11\linewidth]{figures/teaser/t1224-1/res_not_00.png} &
  \includegraphics[width=0.11\linewidth]{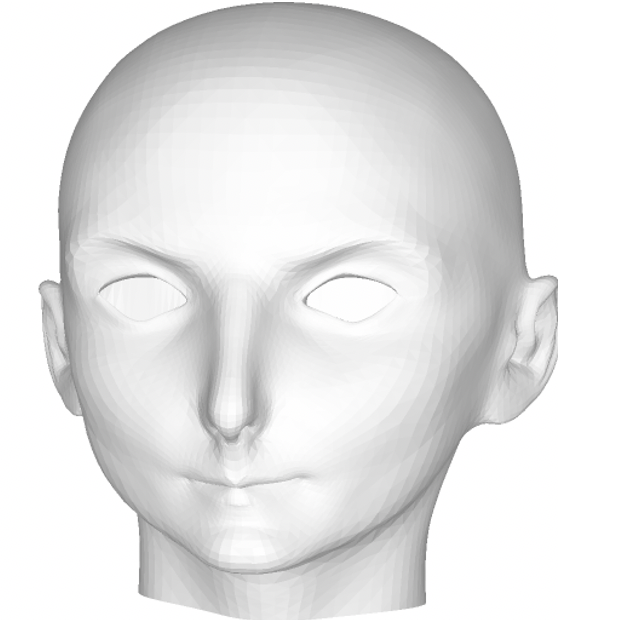}&
  \includegraphics[width=0.11\linewidth]{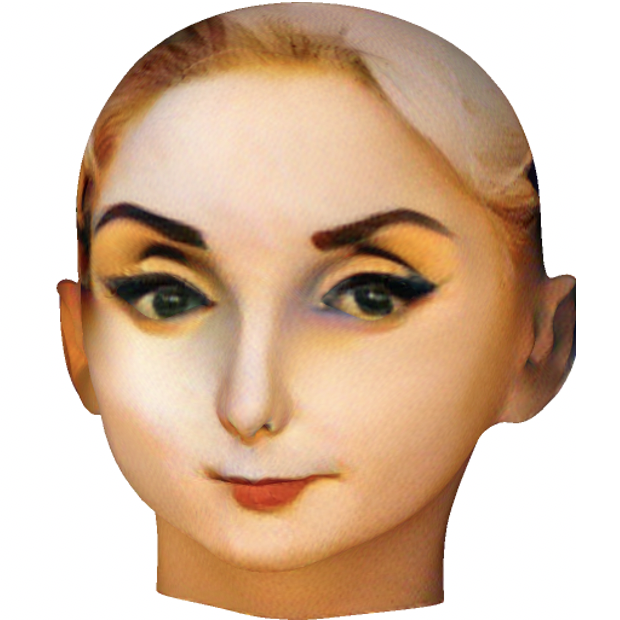} &
  \includegraphics[width=0.11\linewidth]{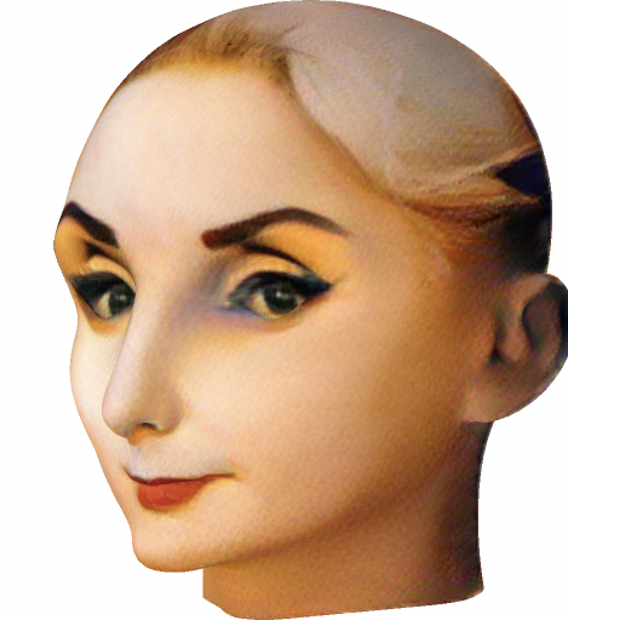} &
  \includegraphics[width=0.11\linewidth]{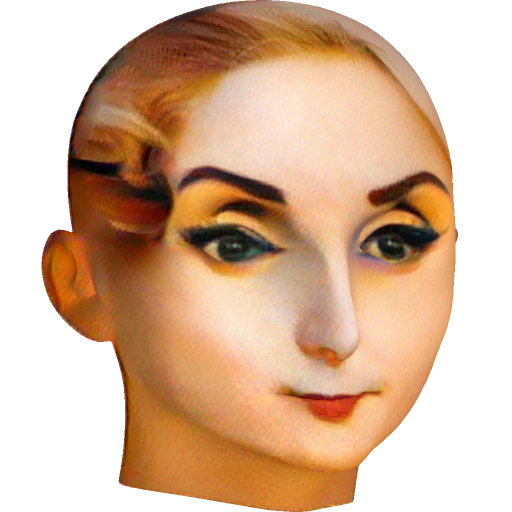} &
  \includegraphics[width=0.11\linewidth]{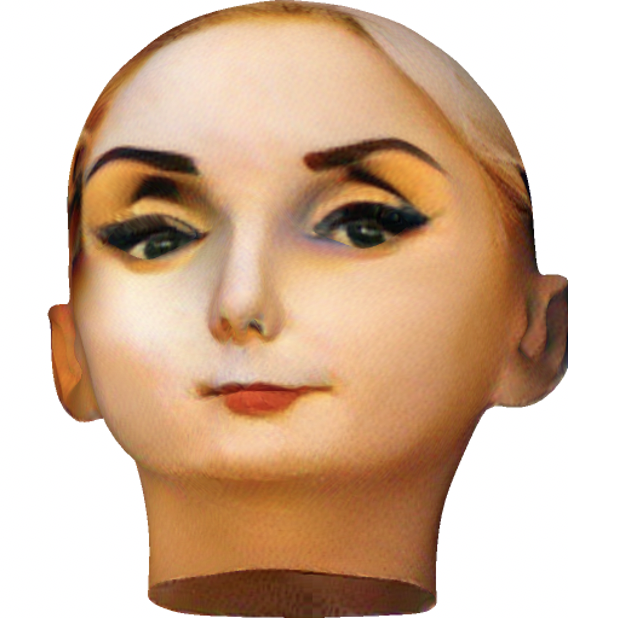} &
  \includegraphics[width=0.11\linewidth]{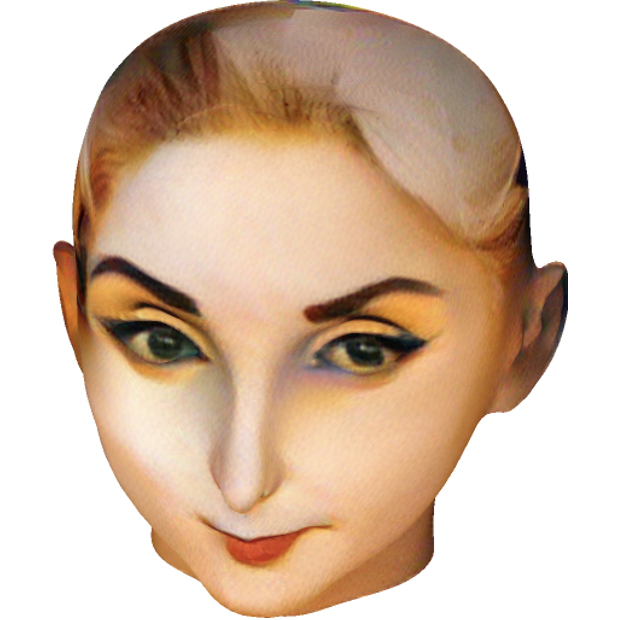} \\
  Content & Style & Rec. Geo. & Def. Geo. & \multicolumn{5}{c}{Multi-View Results}\\
  \end{tabular}}
	\caption{Our exemplar-based 3D portrait stylization method is able to generate a
stylized 3D face model with both the geometry exaggerated and the texture
transferred from a style image, while preserving the identity from the original
content portrait.}
	\label{fig:teaser}
\end{figure*}

\IEEEPARstart{F}{ine} arts featuring human faces are pervasive in our daily lives. Whether it be a piece of classical portraiture, a character from one's favorite cartoon, or even a funny caricature of a popular celebrity, we are surrounded by a vast amount of artistic interpretations of face, and constantly inspired to extend these unique visual styles to our own photos. The urge to automatically synthesize head portraits that acquire particular styles while preserving the original identities has been motivating much successful research in the past decades, especially with the striking advancement of neural-based image style transfer techniques \cite{gatys2015neural,liao2017visual,kolkin2019style}.

Artistic portrait style transfer, however, is so much more than just matching the local texture statistics. Intuitively, an artistic portrait style usually spans not only the color appearance but also the geometric shape and structure. In light of this, much encouraging effort \cite{yaniv2019face,cao2018carigans,shi2019warpgan}, instead of concentrating solely on the texture appearance, has been made recently to also handle the geometry style properly, which enables texture transfer and geometry exaggeration altogether to better convey the style perceptually. Although this pioneering progress has been made, 2D geometry style transfer methods still have the following key limitations.

First of all, texture and geometry styles are not orthogonal in the projected 2D space, they mutually interact with each other. Through direct manipulation in the original image plane, most existing 2D-based methods do not explicitly decouple texture and geometry styles from each other, which makes it difficult to accurately transfer texture style without affecting the geometry one, and vice versa, thus significantly degrading the quality.

Also, confined to the 2D image translation frameworks, reusability is highly restricted. The result, a sole image with fixed viewpoint and head pose, does not truly provide any 3D understanding of the geometry shape or disentangled parameterizations of the texture, thus forbidding further adaptations and manipulations that are particularly essential for 3D-aware graphics applications.

To overcome these limitations, we propose a \textit{disentangled 3D parameterization} for artistic portraits, which consists of two aspects: 1) a \textit{geometric shape} that captures the coarse but prominent geometric characteristics with facial landmarks, which compactly represents the geometry style and models the transferring process to ease geometry stylization with way fewer exemplars; and captures the fine geometry characteristics with a 3D mesh which is deformed with the guidance of landmarks. 2) a \textit{canonical texture} that provides an undistorted appearance parameterization, avoiding the interference of exaggerated geometry deformation. This disentangled parameterization ensures directly reusable geometry and texture, ready to drive more applications.

In this paper, with the aforementioned disentangled 3D parameterization, we present, as far as we know, the first \textit{exemplar-based 3D portrait stylization} method. Given an in-the-wild target photo and a single style exemplar, our method generates a 3D head model with texture and geometry both properly stylized from the exemplar and fully disentangled for manipulation as shown in Fig.~\ref{fig:teaser}. Specifically, our method contains two major stages. First, the \textit{geometry style transfer stage} extracts the facial landmarks from the target and translates them using an unsupervised exemplar-based landmark stylization network. Then, with the stylized coarse global shape as the guidance, the face mesh achieves dense geometric exaggeration via landmark-guided deformation. Finally, given the well-aligned canonical texture, the \textit{texture style transfer stage} progressively optimizes the final stylized texture utilizing a multi-view differentiable rendering approach.

We demonstrate the superiority of the proposed method on a variety of artistic portrait styles that are challenging for the previous image style transfer methods, especially in transferring geometry styles. More importantly, our 3D output, with geometry and texture properly parameterized and disentangled, can not only achieve image stylization but also immediately enable multiple novel 3D applications, including recognizable 3D avatars creation and reenactment, cartoon 3D portrait modeling, and personalized 3D character animation. The source code, pre-trained models, and data will be made publicly available to facilitate future research.

\section{Related work}
\subsection{Image Style Transfer}

Image stylization aims to modify the style of an input image while preserving its content, which has been extensively studied in the literature. Early works to address this problem by using hand-crafted features to map a style from one image to the content image~\cite{hertzmann2001image,efros2001image}. 
 Recently, thanks to advances of the convolutional neural network (CNN),
the pioneering work of Gatys et al. ~\cite{gatys2015neural} presents
a neural style transfer framework by leveraging deep features learned from a big dataset, which inspires a large number of following works to either improve the quality or speed. Among these methods, some are optimization-based with advanced features and well-designed losses defined between image pairs~\cite{gatys2015neural, gatys2016image,liao2017visual,gu2018arbitrary,kolkin2019style} while some are end-to-end learning methods with a neural network to perform style transfer, enabling real-time inference~\cite{johnson2016perceptual,ulyanov2016texture,chen2017stylebank,huang2017arbitrary,li2017universal,sheng2018avatar}. 

\begin{figure*}[t]
\centering
 \includegraphics[width=\linewidth]{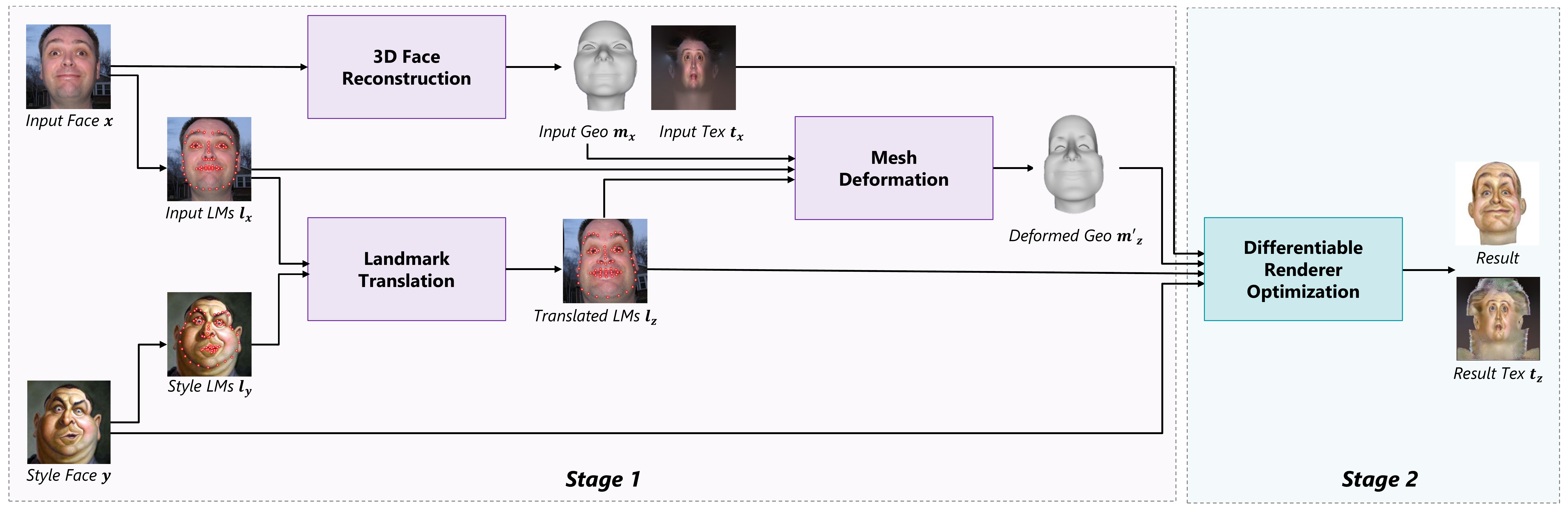}
 \caption{\textbf{System Overview.} Our system contains two stages. In the first geometry stage, we use facial landmark translation to capture the coarse geometry style and then guide the deformation of the dense 3D face geometry. In the second texture stage, we focus on performing style transfer on the canonical texture by adopting a differentiable renderer to optimize the texture in a multi-view framework. }
 \label{fig:system_overview}
\end{figure*}

Moreover, a series of works based on the generative adversarial network (GAN) for a general image-to-image translation is also widely applied to the task of style transfer~\cite{zhu2017unpaired,liu2017unsupervised,huang2018multimodal,lee2019drit}. These methods have been proposed for unpaired many-to-many image translation and achieve success on color or texture transfer with~\cite{zhu2017unpaired,liu2017unsupervised} or without ~\cite{huang2018multimodal,lee2019drit} style exemplar. Compared to neural style transfer, this kind of method's limitation is that it can only support stylization images into a specific domain, such as emoji, manga, or Van Gogh's painting. It does not allow arbitrary style transfer.
Besides the above general-purpose methods, there are approaches focusing on the specific domain of portrait style transfer with specific properties of head portraits taken into account~\cite{selim2016painting,fiser2017example,futschik2019real}.

However, the successes of the above image stylization methods are all limited to color and texture transfer, unable to handle geometry style transfer (\textit{e.g.} shape deformation of content). 
There are only some portrait style transfer works considering geometry deformation (\textit{e.g.} caricature, cartoon, or artistic portraiture). The pioneering works of caricature generation~\cite{brennan1985dynamic,akleman1997making} design interactive tools to implement face shape exaggeration by modifying the face features from average represented as drawing lines. To automate this exaggeration procedure, some approaches~\cite{liang2002example,cao2018carigans,li2018carigan} propose to learn exaggeration prototypes from a large-scale dataset with 2D face landmarks as shape representation. Among them, the representative work~\cite{cao2018carigans} proposes to decouple geometry deformation and texture transfer and learns the geometric mapping in landmark space from real faces to caricatures using Generative Adversarial Networks (\textit{i.e.}, CycleGAN)~\cite{zhu2017unpaired}. Recently, another GAN-based method~\cite{shi2019warpgan} proposes to jointly learn texture rendering and geometric warping to enhance spatial variability. These methods achieve impressive results compared to those with texture transferred only, but they rely on facial features of caricatures and fail to generalize to other artistic portraits. The work~\cite{yaniv2019face} extends geometry style transfer from the caricature domain to various artistic styles based on landmark statistics learned from an specialized artistic portraits domain. However, all these image-level geometry style transfer methods are limited by the fact that geometry and texture are not orthogonal in the projected 2D space. They mutually interact with each other, which makes it difficult to accurately transfer texture style without affecting the geometry one and vice versa. In contrast, we propose a disentangled 3D parameterization for artistic portrait stylization to overcome these limitations and support more 3D applications.

\subsection{3D Style Transfer}
A simple extension of 2D style transfer into 3D domain is to generalize neural style transfer to 3D meshes by applying image style transfer on renderings. The main idea behind these methods is the differential rendering, which allows the backpropagation of style transfer objectives form the image domain to the 3D domain by using approximate derivatives~\cite{kato2018neural} or analytical derivatives~\cite{liu2018paparazzi, mordvintsev2018differentiable}. According to whether the vertices, texture, or both are allowed to be changed during the backpropagation, existing 3D style transfers achieve three different effects. The method~\cite{liu2018paparazzi} proposes a general-purpose back-end optimization to edit 3D mesh vertex positions only according to image changes. When it is applied to style transfer, some fine-scaled deformations on the mesh appears to simulate the reference style.  In contrast, an under-explored tool~\cite{mordvintsev2018differentiable} optimizes the texture of the 3D object for stylization, but keeps its geometry unchanged. \cite{kato2018neural} provides gradients of an image with respect to the vertices and textures of a mesh so as to realize both texture and fine-scaled geometry style transfer using loss functions defined on 2D images. However, these three methods are general methods for all kinds of 3D models and style images, and thus they will generate less satisfying results for portraits without considering the face semantics in both geometry and texture aspects.

3D portrait stylization is also a popular topic in computer graphics and most related works focus on 3D caricature model creation from a normal 3D face model. This is typically implemented by detecting distinctive facial features and then exaggerating them on the face mesh using deformation techniques. Some methods~\cite{lewiner2011interactive,vieira2013three} exaggerate an input 3D model by magnifying the differences between the input model and a template model while some works~\cite{liu2009semi,wu2018alive} attempt to model 3D caricatures from 2D images. Recently, some deep learning-based interactive systems~\cite{han2017deepsketch,han2018caricatureshop} are proposed for modeling 3D caricature faces by interactively drawing sketches. The approach~\cite{ye2020carigan} builds a 3D parametric model for caricature face reconstruction targeting the problem of limited extrapolation capability of the existing parametric models based on normal faces, and support a certain level of user control on the reconstructed faces. Our work addresses a new challenge of automatically transferring geometry and texture styles from an arbitrary portrait art image to a 3D face model, which is not limited in the caricature domain.


\section{System Overview}
Given a pair of a portrait art image $y \in Y$ as the reference style and a portrait photo $x \in X$ as the content, our method aims to generate a stylized 3D face model $m_z$, with both the geometry (represented by the vertices $v_z$) exaggerated and the texture $t_z$ transferred from $y$, while preserving the identity of $x$, as shown in Fig.~\ref{fig:teaser}.

However, with such a complicated task that involves one-shot geometry translation, multi-view texture stylization, and 3D reconstruction \& parameterization, an end-to-end solution that jointly learns both geometry and texture could be difficult. Therefore, we propose a two-stage framework, as shown in Fig.~\ref{fig:system_overview}. Specifically, in the first stage, we focus on the geometry style and disentangle the texture from the geometry. We capture the coarse geometry style with facial landmark translation, which is then used to guide the generation of the 3D face geometry $v_z$ as a joint reconstruction and deformation step. With $v_z$, we project the texture onto the parameterization space to make it independent of the geometry. While in the second stage, we focus on performing style transfer on the canonical texture while fixing the geometry. We adopt a differentiable renderer to render the textured model into different views and optimize the texture $t_z$ via a multi-view framework. In the following sections, we will describe these two stages in detail.

\section{Geometry Style Transfer}
\label{sec:geo}
This section presents the geometry style transfer stage, which learns the geometry style from a piece of portrait art $y$ and applies it to a 3D face model $m_x$ reconstructed from a portrait photo $x$. Since the 2D geometry style cannot be directly applied to a 3D mesh, we leverage the landmarks as a bridge representing the coarse global shape. The geometry information of $x$ and $y$ is represented by their corresponding landmarks $l_x \in L_X$ and $l_y \in L_Y$ respectively. And we train a network to map the landmarks from the normal face domain $L_X$ to the artistic face domain $L_Y$. Instead of using a single-modal translation network whose result cannot be controlled by the reference $l_y$ as in CariGANs\cite{cao2018carigans}, we design a novel multi-modal landmark translation network that supports translating input landmarks $l_x$ to $l_z$ with the geometry style of the reference $l_y$. 
With the translated landmarks $l_z$, we then deform the 3D mesh of $m_x$ using landmark-guided Laplacian deformation to generate the stylized geometry $v_z$, while preserving its texture unchanged.

\subsection{Multi-Modal Landmark Translation Network}

\begin{figure}[t]
\centering
\setlength{\tabcolsep}{0.1mm}{
\begin{tabular}{cc}
  Class 1 &
  \includegraphics[width=0.85\linewidth]{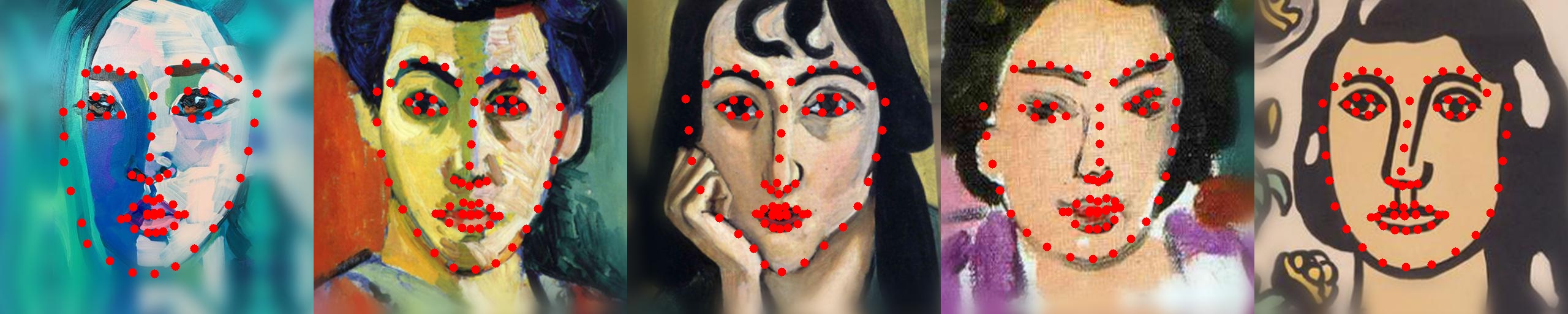} \\
  Class 2 &
  \includegraphics[width=0.85\linewidth]{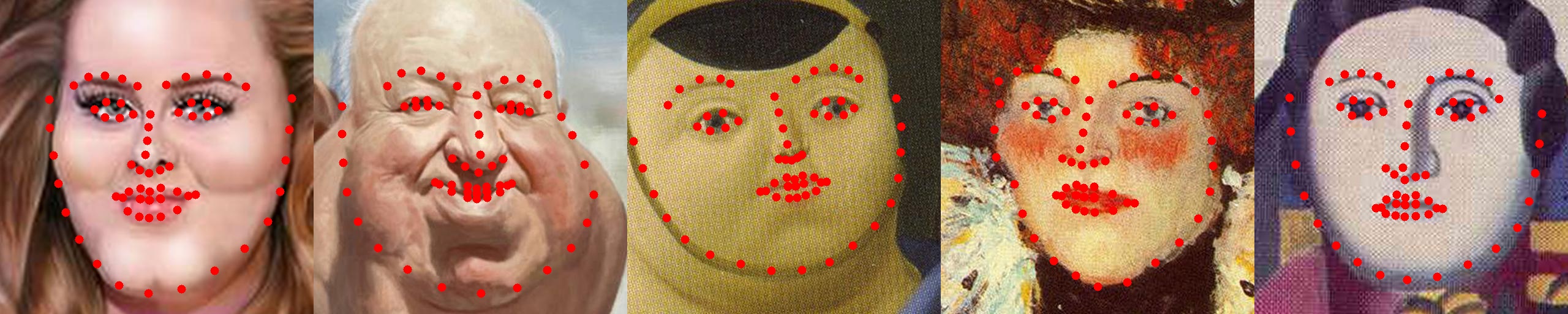} \\
  Class 3 &
  \includegraphics[width=0.85\linewidth]{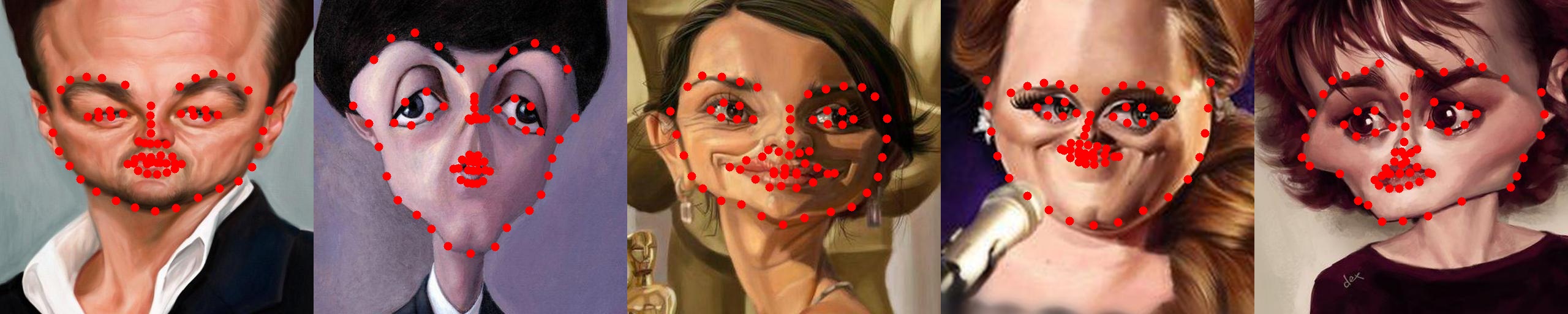} \\
   &
  \includegraphics[width=0.7\linewidth]{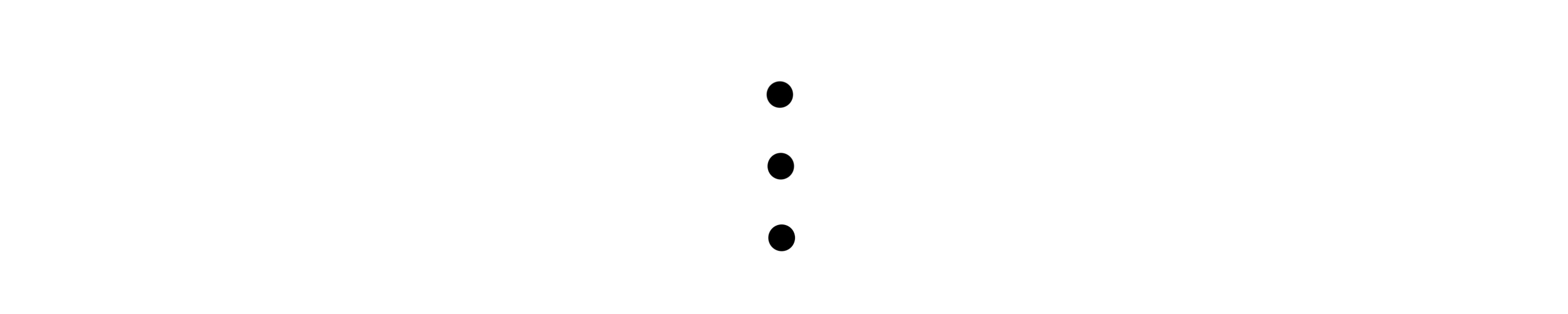} \\
  Class 25 &
  \includegraphics[width=0.85\linewidth]{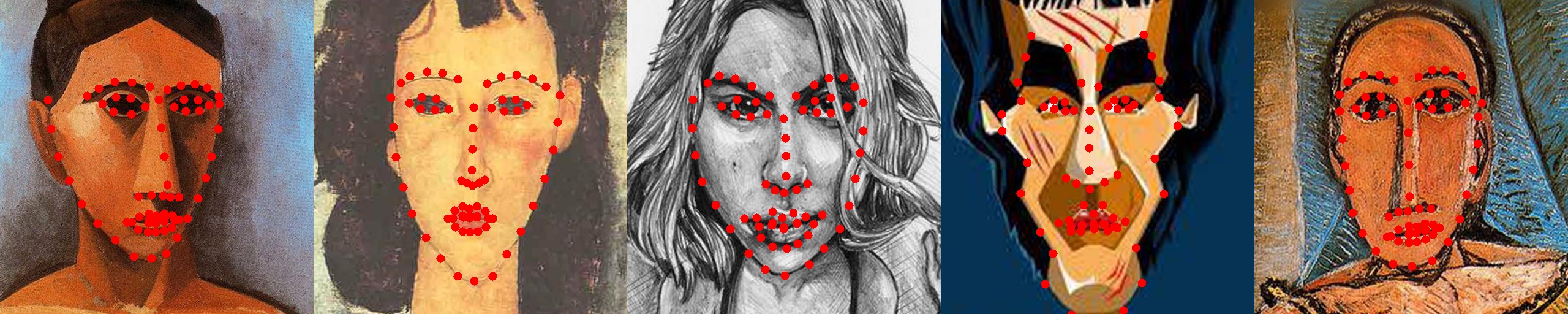} \\
  \end{tabular}}
 \caption{\textbf{Visualization of samples from different classes with landmarks.}}
 \label{fig:landmark_class}
\end{figure}

\textbf{Training Data.}
Face shape can be represented by 2D face landmarks for both real portrait photos and portrait arts. We use DLib~\cite{dlib} and the face of art \cite{yaniv2019face} to automatically detect $68$ face landmarks for real photos and artistic images, respectively, and then slightly adjust the landmark locations by manual to increase accuracy. To centralize the facial shapes, all landmarks are aligned to the average face via three points (center of two eyes and center of the mouth) using the affine transformation. A total of $6269$ samples are collected for the normal face domain $L_X$ and $11415$ samples are collected for the art domain $L_Y$. To further reduce the dimensionality of landmark representation, we perform principal component analysis (PCA) on the landmarks of all samples in $l_x$ and $l_y$. We take the top $32$ principal components to recover $99.58\%$ of total variants. Then the $68$ landmarks of each sample are represented by a vector of $32$ PCA coefficients. This representation also helps constrain the face structure during the mapping learning. In addition, for the samples in $l_y$, we cluster them into $25$ classes by using the K-means algorithm. 
The clustering result can be considered as the class of geometry style. We assume that the landmarks within the same class should have similar shape styles, which will be used to define the classification loss (Eq.~\ref{eq:class}) in the network training. Fig.~\ref{fig:landmark_class} shows several aligned samples with overlaid landmarks from different classes. 

\begin{figure}[t]
\centering
 \includegraphics[width=\linewidth]{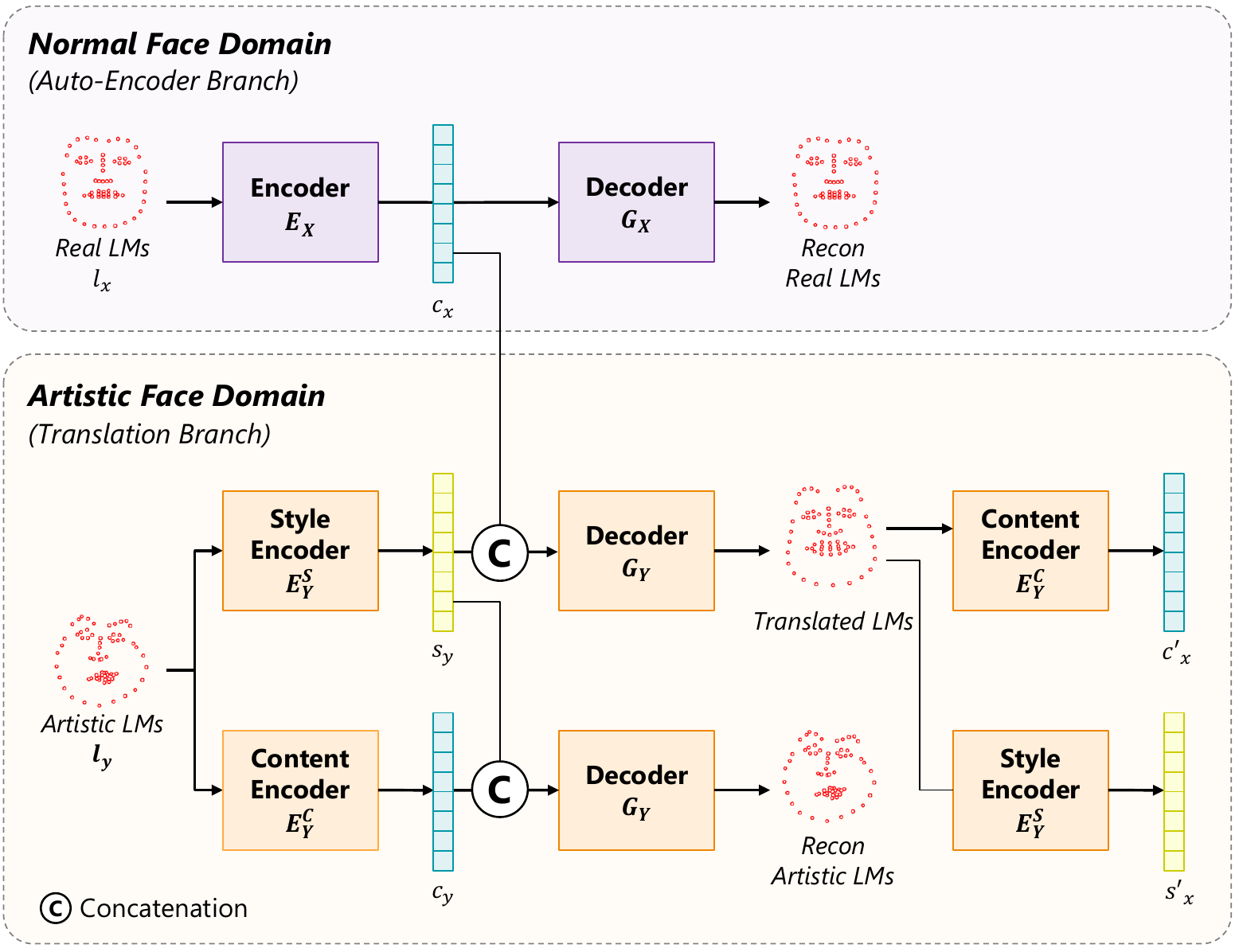}
 \caption{\textbf{Architecture of the multi-modal landmark-to-landmark translation network.}}
 \label{fig:landmark_network}
\end{figure}

\textbf{Network Architecture.} Since samples in $L_X$ and $L_Y$ are unpaired, we should adopt one unpaired translation network to learn the mapping $L_X \rightarrow L_Y$. CariGAN~\cite{cao2018carigans}, which is inspired by the unpaired image-to-image translation network CycleGAN~\cite{zhu2017unpaired}, has shown success in learning the landmark mapping from the normal face domain to the caricature domain. However, CycleGAN is a single modal network that cannot control its result by using the reference landmark $l_y$, and thus it does not satisfy the requirement of the exemplar-based style transfer. Given that, we turn to the multi-modal image-to-image translation networks, such as MUNIT~\cite{huang2018multimodal} and DRIT++~\cite{lee2019drit}. Our first attempt is to adapt these networks to handle landmarks by replacing all CONV-ReLu blocks with MLP blocks and then train them to learn the mapping $L_X \rightarrow L_Y$ with its default losses and settings. Unfortunately, this attempt is not successful because both MUNIT and DRIT++ learn the disentanglement of content and style components implicitly, but for geometry, there is no clear definition of content and style. To solve this problem, we design a novel multi-modal landmark-to-landmark translation network with explicit disentanglement as shown in Fig.~\ref{fig:landmark_network}. 

Our motivation is to define the content component as the face shape in the normal face domain while defining the style component as the artistic exaggeration beyond the normal face shape. To achieve this explicit disentanglement, the architecture of our network consists of two branches. One branch is an auto-encoder for the landmarks in the normal face domain as shown in the top row of Fig.~\ref{fig:landmark_network}. Given a content face shape $l_x \in L_X$, we extract its code $c_x$ using an encoder $E_X$, \textit{i.e.} $c_x=E_X(l_x)$, and then reconstruct $l_x$ from $c_x$ with a decoder $G_X$. Through this branch, the latent space of the code is learned to represent the shape of a normal human face.

The other branch as shown in the bottom row of Fig.~\ref{fig:landmark_network} aims to learn the translation. Given an art face shape $l_y \in L_Y$, the content encoder $E_Y^C$ and the style encoder $E_Y^S$ factorize it into a content code $c_y$ and a style code $s_y$ respectively, \textit{i.e.} $(c_y,s_y)=(E_Y^C(l_y),E_Y^S(l_y))$. 
We enforce that the output of two distinct networks, $E_Y^C$ and $E_X$, falls into the same domain.
In this way, we explicitly define the the content code of the art landmarks as the corresponding normal human face shape, while the style code as the complement, representing the artistic shape aggregation component from the art face. 
By recombining $c_y$ and $s_y$, the decoder $G_Y$ can reconstruct the input art landmarks $l_y$, while by combining $c_x$ and $s_y$, $G_Y^C$ can transfer the geometry style from $l_y$ to $l_x$ and generate the translation result in the art domain $L_Y$, denoted as $l_z=G_Y(c_x,s_y)$. 

\textbf{Losses.}
To simplify the training, we first train the auto-encoder branch and then fix it to train the translation branch. 

The auto-encoder branch is constrained by a single reconstruction loss which penalizes the L1 distance between the landmarks of a normal face and the result landmarks reconstructed from the encoder's output code, \textit{i.e.},
\begin{equation}
    \mathcal{L}_{\text {recon}}^{X}(E_X, G_X)=\mathbb{E}_{l_x \sim L_X}\left[\left\|G_X\left(E_X\left(l_x\right)\right)-l_x\right\|_{1}\right].
\end{equation}
Similarly, the translation branch also includes a reconstruction loss, which penalizes the L1 distance between the landmarks of an artistic face and the result landmarks reconstructed using the content code and style code, \textit{i.e.},
\begin{equation}
\resizebox{1.04\hsize}{!}{
    $\mathcal{L}_{\text {recon}}^{Y}(E_Y^C,E_Y^S,G_Y)=\mathbb{E}_{l_y \sim L_Y}\left[\left\|G_Y\left(E_Y^C\left(l_y\right),E_Y^S\left(l_y\right)\right)-l_y\right\|_{1}\right]$}.
\end{equation}

For the translation result $l_z$ generated by combining the encoding $c_x=E_X(l_x)$ of the normal face domain and the style code $s_y=E_Y^S(l_y)$ of the artistic face domain, we define an adversarial loss, which encourages generating landmarks indistinguishable from the real landmarks sampled from the domain $L_Y$:
\begin{equation}
\resizebox{1\hsize}{!}{$\begin{array}{ll}
\mathcal{L}_{\mathrm{adv}}^{X}(E_X,E_Y^S,G_Y,D_Y) &=\mathcal{L}_{\mathrm{adv}}^{X}(E_X,E_Y^S,G_Y,D_Y)\\[.25cm]
&+\:\mathbb{E}_{l_y \sim L_Y }\left[\log D_Y\left(l_y\right)\right],\end{array}$}
\end{equation}
where $C$ and $S$ denote the domains of content and style codes respectively. $D_Y$ is the discriminator learned to distinguish between real and fake samples in the domain $L_Y$.

To further encourage the generated landmarks $l_z$ to be faithful to the geometry style of $l_y$, we feed them to a classifier $CL$ and require them to be classified to the same class of $l_y$ by imposing a classification loss: 
\begin{equation}
\label{eq:class}
\resizebox{1.04\hsize}{!}{$
 \mathcal{L}_{class}(E_X,E_Y^S,G_Y,CL)=\mathbb{E}_{c_x \sim C, s_y \sim S}\left[\mathcal{H}\left(CL \left(G_Y\left(c_x, s_y\right)\right), lb_y\right)\right]
 $}
 \end{equation}
where $\mathcal{H}$ denotes the cross entropy function, and $lb_y$ is the class label of the input $l_y$. 

Beside the constraints on the generated results, we also define losses on the content code and style code, to guarantee the accuracy of their disentanglement. We further feed the generated translation result into the content encoder $E_Y^C$ and the style encoder $E_Y^S$ to re-factorize it into a content code $c'_x$ and a style code $s'_y$ respectively. For the content code $c'_x$, we require it to be the same as the code $c_x$ of the normal face landmarks $l_x$ generated by the encoder $E_X$:
\begin{equation}
     \mathcal{L}_{\text {recon}}^{C}(E_Y^C,G_Y)=\mathbb{E}_{c_x \sim C, s_y \sim S}\left[\left\|E_Y^C\left(G_Y\left(c_x, s_y\right)\right)-c_x\right\|_{2}\right].
\end{equation}

This loss term enforces that the content codes generated by $E_Y^C$ and the photo codes generated by $E_X$ are in the same domain, thus explicitly specifying the disentanglement of the content component. The style code is the complementary to the content code, and we also require $s'_y$ to reconstruct the input style code $s_y$:
\begin{equation}
  \mathcal{L}_{\mathrm{recon}}^{S}(E_Y^S,G_Y)=\mathbb{E}_{c_x \sim C, s_y \sim S}\left[\left\|E_Y^S\left(G_Y\left(c_x, s_y\right)\right)-s_y\right\|_{2}\right].
 \end{equation}

Moreover, the KL divergence loss $\mathcal{L}_{\mathrm{KL}}$ is employed to match the distribution of style code to be a Gaussian distribution $\mathcal{N} (0, 1)$. Note that the KL divergence loss is necessary. Without it, the style encoder may degenerate, \textit{i.e.}, the content code may contain most of the information, but the style code can still satisfy almost all constraints even if it stays the same.

In summary, the auto-encoder branch is first trained with $\mathcal{L}_{\text {recon}}^{X}$, and then the translation branch is trained by optimizing the combined loss function:
\begin{align}
\mathcal{L}_{translation}&=\lambda_{\text {recon}}^{Y} \mathcal{L}_{\text {recon}}^{Y}\nonumber +\lambda_{\text {recon}}^{C} \mathcal{L}_{\text {recon}}^{C}
+\lambda_{\text {KL}} \mathcal{L}_{\text {KL}}\\
&+\lambda_{\text {recon}}^{S} \mathcal{L}_{\text {recon}}^{S}
+\lambda_{\text {adv}} \mathcal{L}_{\text {adv}}
+\lambda_{\text {class}} \mathcal{L}_{\text {class}},
\end{align}
{where $\lambda_{\text {recon}}^{Y}$, $\lambda_{\text {recon}}^{C}$, $\lambda_{\text {KL}}$, $\lambda_{\text {recon}}^{S}$, $\lambda_{\text {adv}}$ and $\lambda_{\text {class}}$ are weights to balance the multiple objectives. They are set to $1$, $0.5$, $1$, $1$, $1$ and $1$ respectively in our experiments.}

\textbf{Implementation Details.} To incorporate the PCA landmark representation, we take MLP and FC layers as the building blocks of our network. Specifically, each content encoder consists of $8$ MLP layers, the style encoder consists of $16$ MLP layers, and each decoder consists of $8$ MLP layers. The discriminator and classifier consist of $4$ FC layers. The Adam optimizer is used with a fixed learning rate of $0.0005$, a batch size of $68$ data items, and $800$ epochs. Training the landmark translation network takes $20$ hours for each stage, on a workstation with a single NVIDIA V100 GPU.

\subsection{3D Face Deformation}
{In this step, we reconstruct the 3D geometry of the content face $x$ in the space of morphable models, which parameterizes the dense geometry with a consistent and well-corresponded topology. With it, we can use the translated landmark $l_z$ to guide the deformation of the 3D face models, generating deformed models that reveal the geometry style from $y$. }

\textbf{Face Model Reconstruction.} The 3D face geometry $m_x$ of the input portrait image $x$ is reconstructed with FaceWarehouse~\cite{cao2014facewarehouse}. It contains a vertex set $v_x$, a face set $f$, a texture map $t_x$ and a projection matrix $P$. The reconstructed model can be further represented by $m_x=(v_x, f, t_x, P)$. Among all vertices, we manually specify an index set: $ID = \{id_i, i = 1,...,68\}$, where the vertex $v_x[id_i]$ corresponds to the i-th landmark. Since all face models share the same topology, we only need to specify once.

\textbf{Landmark-Guided Laplacian Deformation.}
Given the correspondence between the subset of vertices and the translated landmark $l_z$, we can use the landmarks to guide the deformation of the mesh. During the deformation, only the coordinates of the vertices need to be optimized while face connections, texture map and projection matrix remain unchanged. Thus, we can denote the deformed face model as $m'_z=(v_z, f, t_x, P)$ with its stylized geometry $v_z$ transferred from the style reference $y$, while keeping the texture $t_x$ unchanged. The deformation is an optimization-based process with two energy terms. The landmark term $\mathcal{L}_{landmark}$ forces the specified vertex $v_z[id_i]$ to be aligned with its corresponding landmark $l_z[i]$ after projection to the 2D image plane:
\begin{equation}
    \mathcal{L}_{landmark}=\sum_{i=1}^{68}\left\|v_z[id_i]*P-l_z[i]\right\|_{2}.
\end{equation}
Here $v_z[id_i]*P$ is the process to project a 3D vertex $v_z[id_i]$ onto the 2D image plane.

Since the landmarks are sparse, $\mathcal{L}_{landmark}$ alone will only move the specified vertices while all other vertices are unchanged. To smoothly move other vertices to the target positions while preserving the local geometry details of the original mesh, we use an additional term $\mathcal{L}_{Laplacian}$ for regulation:
\begin{equation}
\mathcal{L}_{Laplacian}=\left\|Lap(v_z, f)-Lap(v_x, f)\right\|_{2},
\end{equation}
where $Lap(v, f)$ denotes the graph Laplacian matrix of a model with a vertex set $v$ and a face set $f$.

In summary, the deformation is to optimize the coordinates of the vertex $v_z$ by minimizing the energy function:
\begin{equation}
\mathcal{L}_{deformation}= \mathcal{L}_{landmark}+\alpha\mathcal{L}_{Laplacian},
\end{equation}
where {$\alpha=10^7$} is the balancing weight. The Adam optimizer with a fixed learning rate of {$0.01$} is used for the optimization.

\section{Texture Style Transfer}
\label{sec:tex}
\begin{figure}[t]
\centering
 \includegraphics[width=1.05\linewidth]{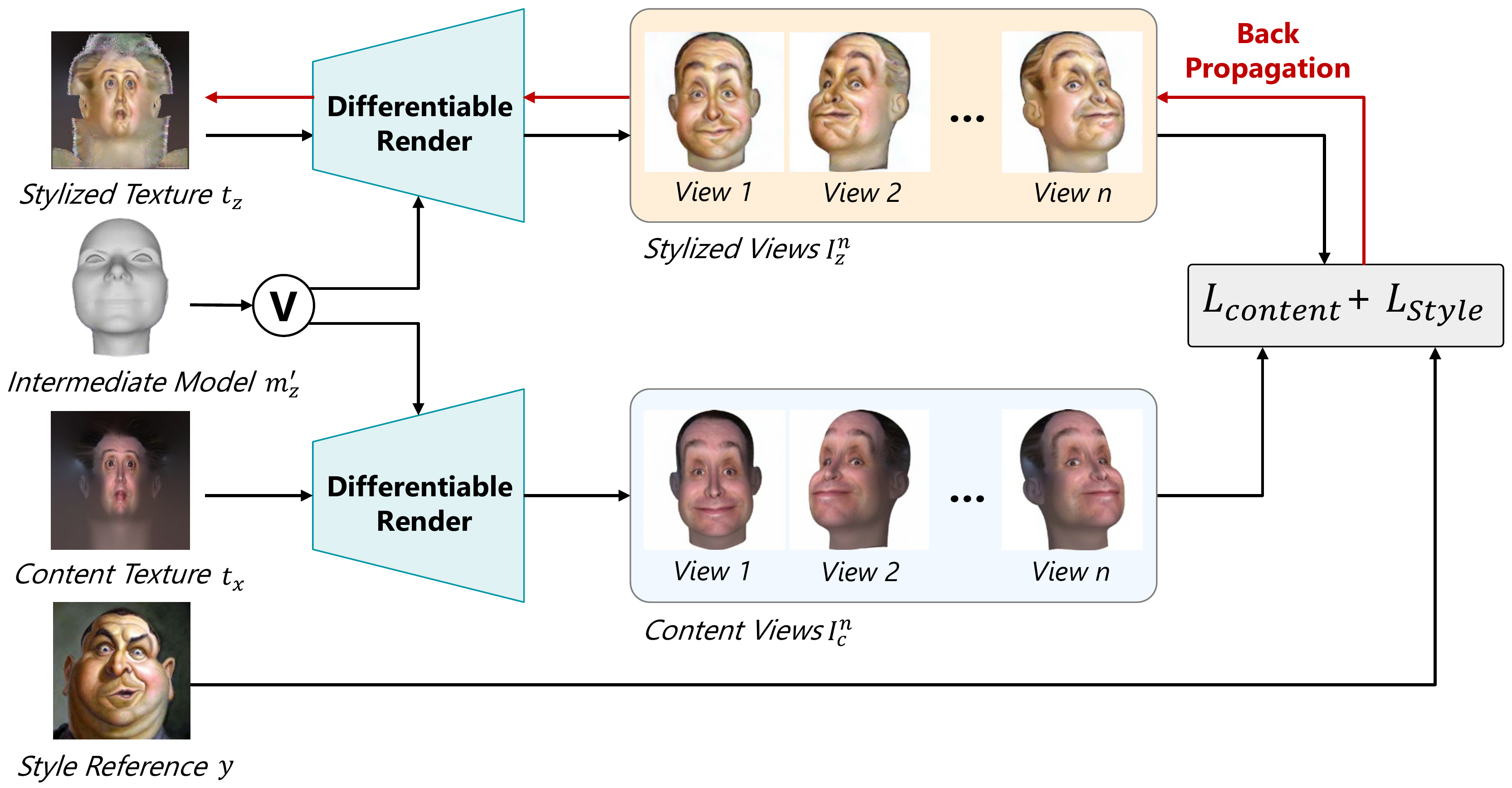}
 \caption{\textbf{Multi-view optimization framework.}}
 \label{fig:opti_framework}
\end{figure}

The intermediate model $m'_z=(v_z, f, t_x, P)$ generated in the first stage (Fig.~\ref{sec:geo}) has an artistically stylized geometry but a photo-like texture. {Since the 3D geometry is represented in a 3D morphable space, we can parameterize the texture into the canonical space using the same UV parameterization.} The second stage aims to transfer the texture style from the reference image $y$ to the parameterized content texture $t_x$, generating the final stylized model $m_z=(v_z, f, t_z, P)$ with both of the geometry $v_z$ and the texture $t_z$ stylized. One naive way to conduct this is to treat the texture map $t_x$ as an image with normal content and apply the existing image style transfer methods (\textit{e.g.} \cite{gatys2015neural}) to it. However, this naive baseline has two drawbacks. First, flattened textures tend to distort their semantic structures, resulting in poor performance of semantic-level style transfer. Second, the transferred texture patterns on the texture map can be stretched after rendering with the deformation of the 3D model, leading to unpleasant visual artifacts. To address the above challenges, we propose a multi-view optimization framework, as shown in Fig. \ref{fig:opti_framework}. We use a differentiable renderer to render model $m_z$ with texture $t_z$ into different views and impose the image style transfer loss on the frames of each view. This objective will be back-propagated through a differentiable renderer to update the texture so that we can obtain the result $t_z$ with the texture styles of $y$.

\subsection{Multi-View Optimization Framework}

For the stylized face model $m_z$, we first initialize its texture $t_z$ with the content texture $t_x$ as in the intermediate model $m'_z$, and then optimize $t_z$ using an iterative method. In the n-th iteration, we feed the intermediate model $m'_z$ and the stylized model $m_z$ into the differentiable renderer~\cite{kato2018neural} to get the rendered content views $I_c^n=\mathcal{R}(m'_z,\theta^n)$ and stylized views $I_z^n=\mathcal{R}(m_z,\theta^n)$ respectively, where $\mathcal{R}$ denotes the rendering process and $\theta^n$ denotes the view parameters. In the rendering process, we do not adopt any lighting and shading model, since the artistic style rendering does not follow any photo-realistic illumination model. Our goal is to deploy the visual attributes of $y$ onto $I_z^n$ while keeping minimum distortion of the underlying semantics and layout of $I_c^n$. This is a standard image style transfer task, so we define the style transfer loss function $L_{texture} (I_z^n,I_c^n,y)$ to optimize $I_z^n$. The rendering process is differentiable, and the gradients on $I_z^n$ can be back-propagated to both texture map $t_z$ and vertices $v_z$ of the model $m_z$. As the ambiguity in updating two variables leads to obvious artifacts on both geometry and texture (with examples shown in Sec. 2 in the supplemental document), we choose to fix $v_z$ and optimize $t_z$ only in this stage.

In each iteration, the $I_z^n$ rendered in a single view can only cover a partial region of the texture $t_z$. Therefore, we need to randomly sample the camera position in the range of horizontal angle $\pm30$ degrees and vertical angle $\pm20$ degrees to render the 3D model with an updated texture $t_z$ in the next iteration and repeat the above optimization. After several iterations, we can seamlessly combine the multi-view results to optimize the entire stylized texture $t_z$.

\subsection{Loss Function}
\label{sec:tex_loss}
According to the loss function, the existing image neural style transfer methods can be categorized into two groups. One group is to match the global statistics of deep features~\cite{gatys2015neural}, like the Gram matrix, so it transfers the global styles. The other group is to explicitly find local feature matching and thus can achieve semantic-level transfer~\cite{liao2017visual,li2016combining,kolkin2019style}, like eye-to-eye or mouth-to-mouth transfer. Considering that our inputs are both faces with clear semantic correspondences, the latter is more suitable. So our style transfer loss function $L_{texture} (I_z^n,I_c^n,y)$ is inherited from STROTSS~\cite{kolkin2019style}, one of the state-of-the-art methods belonging to the second group, which builds the feature matching via a relaxed optimal transport algorithm.

\textbf{Style Loss.} 
The style loss $\mathcal{L}_{style}$ is defined between the stylized views $I_z^n$ and the style reference $y$. We first feed these two images into a pre-trained VGG16 network and extract their multi-layer feature maps. We use bilinear upsampling to match the spatial dimensions of the feature maps of different layers to those of the input image, and then concatenate them along the channel dimension. This yields a hypercolumn at each pixel, which includes low-level features that capture edges and colors to high-level features that capture semantics. Let $A=\{A_1,...,A_k\}$ be a set of $k$ feature vectors of $I_z^n$ and $B= \{B_1,...,B_k\}$ be a set of $k$ feature vectors of $y$, where $k$ is the number of pixels. The style loss is derived from the Earth Movers Distance (EMD):
\begin{align}
    \mathcal{L}_{style}(I_z^n,y)&=EMD(A,B) \nonumber\\
    &=\min_{T\geq 0}\sum_{ij}T_{ij}Cost_{ij}\nonumber\\
    & s.t. \sum_{j}T_{ij}=i/k \quad \text{and} \quad \sum_{i}T_{ij}=i/k,
\end{align}
where $T$ is the ``transport matrix'', which defines partial pairwise assignments between pixel $i$ in $A$ and pixel $j$ in $B$. $Cost$ is the ``cost matrix'' which measures the feature cosine distance between each pixel in $A$ and each pixel in $B$, \textit{i.e.} $Cost_{ij}=cos(A_i,B_j)$. Directly optimizing the EMD problem is time-consuming, so we take the suggestion of \cite{kolkin2019style} to relax it. The details of relaxation can be found in the paper \cite{kolkin2019style}.

\textbf{Content Loss.} 
The content loss is defined between the stylized views $I_z^n$ and the content views $I_c^n$. The feature representations of both $I_z^n$ and $I_c^n$ are similarly constructed by the feature exaction and concatenation of VGG16 as described above. Let us denote $A=\{A_1,...,A_k\}$ as a set of $k$ feature vectors of $I_z^n$ and $B= \{B_1,...,B_k\}$ as a set of $k$ feature vectors of $I_c^n$, where $k$ is the number of pixels. The content loss is defined as the differences between the self-similarity patterns of two feature maps, \textit{i.e.},
\begin{equation}
\resizebox{1.04\hsize}{!}{$
    \mathcal{L}_{content}(I_z^n,I_c^n)=\frac{1}{k^2}\sum_{i,j}\left\|\frac{cos(A_i,A_j)}{\sum_{i}cos(A_i,A_j)}-\frac{cos(B_i,B_j)}{\sum_{i}cos(B_i,B_j)}\right\|.
$}
\end{equation}
In other words, the normalized cosine distance between the feature vectors extracted from any pair of pixels should remain constant between the content view and the stylized view. This is better than directly minimizing the L1 and L2 distance between two feature vectors $A_i$ and $B_i$, because self-similarity matching can preserve the semantic and spatial layout broadly, while allowing pixel values in the $I_z^n$ to differ drastically from those in $I_c^n$.

The overall loss for texture transfer is a weighted combination of two terms:
\begin{equation}
\label{eq:tex}
\mathcal{L}_{texture}=\mathcal{L}_{style}+\beta \mathcal{L}_{content},
\end{equation}
where $\beta=1.0$ is weight to balance two terms. The stylized texture $t_z$ is optimized by minimizing $\mathcal{L}_{texture}$ with the RMSprop optimizer using learning rate of $0.002$ in $600$ iterations. Then we obtain the final result, a face model $m_z=(v_z,f,t_z,P)$, with both the stylized geometry reflected in vertices $v_z$ and stylized texture reflected in the texture map $t_z$.

\begin{figure*}[htp]
  \centering
  \includegraphics[width=0.91\linewidth]{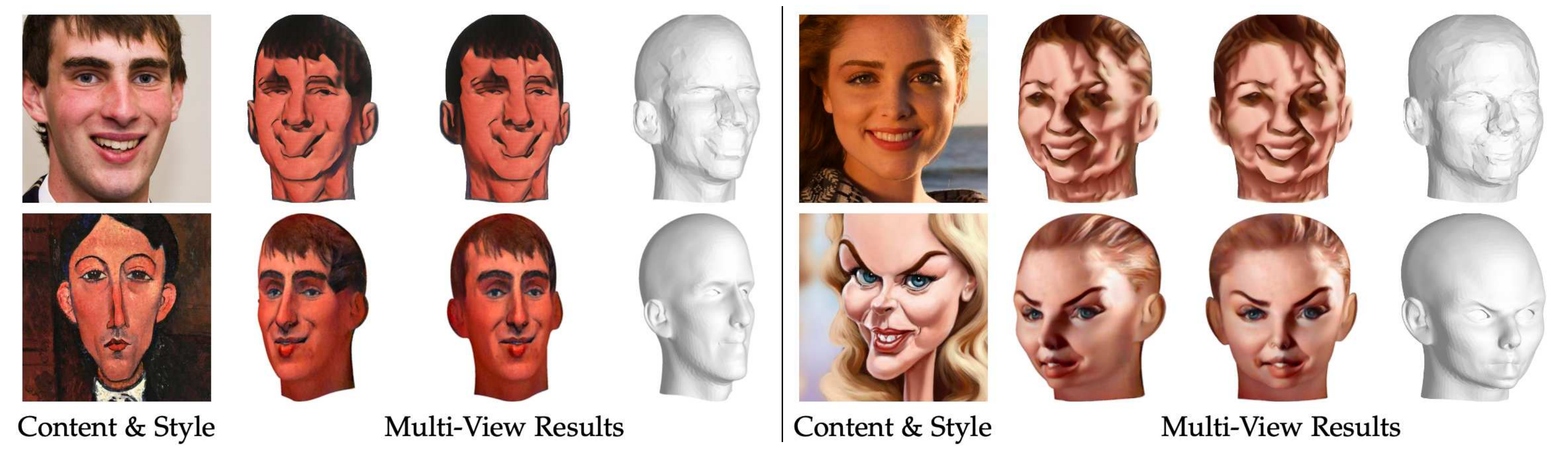}
  \caption{\textbf{Comparison between single-stage (upper row) and two-stage (lower row) frameworks.}}
 \label{fig:cmp_two_one}
\end{figure*}

\begin{figure*}[htp]
  \centering
  \setlength{\tabcolsep}{0.2mm}{
\begin{tabular}{ccccccc}
  \includegraphics[width=0.125\linewidth]{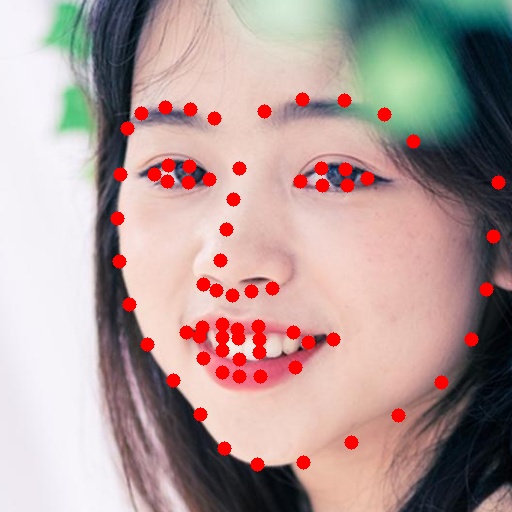} &
  \includegraphics[width=0.125\linewidth]{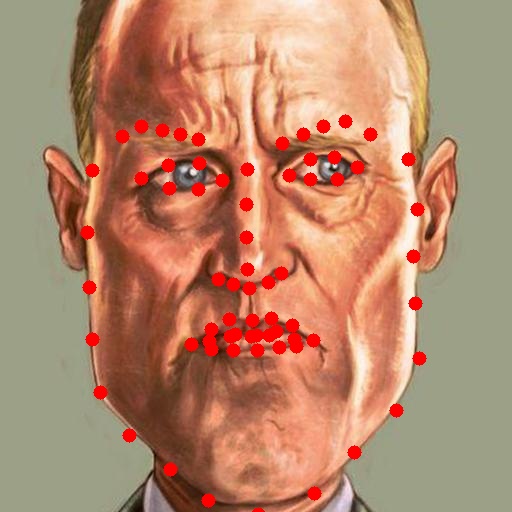} &
  \includegraphics[width=0.125\linewidth]{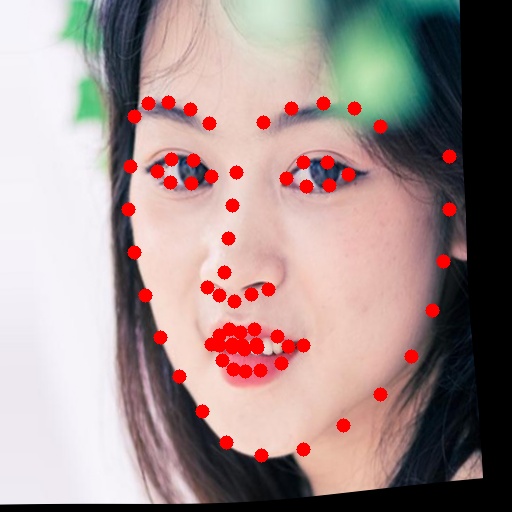} &
  \includegraphics[width=0.125\linewidth]{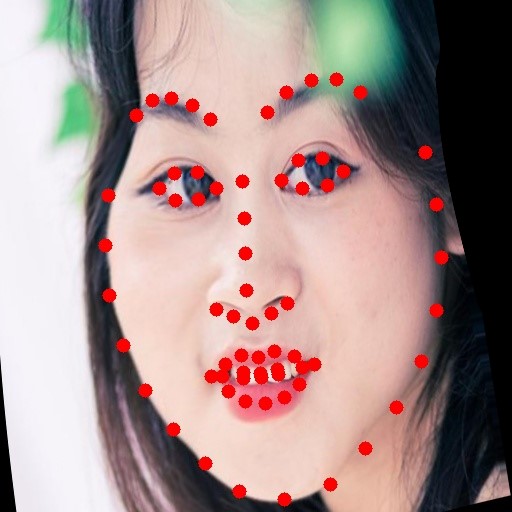} &
  \includegraphics[width=0.125\linewidth]{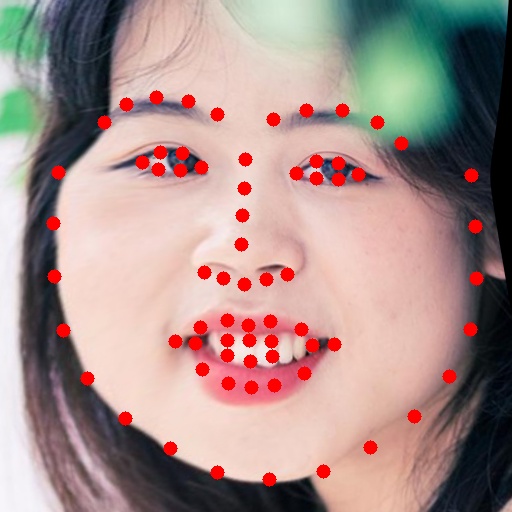} &
  \includegraphics[width=0.125\linewidth]{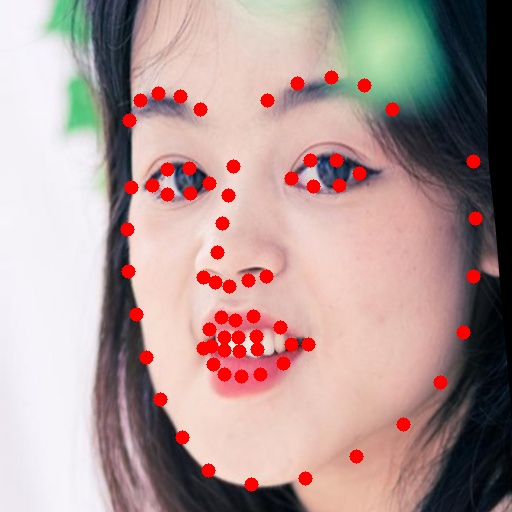} &
  \includegraphics[width=0.125\linewidth]{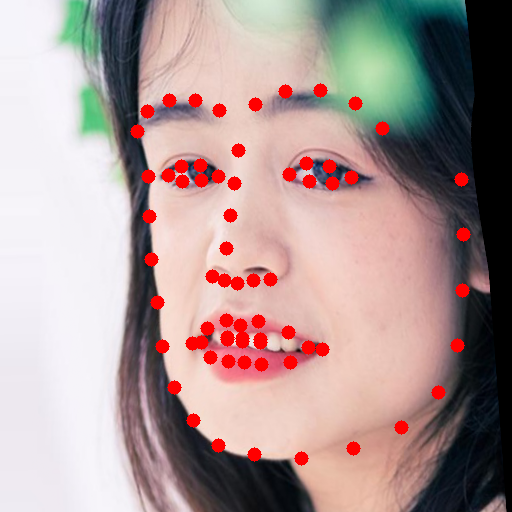} \\
   &
  \includegraphics[width=0.125\linewidth]{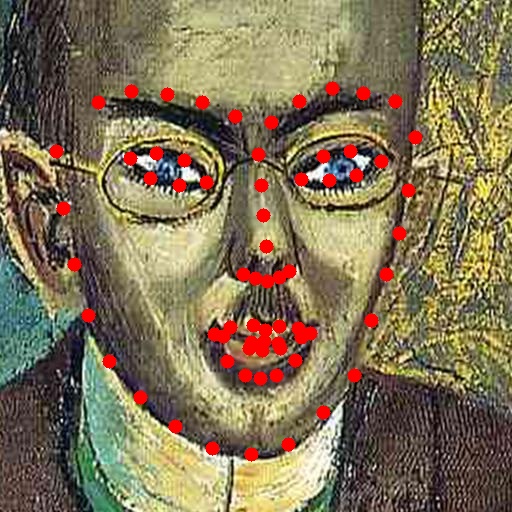} &
  \includegraphics[width=0.125\linewidth]{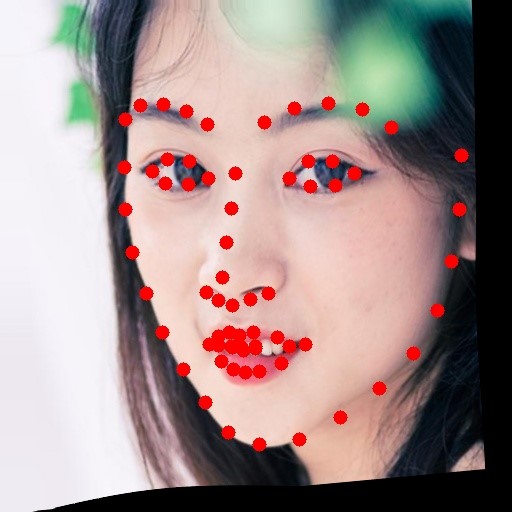} &
  \includegraphics[width=0.125\linewidth]{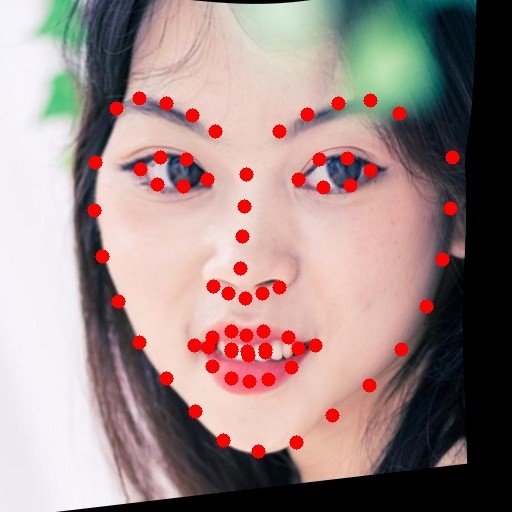} &
  \includegraphics[width=0.125\linewidth]{figures/ablation_landmark/example1/cyc.jpg} &
  \includegraphics[width=0.125\linewidth]{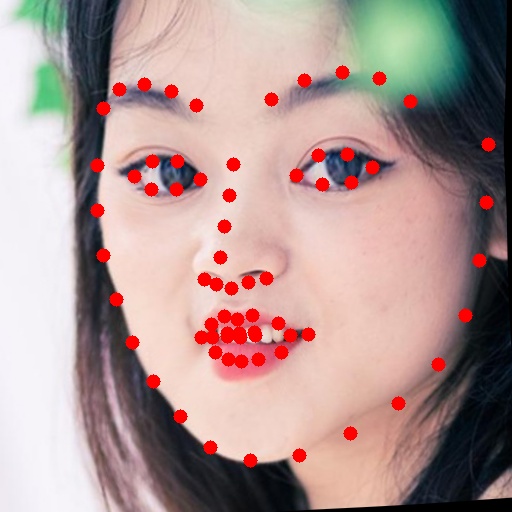} &
  \includegraphics[width=0.125\linewidth]{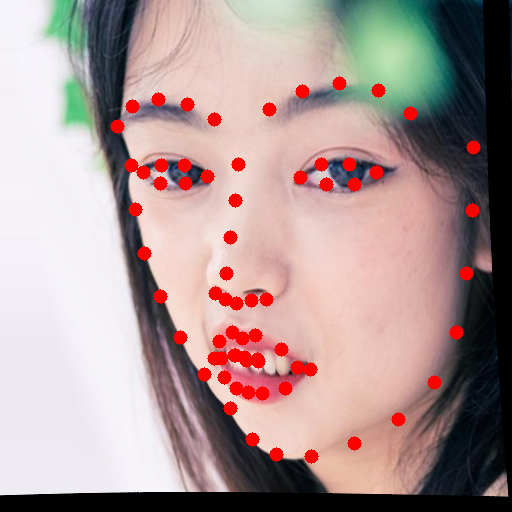} \\\hline
  & \\[\dimexpr-\normalbaselineskip+3pt]
  \includegraphics[width=0.125\linewidth]{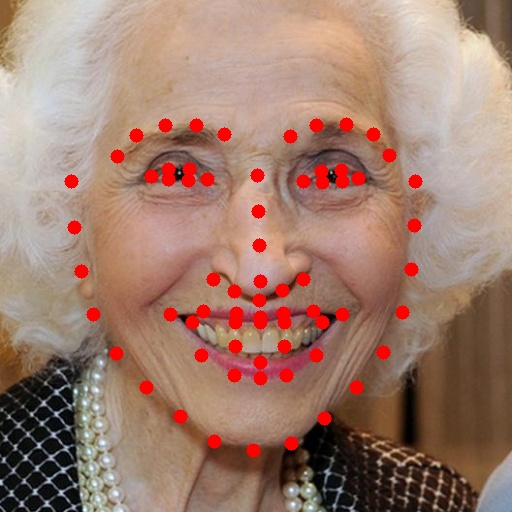} &
  \includegraphics[width=0.125\linewidth]{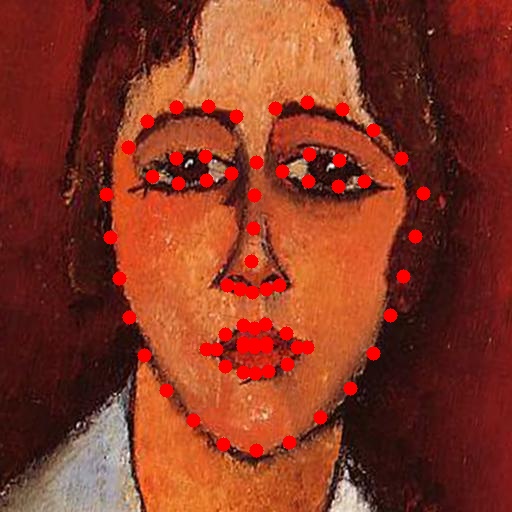} &
  \includegraphics[width=0.125\linewidth]{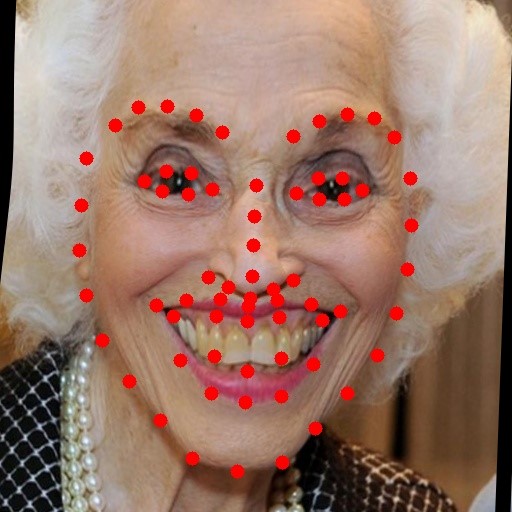} &
  \includegraphics[width=0.125\linewidth]{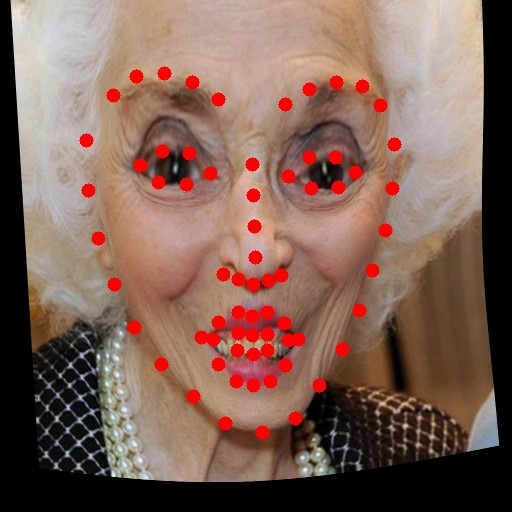} &
  \includegraphics[width=0.125\linewidth]{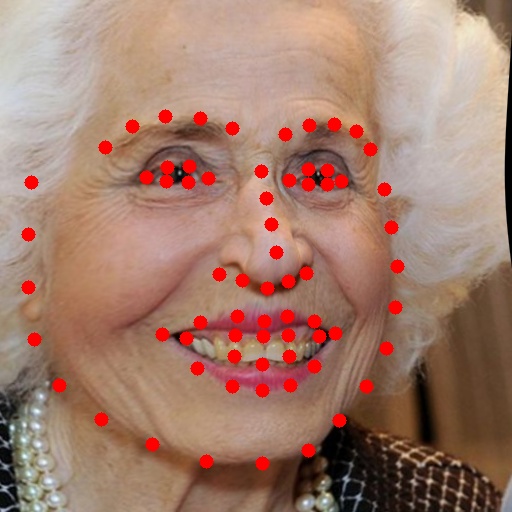} &
  \includegraphics[width=0.125\linewidth]{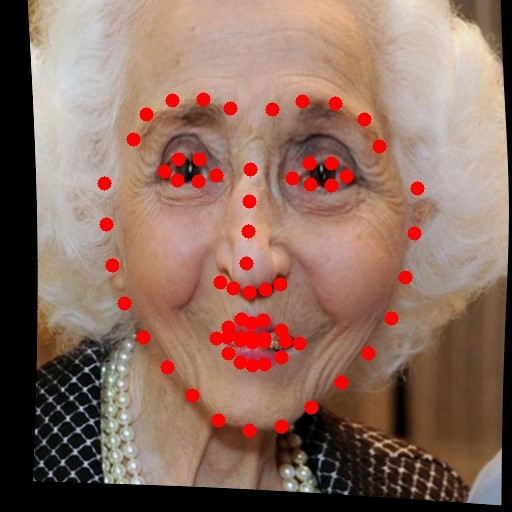} &
  \includegraphics[width=0.125\linewidth]{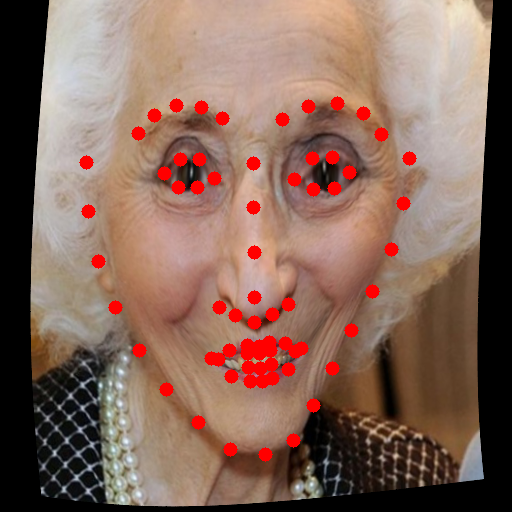} \\
   &
  \includegraphics[width=0.125\linewidth]{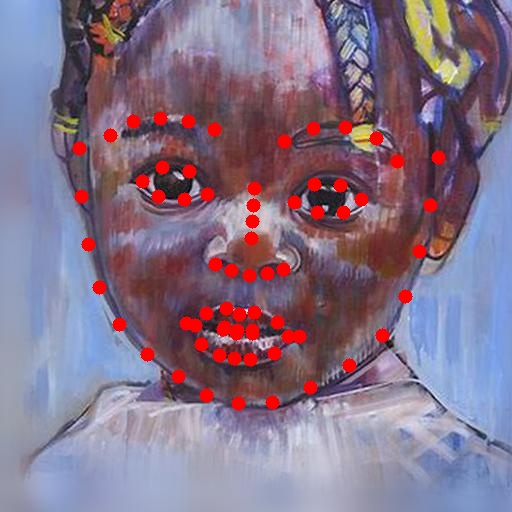} &
  \includegraphics[width=0.125\linewidth]{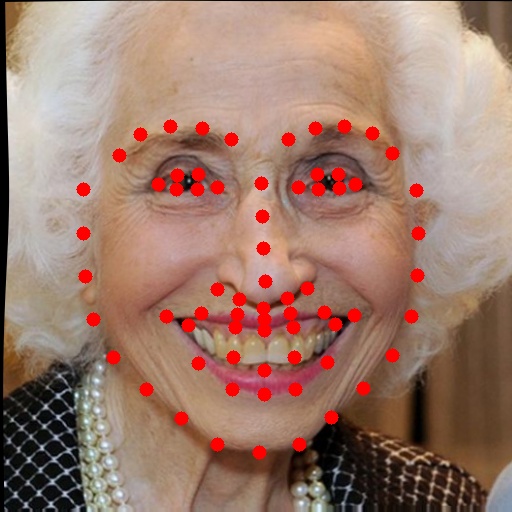} &
  \includegraphics[width=0.125\linewidth]{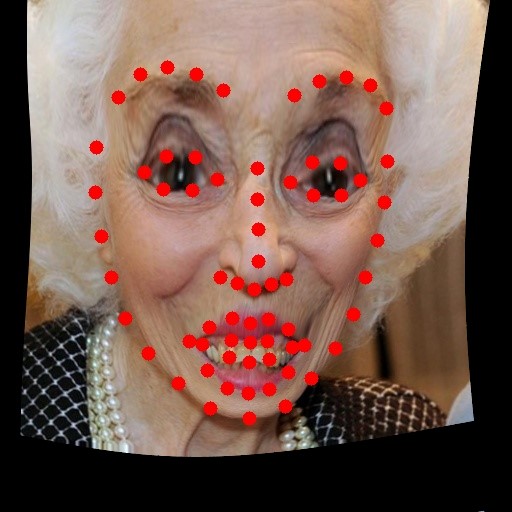} &
  \includegraphics[width=0.125\linewidth]{figures/ablation_landmark/example2/cyc.jpg} &
  \includegraphics[width=0.125\linewidth]{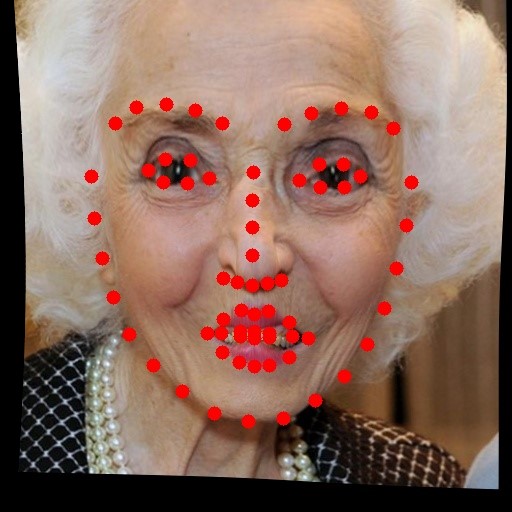} &
  \includegraphics[width=0.125\linewidth]{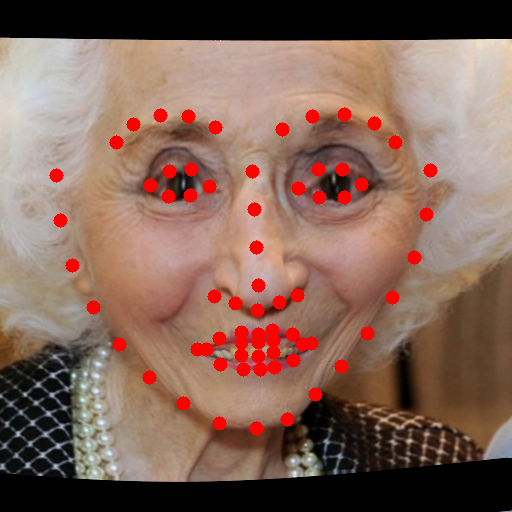} \\
  Content & Style & MUNIT & DRIT++ & CycleGAN & Ours w/o c-loss & Ours \\
  \end{tabular}}
\caption{\textbf{Comparisons on landmark translation.} Content image is warped based on the different landmark translation results. From left to right: MUNIT~\cite{huang2018multimodal}, DRIT~\cite{lee2019drit}, CycleGAN~\cite{zhu2017unpaired}, ours without classification loss, and ours.}
	\label{fig:cmp_geometry}
\end{figure*}

\begin{figure*}[t]
    \centering
    \includegraphics[width=1.0\linewidth]{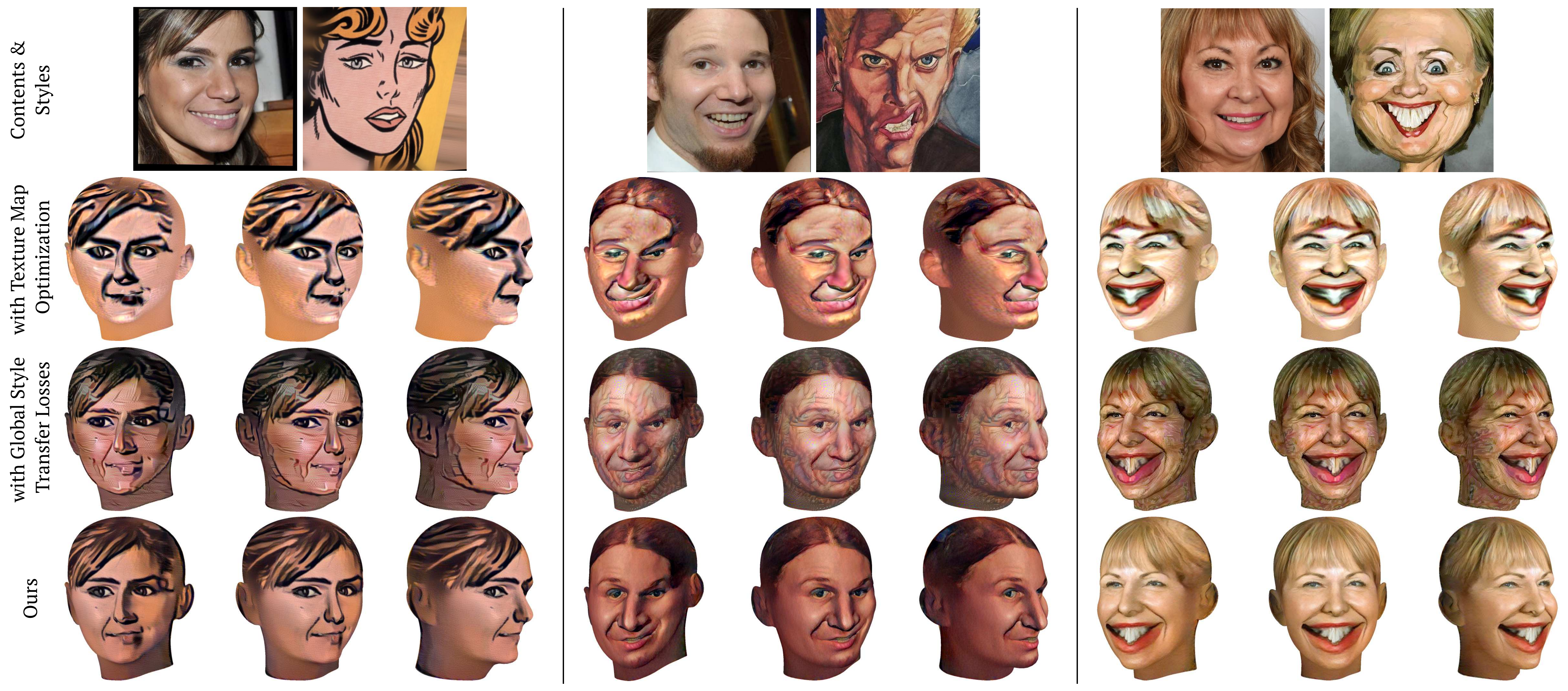}
 \caption{\textbf{Comparisons on texture stylization.} We compare our results using multi-view optimization and local style transfer losses with the results using single texture map optimization and the results using global style transfer losses.}
 \label{fig:cmp_texture}
\end{figure*}

\begin{figure*}[t]
    \centering
    \includegraphics[width=1.02\linewidth]{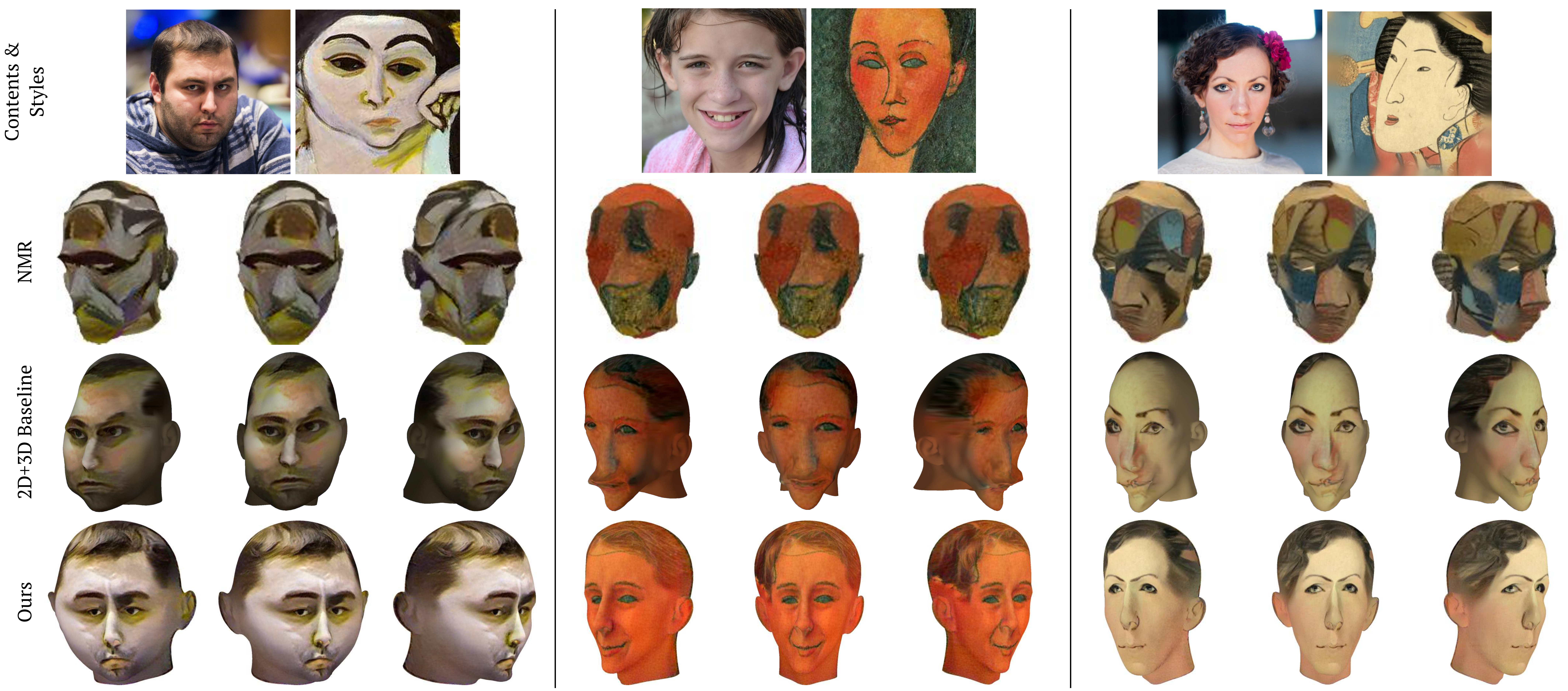}
 \caption{\textbf{Comparison with 3D geometric stylization methods.} From top to bottom: Neural 3D Mesh Renderer (NMR) \cite{kato2018neural}, the 2D+3D baseline (\cite{yaniv2019face} to stylize both geometry and texture in the image plane, and then FacewareHouse \cite{cao2014facewarehouse} to reconstruct the face), and ours.}
 \label{fig:cmp_res}
\end{figure*}

\section{Experiments}
In this section, we validate the effectiveness of key designs in our method through ablation studies.

\subsection{Two Stages v.s. Single stage}

To show that the two-stage framework is effective in learning both geometry and texture styles, we design an ablation study to compare it with the single-stage framework, in which we do not have a separate geometry style transfer stage and let the geometry style be learned jointly with the texture styles. That means during the multi-view optimization, the style transfer objectives will be back-propagated to variables $v_z$ and $t_z$ via the differentiable renderer and update them. However, the freedom in updating two variables will also cause some ambiguities: whether the vertex should be moved or the texture map's color should be changed to simulate the reference style. Thus, it leads to obvious artifacts on both geometry and texture, as shown in Fig. \ref{fig:cmp_two_one}.

\subsection{Geometry style Transfer}

\begin{table}[h]
\resizebox{0.5\textwidth}{!}{
\begin{tabular}{@{}cccccc@{}}
\toprule
methods & CycleGAN & MUNIT & DRIT++ & Ours w/o c-loss & Ours \\ \midrule
FIDs    & 49.432 & 43.828 & 32.427 & 30.802  & 23.587\\
\bottomrule
\end{tabular}}
\caption{\textbf{FID comparison} for landmark-to-landmark translation with different network architectures.}
\label{tab:fid}
\end{table}

In the first stage of our framework, the landmark-to-landmark translation network is the key to learn the geometry style from the reference and guide the 3D model deformation. Since the landmark-to-landmark translation task is analogical to the image-to-image one, some standard image-to-image network architectures like CycleGAN \cite{zhu2017unpaired}, MUNIT \cite{huang2018multimodal}, and DRIT++ \cite{lee2019drit} can also be migrated to landmarks, by simply replacing its Convolution and ReLU layers with fully connected layers to incorporate the point vector input. CariGAN \cite{cao2018carigans} has already proved that migrating the CycleGAN from image to landmarks in this way is successful. For the comparison purpose, we train the migrated CycleGAN, MUNIT, and DRIT++ on the same training dataset as our landmark-to-landmark translation network, and test these three models and ours on 500 samples in the test set. FID metric is adopted to evaluate the authenticity of generated landmarks. Unlike images where the FID is calculated between perceptual features, here it is calculated between 32-dims PCA coefficients to measure the distribution similarity between generated landmarks and the ground truth dataset of artwork landmarks. The comparison results are shown in Table. \ref{tab:fid}. Our multi-modal landmark-to-landmark translation network achieves the best score. The visual results shown in Fig. \ref{fig:cmp_geometry} also reflect this observation. The results of CycleGAN lack diversity because it is a single-modal network. For a given input, it can only generate a single result that cannot be controlled by the reference style. MUNIT and DRIT++ are both multi-modal. However, their results are not faithful enough to the geometry styles of the reference because their implicit disentanglement scheme is hard to correctly decouple content and style components. Thanks to the explicit disentanglement in our network, the content and style components are clearly defined in our network. Thus our results can better preserve the input identity, and at the same time, more faithfully transfer the style from the reference. Moreover, MUNIT and DRIT++ have modal collapse problems and are insensitive to small shape differences between different references, generating very similar results for them, as shown by the three styles in the second example. In contrast, our method can reflect even these small differences in styles.

Next, we ablate the function of the classification loss (Eq. \ref{eq:class}) in our multimodal landmark-to-landmark translation network by training another version without it. The FID score of the model without the classification loss increases which indicates its results are less similar to real data distribution. That is because the classification loss will encourage the generated landmarks to better captures the common characteristics of all samples in the reference’s class. This point can be more clearly observed by visual results in Ours w/o c-loss column in Fig. \ref{fig:cmp_geometry}.

\subsection{Ablation on Texture Style Transfer}

Once the geometry style is transferred in the first stage, we have two options to transfer the texture style. The first one is to apply image style transfer on the texture map, and the second one is to optimize the texture map via the differential render in the multi-view framework. We compare the results of these two options in Fig. \ref{fig:cmp_texture}. With the first option, the texture has lots of artifacts because the texture map, which is often a severely distorted face, will cause misalignment and texture pattern distortions in rendered results. The second option, our choice, can avoid these problems and achieve much better texture quality by rendering the 3D model into normal face images for style transfer. And the multi-view optimization framework can seamlessly combine the multi-view transfer results into a full texture map. 

Next, we ablate the choice for texture style transfer loss. As claimed, the texture transfer loss is inherited from neural image style transfer methods, which have two major categories. One is to match the global statistics of deep features, while the other is to build pixel-wise feature correspondences and transfer the local statistics. The first category's representative method is Gatys \cite{gatys2015neural} and the STROTSS \cite{kolkin2019style} represents the state of the art of the second category. We apply the loss functions from these two methods, respectively, in our multi-view optimization and show the comparison results in Fig. \ref{fig:cmp_texture}. Since our style transfer focuses on faces with strong semantic correspondences, the STROTSS loss, our choice, leads to more visually pleasing results, where the style of a local part can be precisely transferred to its corresponding part.

\section{Results}
In this section, we first compare our landmark translation method and the whole framework with existing baselines and then conduct a user study to evaluate our performance in identity preservation and style faithfulness.

\subsection{Comparisons}
The 3D neural style transfer is still an underexplored problem. To the best of our knowledge, only three works have attempted to do it with their differentiable rendering tools: Neural 3D Mesh Renderer (NMR) \cite{kato2018neural}, Paparazzi \cite{liu2018paparazzi}, and Differentiable Image Parameterizations (DIP) \cite{mordvintsev2018differentiable}, as shown in Fig. \ref{fig:cmp_res}. Paparazzi focuses on geometry style transfer only, while DIP focuses on texture style transfer only. These two are different from our goal, so we only take NMR, which jointly optimizes geometry and texture, for comparison. Another naive baseline to achieve 3D style transfer is to combine the existing deformable image style transfer method with a 3D face reconstruction method. Specifically, we first run Face of Art \cite{yaniv2019face} to stylize both geometry and texture in the image plane, and then use FacewareHouse \cite{cao2014facewarehouse} to reconstruct the 3D face model from the stylized image. We compare our method with NMR and the 2D+3D baseline in Fig. \ref{fig:cmp_res}. NMR is a style transfer method for general 3D objects. When it is applied to a face model, although it deforms the shape and updates the texture to reflect the style of the given reference, it fails to preserve the facial structure. That is because this method does not consider the face semantics. Once the face semantics are taken into account, the 3D style transfer task becomes much complicated, and it is hard to be solved with a single-stage framework like NMR. The 2D+3D baseline can correctly transfer the geometry style and texture style to an image. Still, the 3D face model generated from the image result cannot fit the 2D geometry well because the face reconstruction algorithm built on the real face basis cannot represent the exaggerated faces. Compared with these two methods, our method can generate much better results of stylized 3D faces.

\begin{figure}[t]
\setlength{\tabcolsep}{0.05mm}{
  \begin{tabular}{cccc}
      \includegraphics[width=0.5\linewidth]{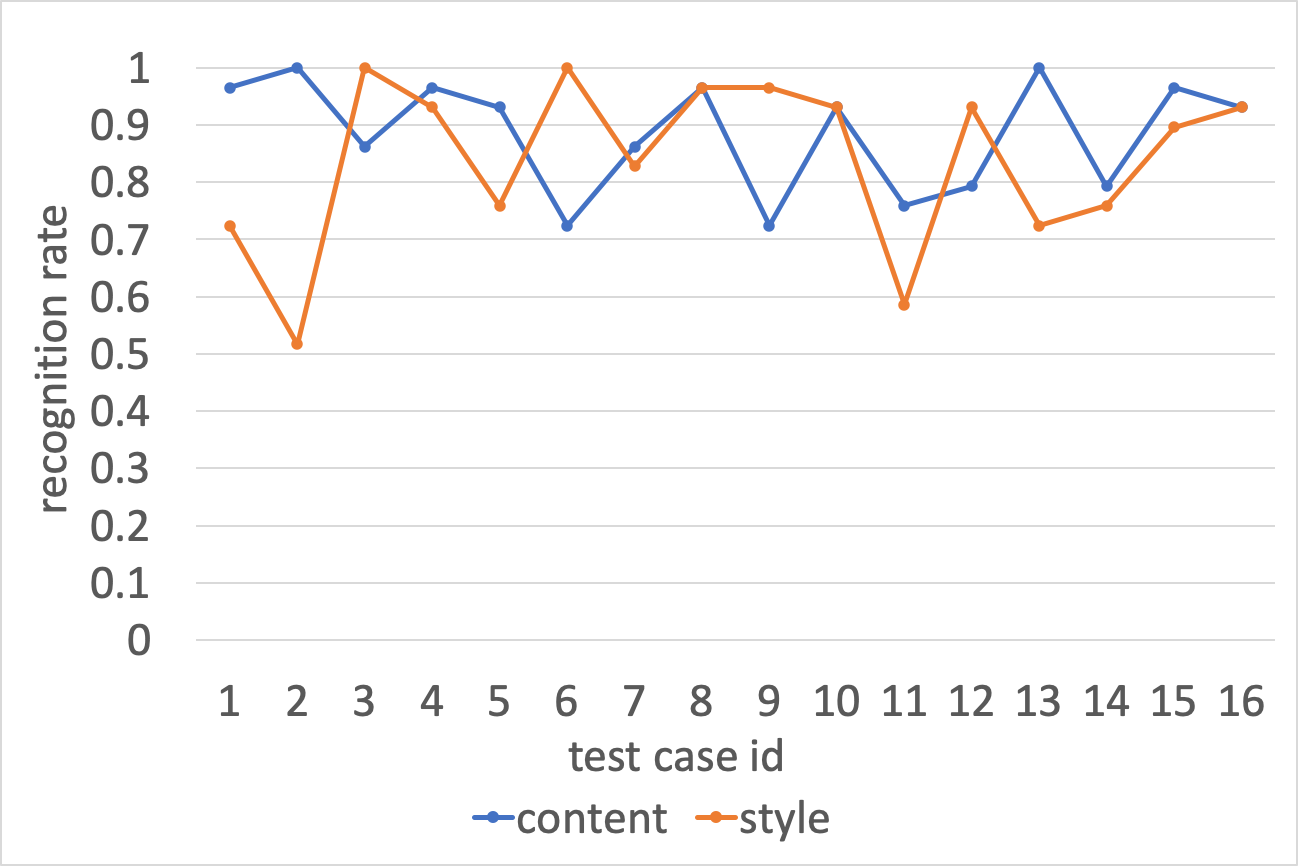} &
      \includegraphics[width=0.5\linewidth]{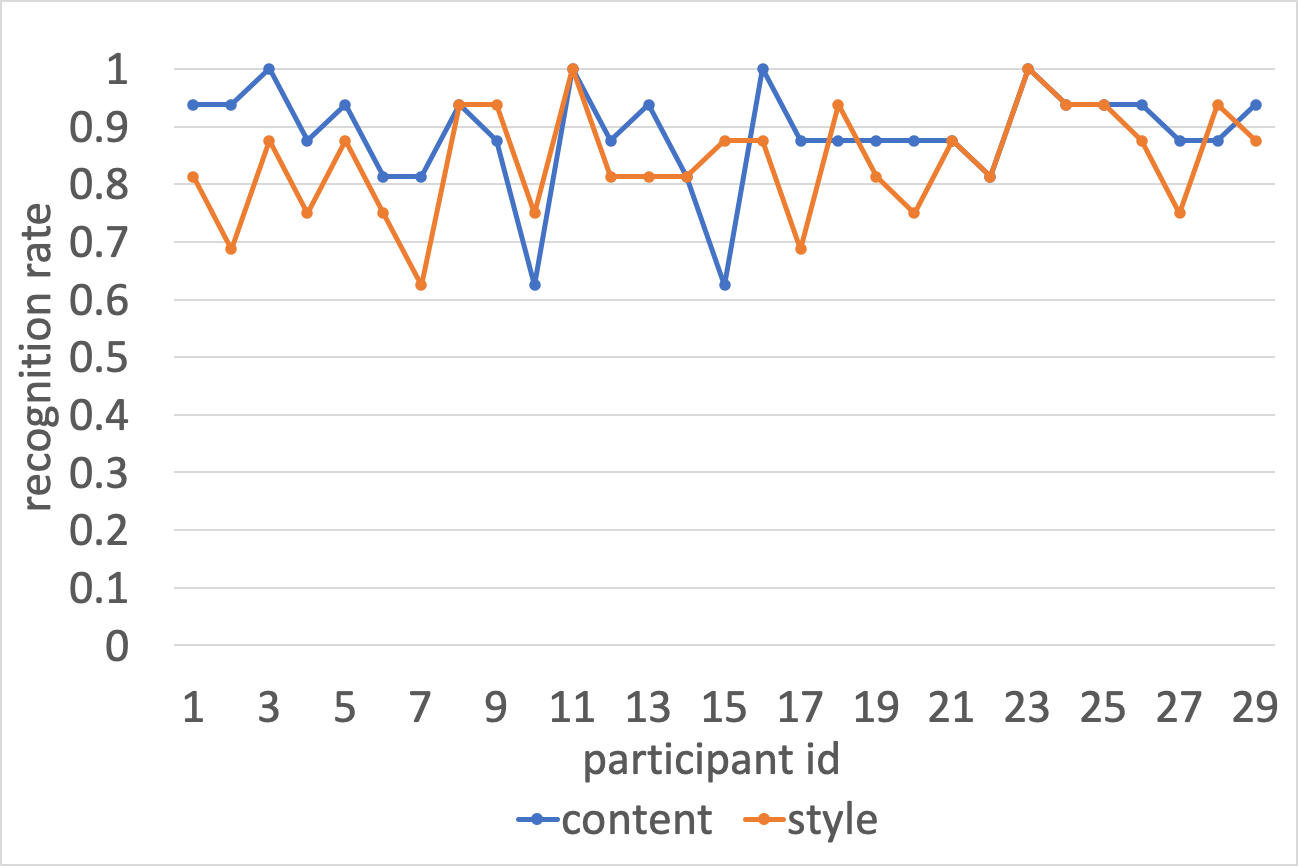} \\
      (a) Rec. rate per test case & (b) Rec. rate per participant \\
  \end{tabular}
  }
 \caption{\textbf{Perceptual study result analysis.} }
 \label{fig:user_study}
\end{figure}

\begin{figure}[t]
\centering
 \includegraphics[width=1\linewidth]{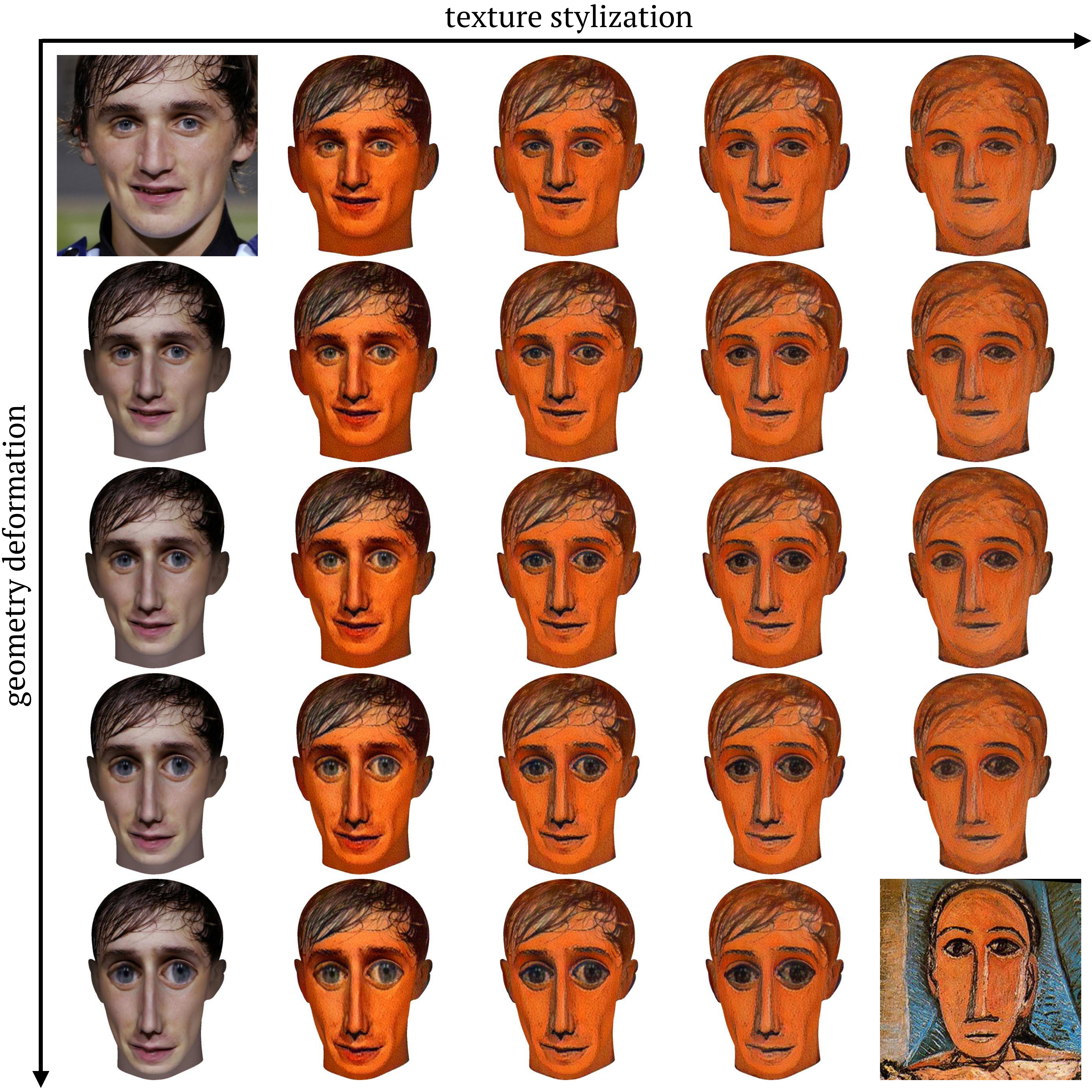}
 \caption{\textbf{An example of user control.} The scale of geometric deformation increases from top to bottom, while the style weight increases from left to right.}
 \label{fig:control}
\end{figure}

\subsection{Perceptual Study}
{Since there is no objective metric to evaluate the quality of style transfer, we conduct a perceptual study to evaluate our method in terms of identity recognizability and style faithfulness. In the study, 16 pairs of portrait photos and portrait art images are randomly selected from our testing set and then fed to the proposed method to generate stylized 3D face models. Each model is then rendered in 3 views for demonstration. Meanwhile, the four most similar images to the content input and the reference style are retrieved respectively from the dataset by considering both the identity similarity using the VGG-Face feature and the appearance similarity using the color histogram. Participants are asked to select the correct content input and style reference from the five options. After collecting feedback from 29 users, the average successful recognition rates are 86.80\% for content and 83.71\% for style, which proves that our stylization results can well preserve both input face identity and reference style. The more detailed visualization of the perceptual study results can be seen in Fig. \ref{fig:user_study}, which shows that our high recognition rate is stable with different test cases and different users. 

In stylization problems, identity preservation is always somewhat contradictory to style faithfulness. To transfer patterns of the style to the content, it is inevitable to cause some loss of identity. Although our method can automatically achieve a good balance as evaluated in the perceptual study, it also has the flexibility to let the user adjust the balance by changing the scale of geometric deformation and the style loss weights in texture style transfer. An example of user-adjusted balance is shown in Fig. \ref{fig:control}. }

 \subsection{Runtime}
The average runtime of each step for one example is listed in Table~\ref{tab:time}. The runtime performance is tested on a workstation with a NVIDIA Tesla V100 GPU. The bottleneck of our current piepline is the multi-view optimization, which can be replaced with a faster feed-forward style transfer network in the future.

\begin{table}[ht]
\resizebox{0.48\textwidth}{!}{
\begin{tabular}{@{}cccc@{}}
\toprule
Step & Landmark Trans. & Face Deformation & Texture Trans.  \\ \midrule
Secs    & 0.2  & 2 & 255  \\
\bottomrule
\end{tabular}}
\caption{\textbf{The average processing time} of each main step per-example.}
\label{tab:time}
\end{table}

\section{Applications}
The 3D stylized face model generated by our method can be applied to many 2D and 3D graphics applications, such as image stylization, personalized emoji or avatar generation, 3D character modeling and animation, 3D printing, etc. In this section, we show four typical applications including both 2D and 3D scenarios.

\begin{figure}[t]
\centering
 \setlength{\tabcolsep}{0.2mm}{
 \begin{tabular}{ccccc}
   &
  \includegraphics[width=0.2\linewidth]{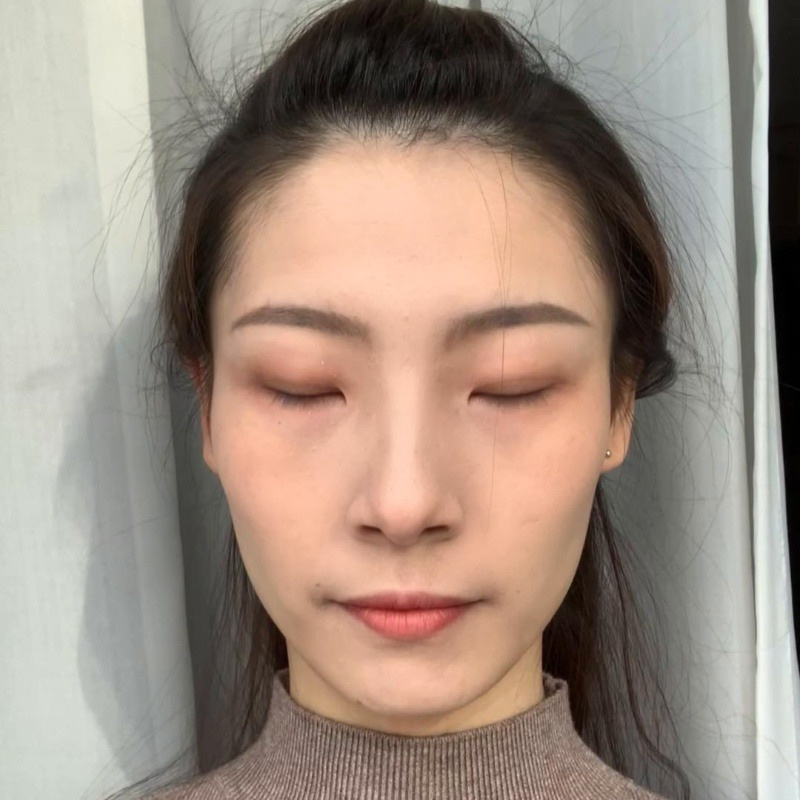} &
  \includegraphics[width=0.2\linewidth]{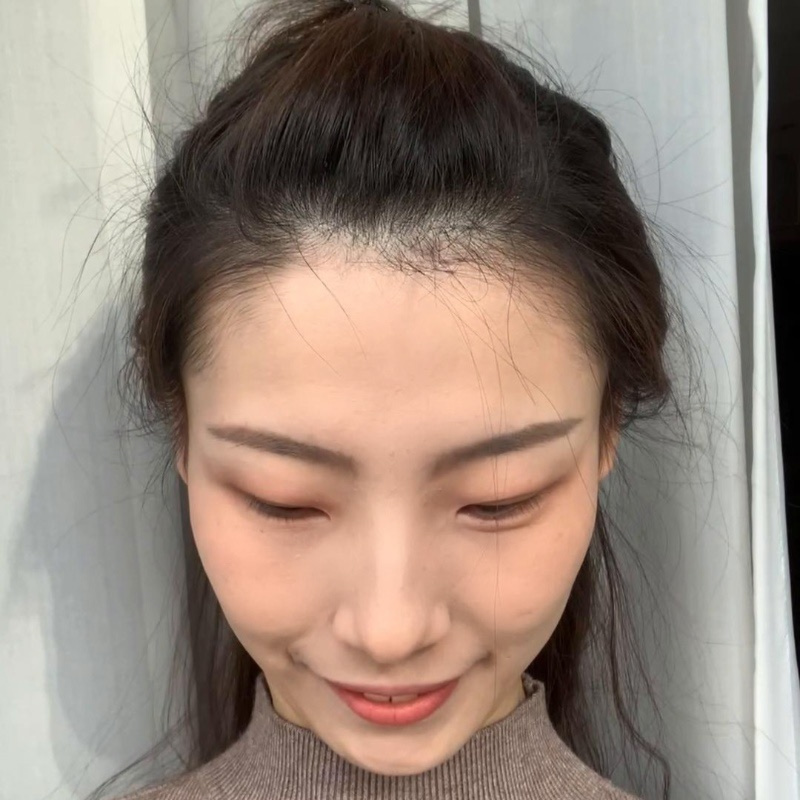} &
  \includegraphics[width=0.2\linewidth]{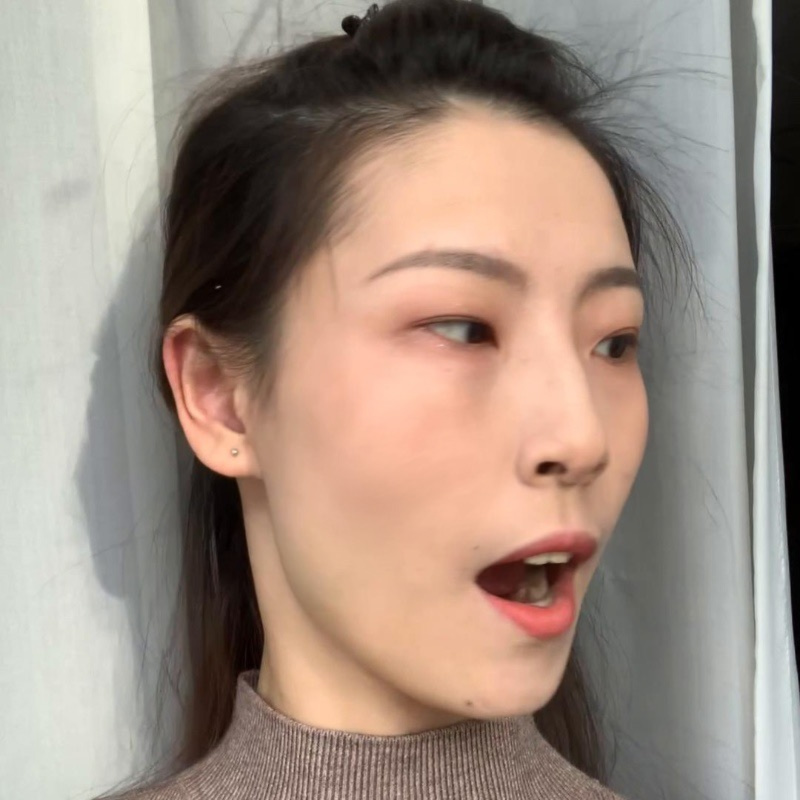} &
  \includegraphics[width=0.2\linewidth]{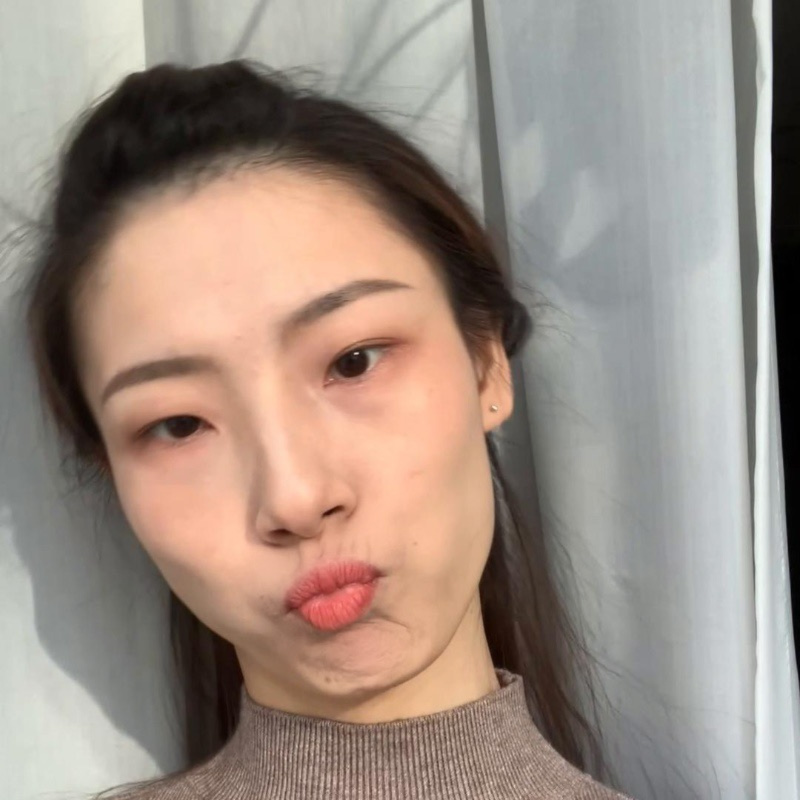}\\
  \includegraphics[width=0.2\linewidth]{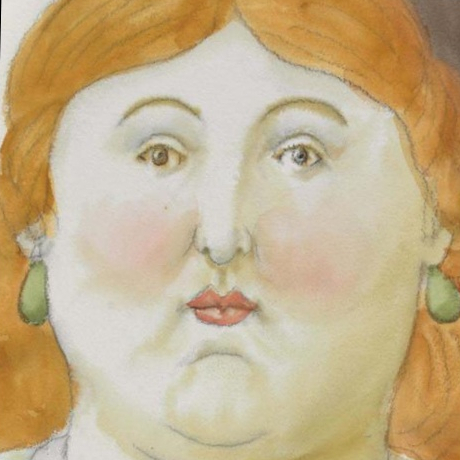} &
  \includegraphics[width=0.2\linewidth]{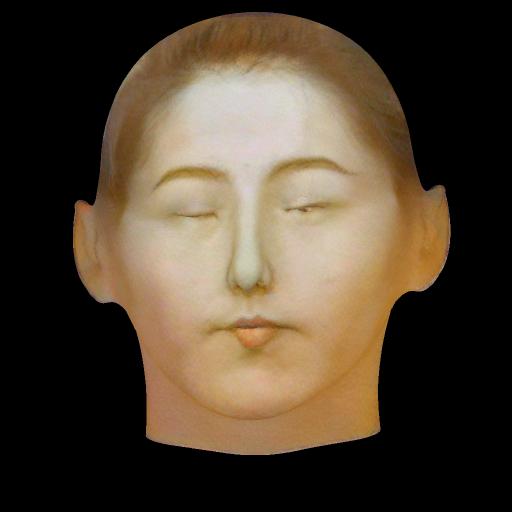} &
  \includegraphics[width=0.2\linewidth]{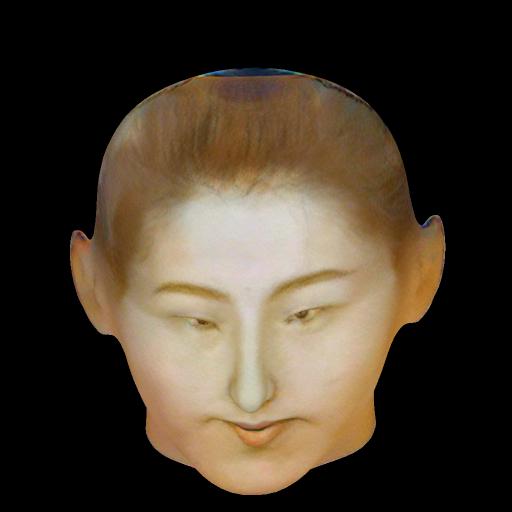} &
  \includegraphics[width=0.2\linewidth]{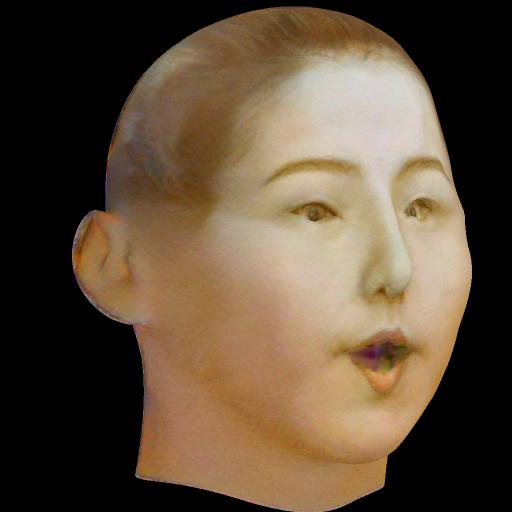} &
  \includegraphics[width=0.2\linewidth]{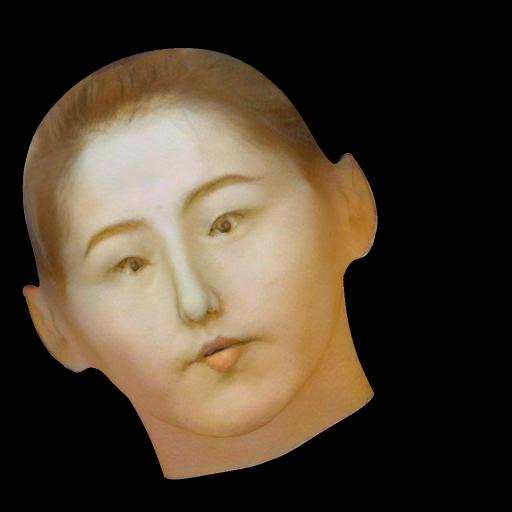}\\
  \includegraphics[width=0.2\linewidth]{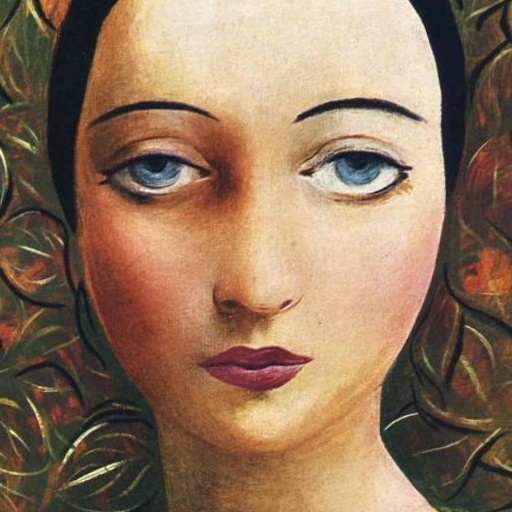} &
  \includegraphics[width=0.2\linewidth]{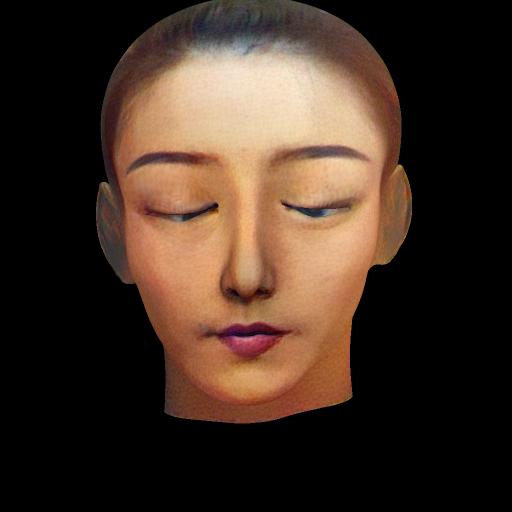} &
  \includegraphics[width=0.2\linewidth]{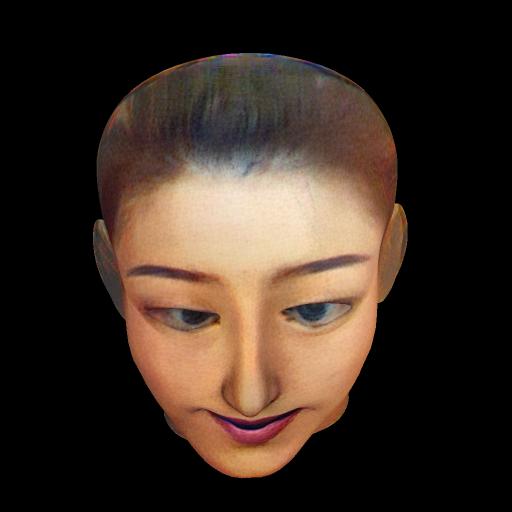} &
  \includegraphics[width=0.2\linewidth]{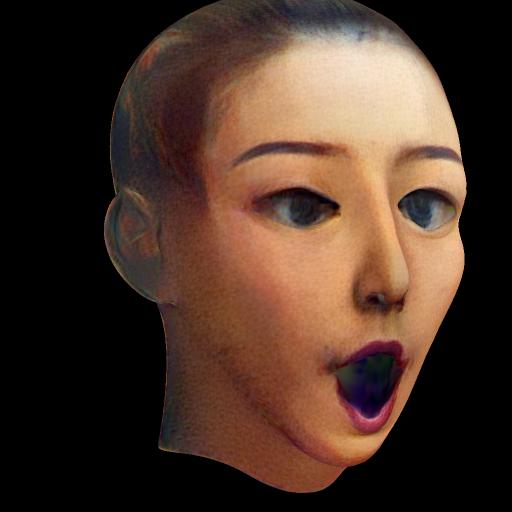} &
  \includegraphics[width=0.2\linewidth]{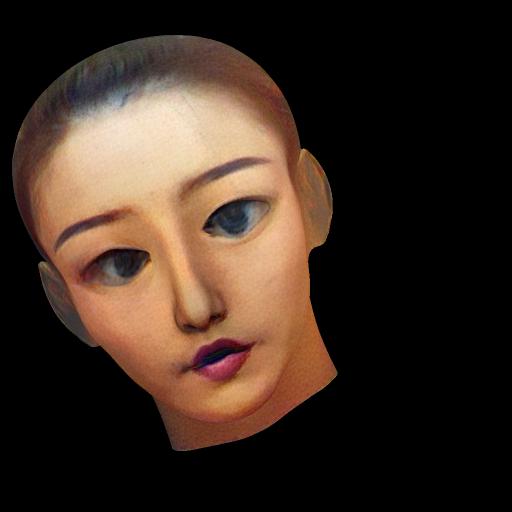}\\
  \includegraphics[width=0.2\linewidth]{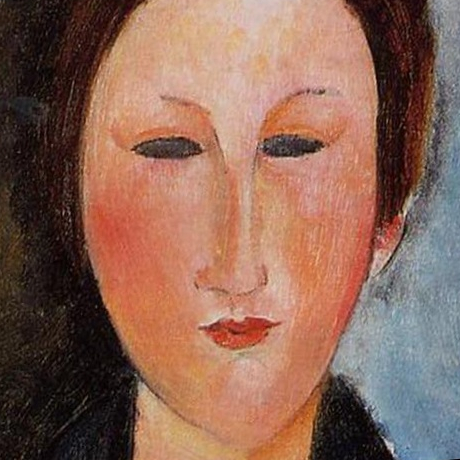} &
  \includegraphics[width=0.2\linewidth]{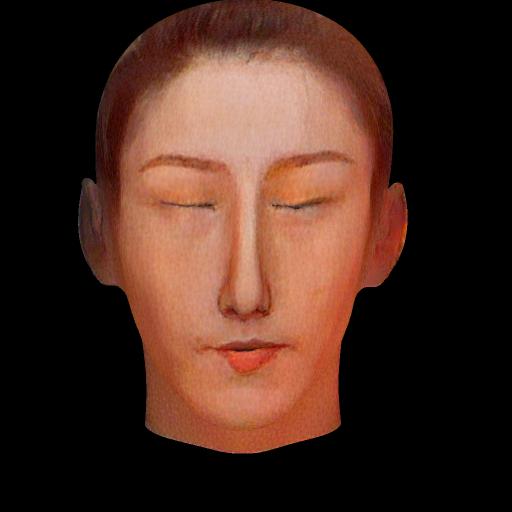} &
  \includegraphics[width=0.2\linewidth]{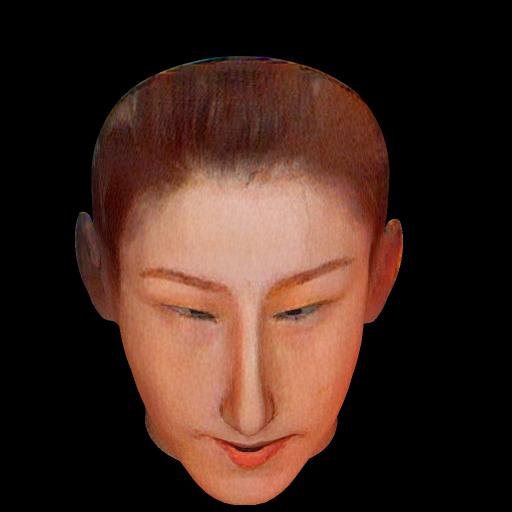} &
  \includegraphics[width=0.2\linewidth]{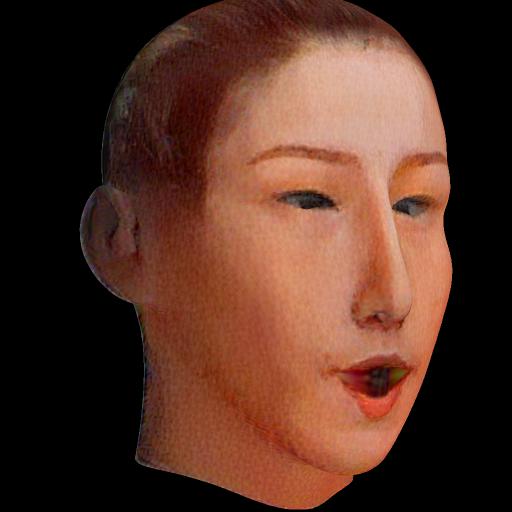} &
  \includegraphics[width=0.2\linewidth]{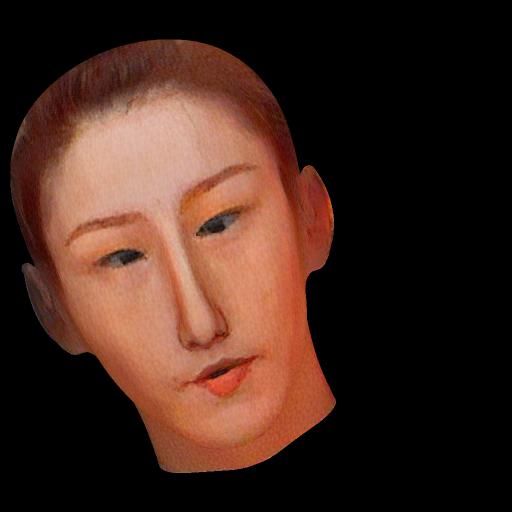}\\
  Style & \multicolumn{4}{c}{Original Frames and Reenactment Results} \\
  \end{tabular}}
 \caption{\textbf{Application of stylized portrait reenactment.} We apply the deformation difference and texture transfer result on each frame of a video to generate stylized video sequence with corresponding expressions.}
 \label{fig:app_video}
\end{figure}

\begin{figure}[t]
\centering
 \setlength{\tabcolsep}{0.2mm}{
 \begin{tabular}{ccccc}
  \includegraphics[width=0.2\linewidth]{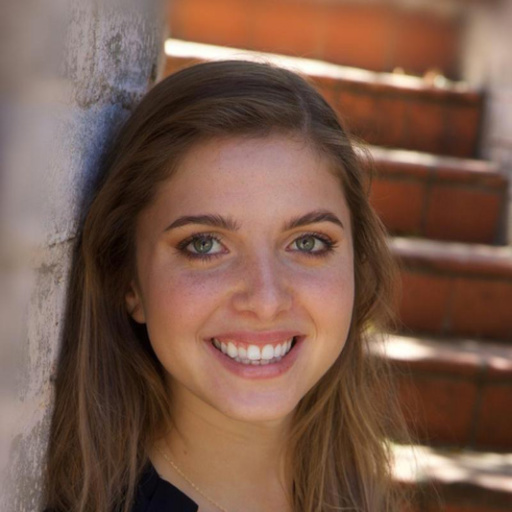} &
  \includegraphics[width=0.2\linewidth]{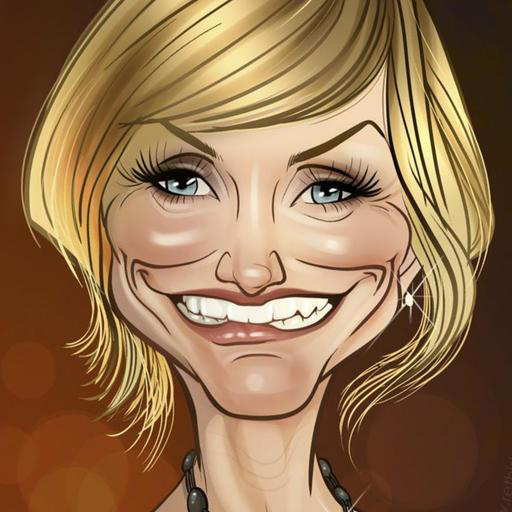} &
  \includegraphics[width=0.2\linewidth]{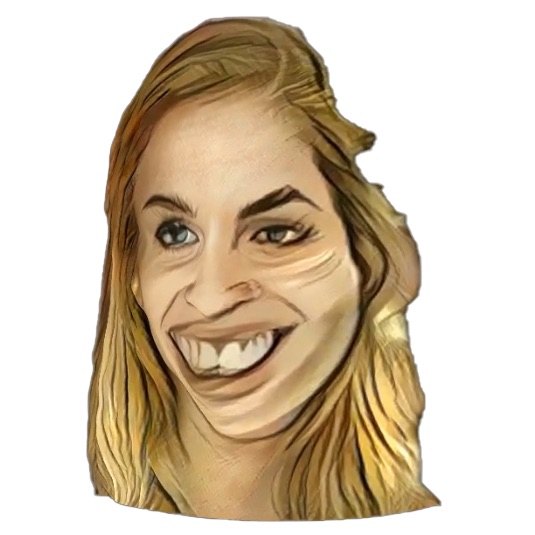} &
  \includegraphics[width=0.2\linewidth]{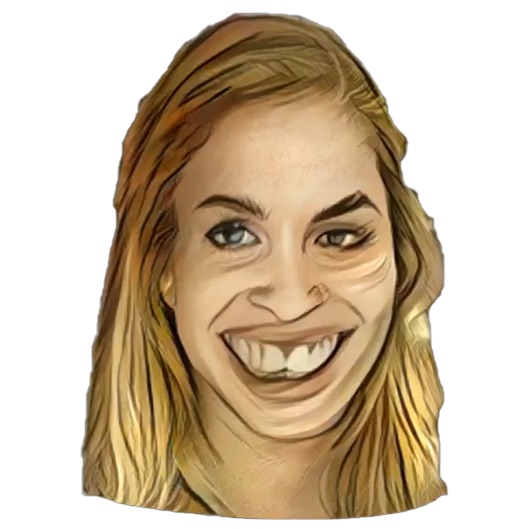} &
  \includegraphics[width=0.2\linewidth]{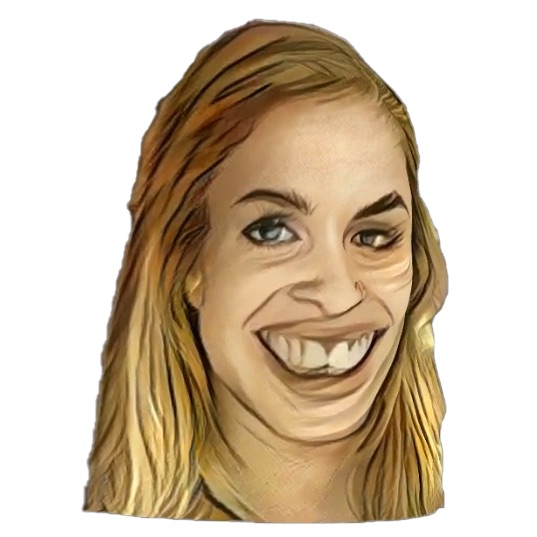}\\
  &
  \includegraphics[width=0.2\linewidth]{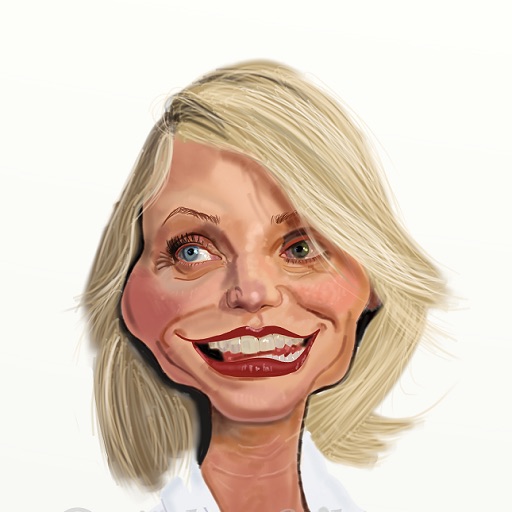} &
  \includegraphics[width=0.2\linewidth]{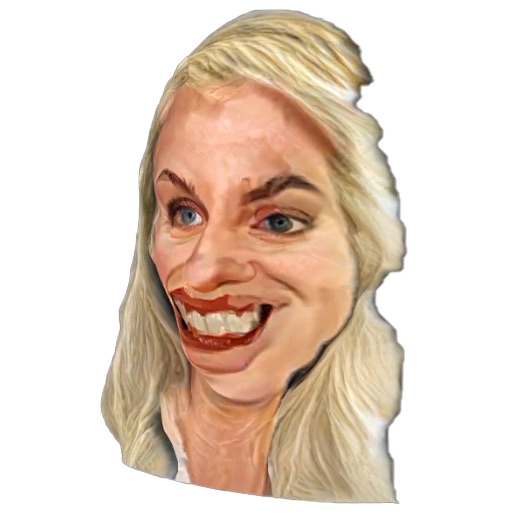} &
  \includegraphics[width=0.2\linewidth]{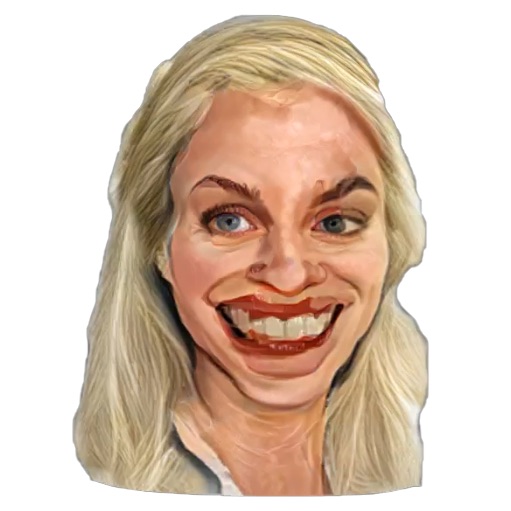} &
  \includegraphics[width=0.2\linewidth]{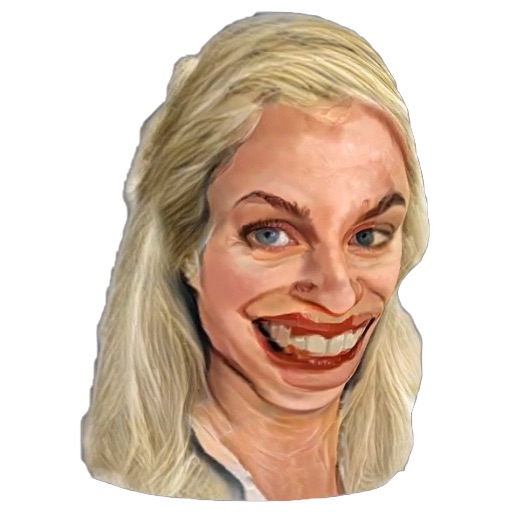}\\\hline
  & \\[\dimexpr-\normalbaselineskip+3pt]
  \includegraphics[width=0.2\linewidth]{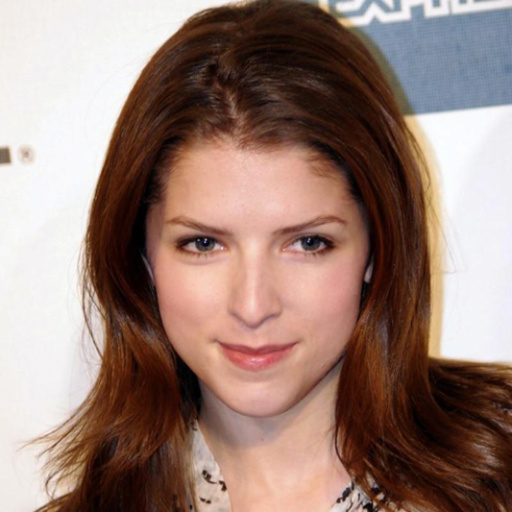} &
  \includegraphics[width=0.2\linewidth]{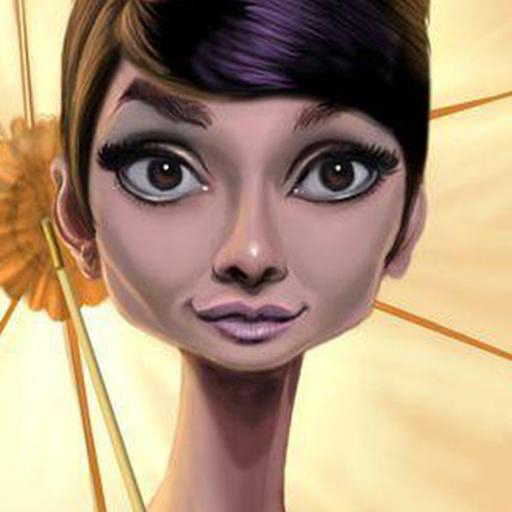} &
  \includegraphics[width=0.2\linewidth]{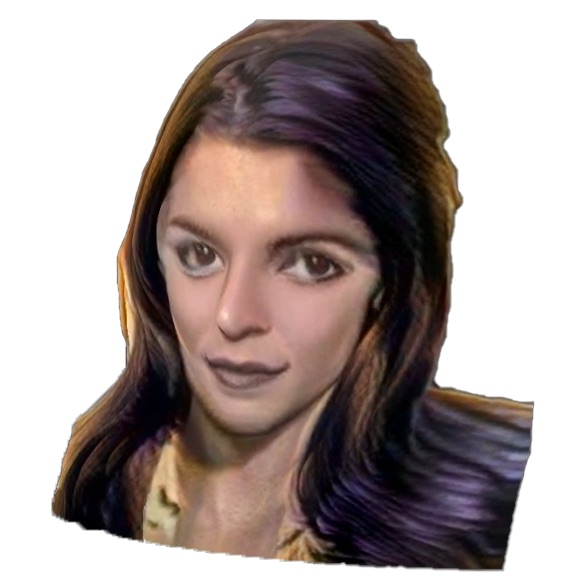} &
  \includegraphics[width=0.2\linewidth]{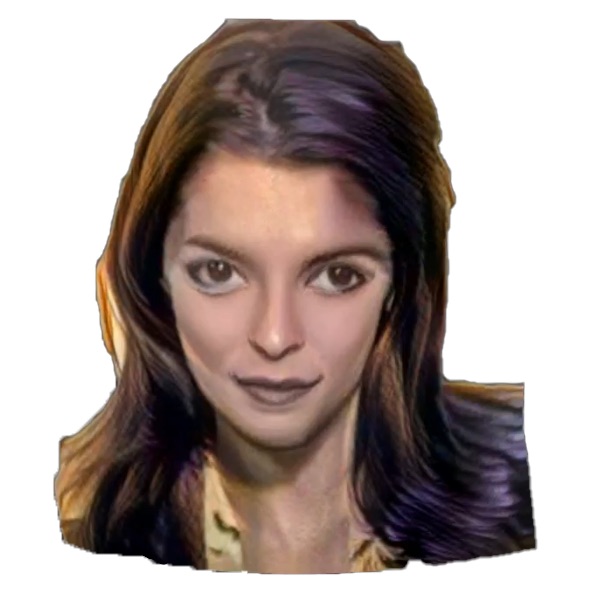} &
  \includegraphics[width=0.2\linewidth]{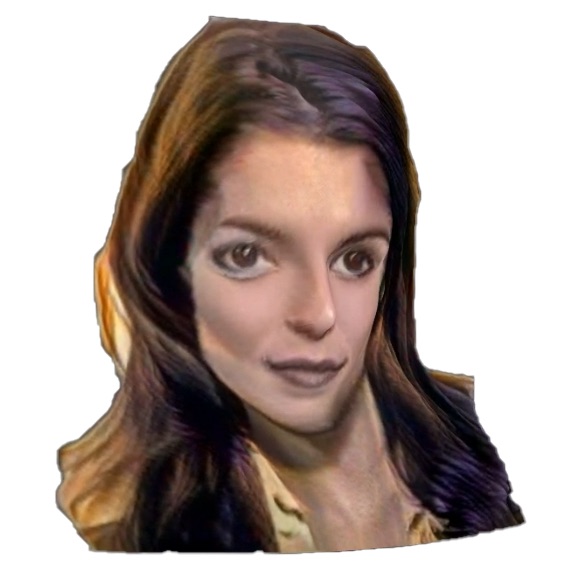}\\
  &
  \includegraphics[width=0.2\linewidth]{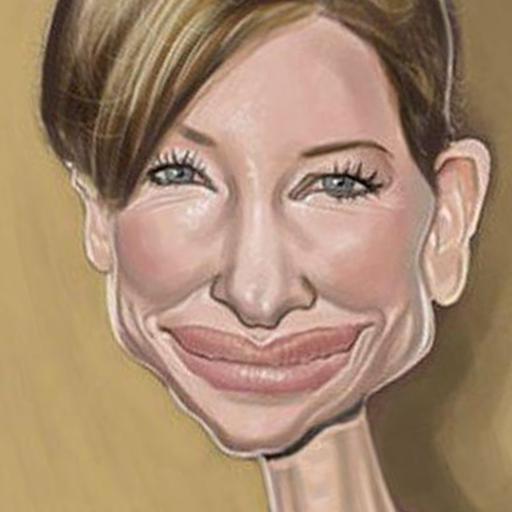} &
  \includegraphics[width=0.2\linewidth]{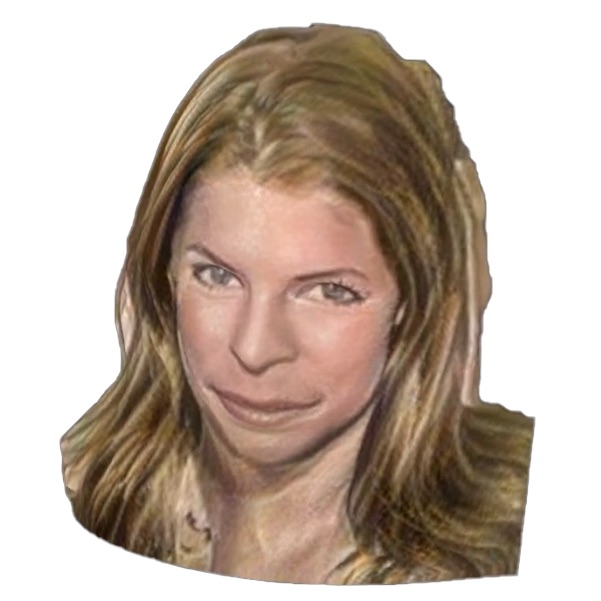} &
  \includegraphics[width=0.2\linewidth]{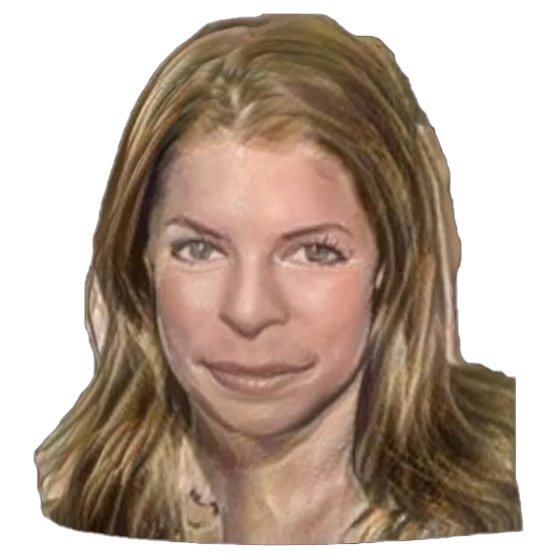} &
  \includegraphics[width=0.2\linewidth]{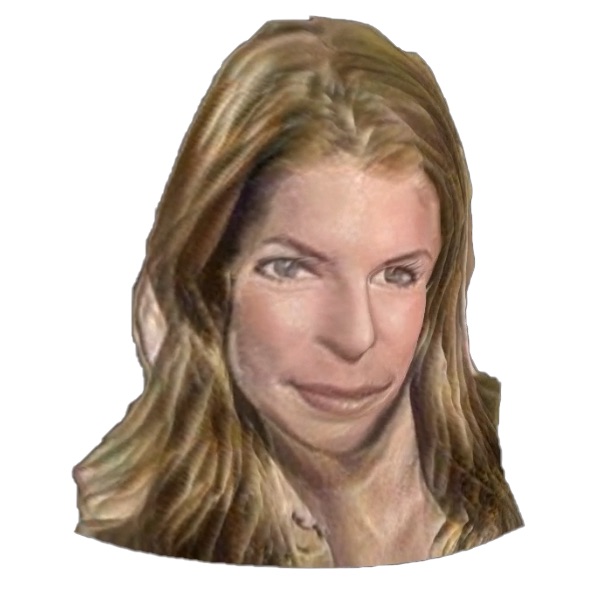}\\
  Content & Style & \multicolumn{3}{c}{Multi-View Results} \\
  \end{tabular}}
 \caption{\textbf{Application of cartoon 3D portrait modeling.} We combine the result of full-portrait 3D modeling method~\cite{chai2015high} with our face model to get a complete stylized portrait model.}
 \label{fig:app_3d_style}
\end{figure}

\begin{figure}[t]
\centering
 \setlength{\tabcolsep}{0.2mm}{
 \begin{tabular}{ccccc}
  \includegraphics[width=0.2\linewidth]{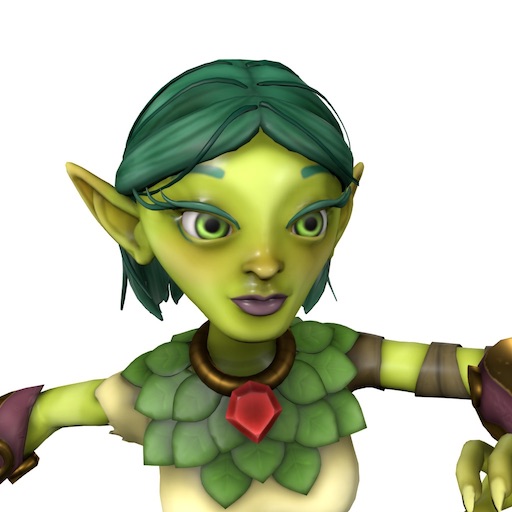} &
  \includegraphics[width=0.2\linewidth]{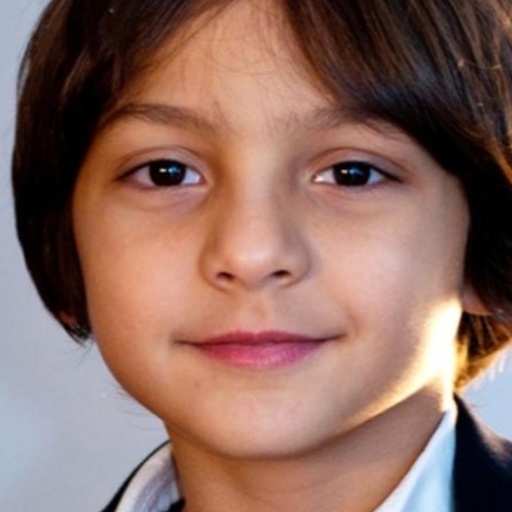} &
  \includegraphics[width=0.2\linewidth]{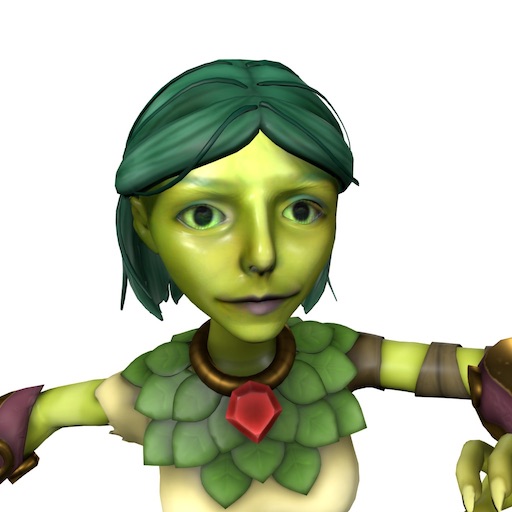} &
  \includegraphics[width=0.2\linewidth]{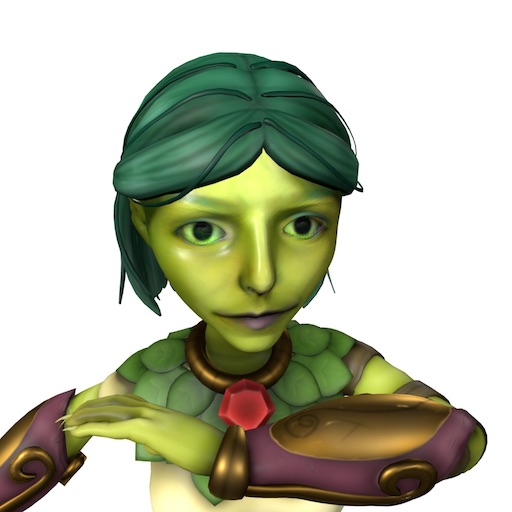} &
  \includegraphics[width=0.2\linewidth]{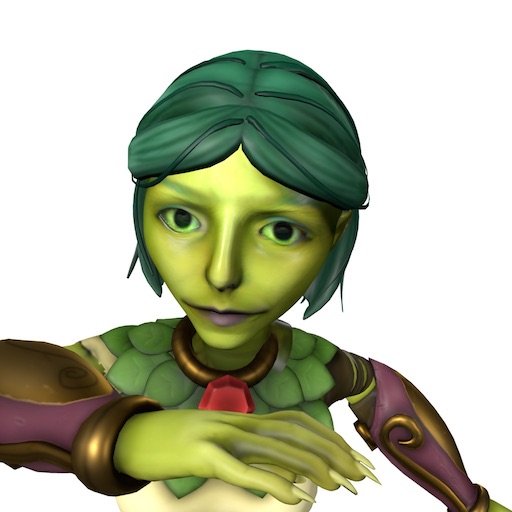}\\
  &
  \includegraphics[width=0.2\linewidth]{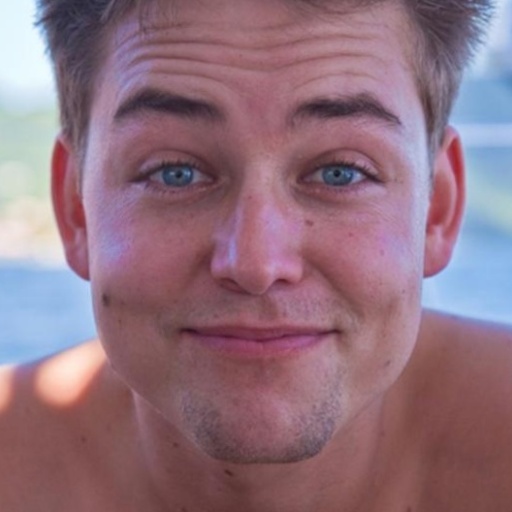} &
  \includegraphics[width=0.2\linewidth]{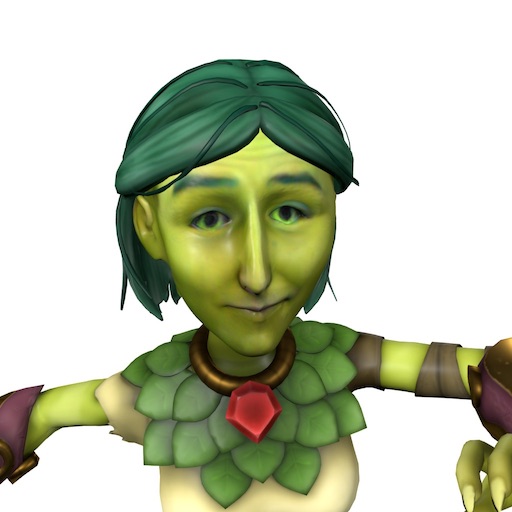} &
  \includegraphics[width=0.2\linewidth]{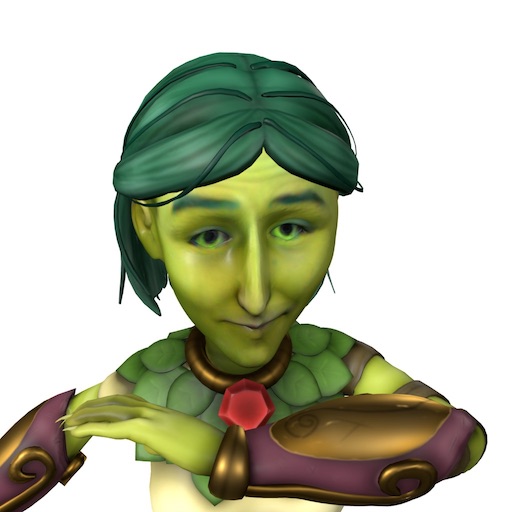} &
  \includegraphics[width=0.2\linewidth]{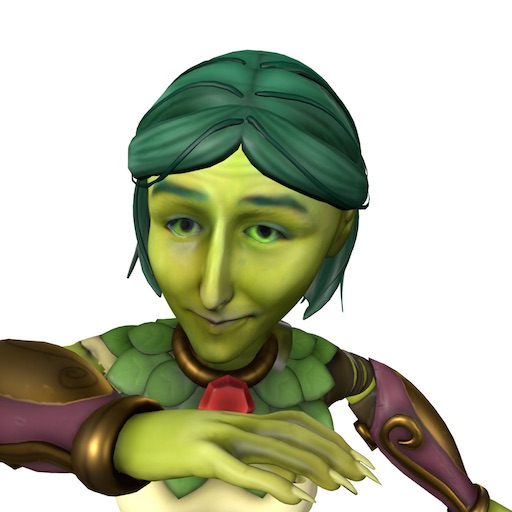}\\\hline
  & \\[\dimexpr-\normalbaselineskip+3pt]
  \includegraphics[width=0.2\linewidth]{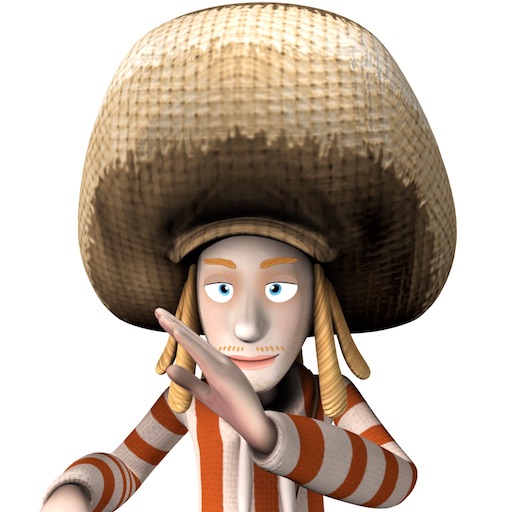} &
  \includegraphics[width=0.2\linewidth]{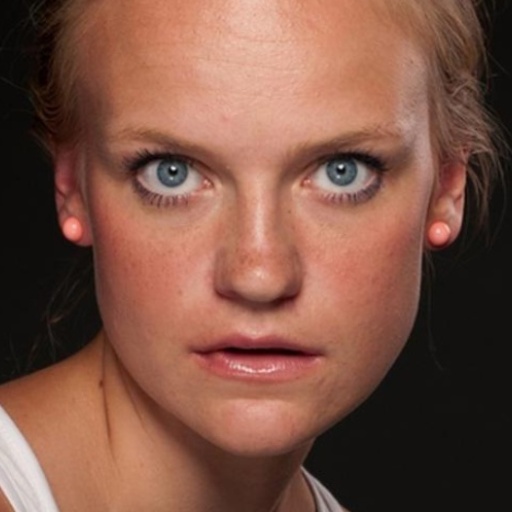} &
  \includegraphics[width=0.2\linewidth]{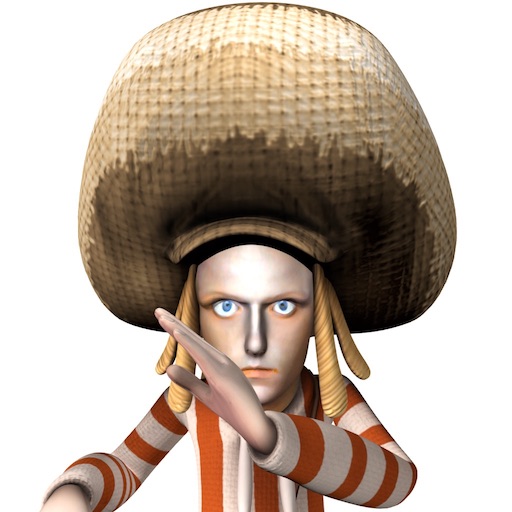} &
  \includegraphics[width=0.2\linewidth]{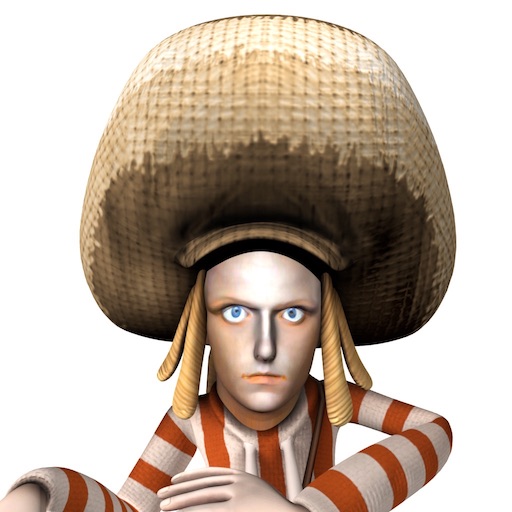} &
  \includegraphics[width=0.2\linewidth]{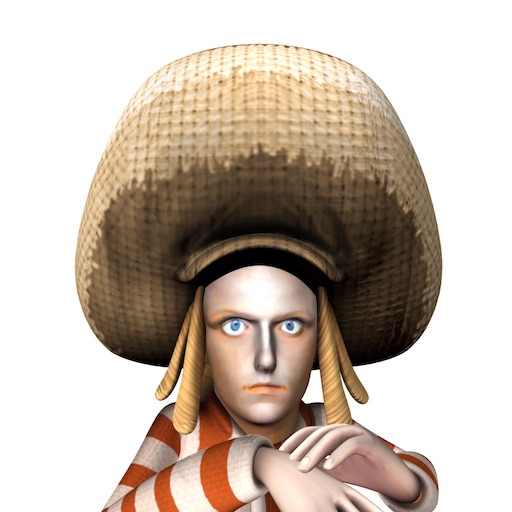}\\
  &
  \includegraphics[width=0.2\linewidth]{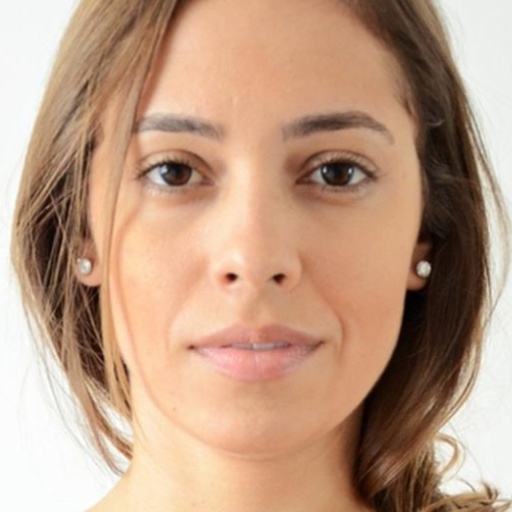} &
  \includegraphics[width=0.2\linewidth]{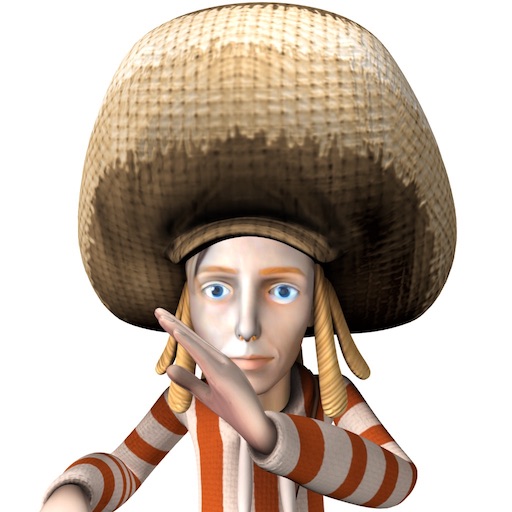} &
  \includegraphics[width=0.2\linewidth]{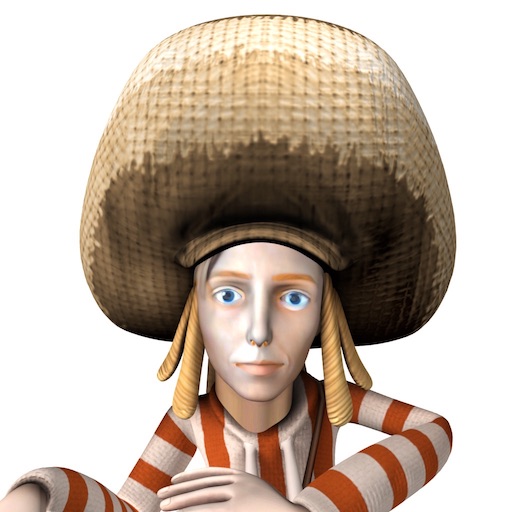} &
  \includegraphics[width=0.2\linewidth]{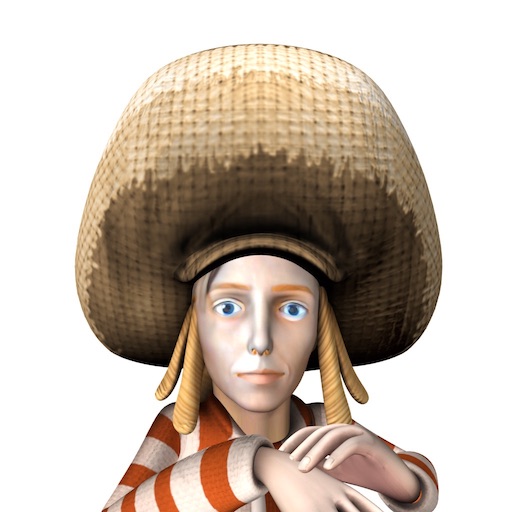}\\
  Model & Content & \multicolumn{3}{c}{Animation Results} \\
  \end{tabular}}
 \caption{\textbf{Application of personalized 3D character animation.} We replace the identity of the original models with other contents, and the new models can further be animated.}
 \label{fig:app_3d_cartoon}
\end{figure}

\begin{figure*}[t]
    \centering
    \setlength{\tabcolsep}{0.1mm}{
    \begin{tabular}{cccccccc}
      \includegraphics[width=0.1235\linewidth]{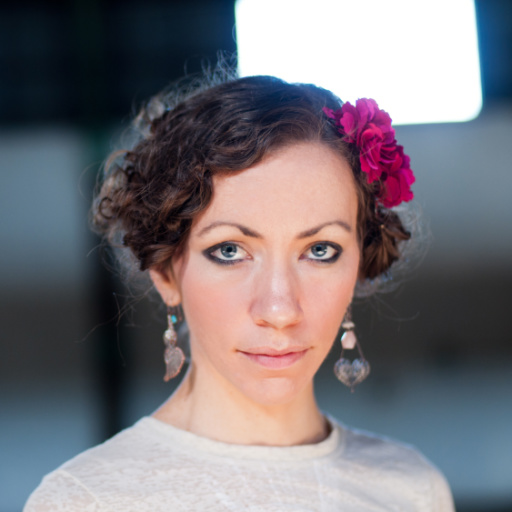} &
      \includegraphics[width=0.1235\linewidth]{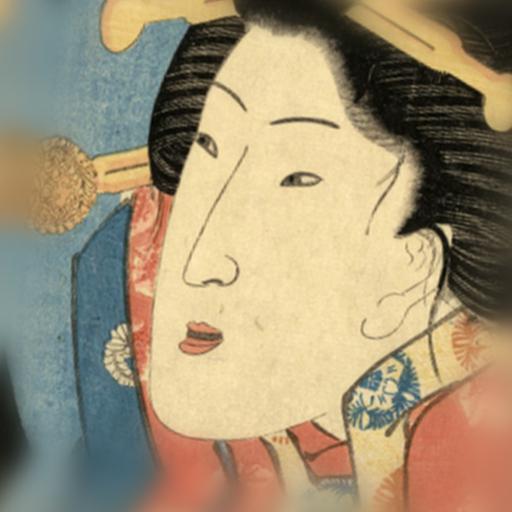} &
      \includegraphics[width=0.1235\linewidth]{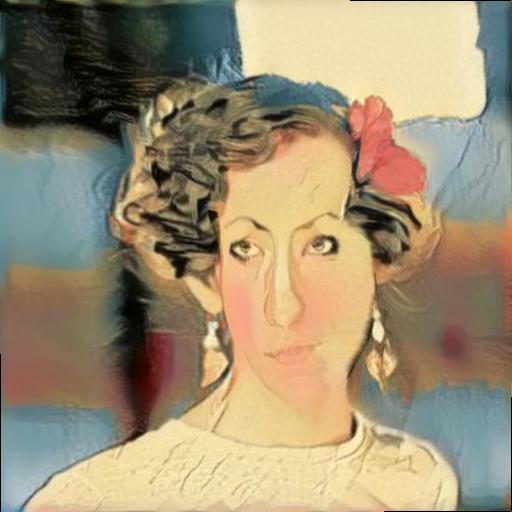} &
      \includegraphics[width=0.1235\linewidth]{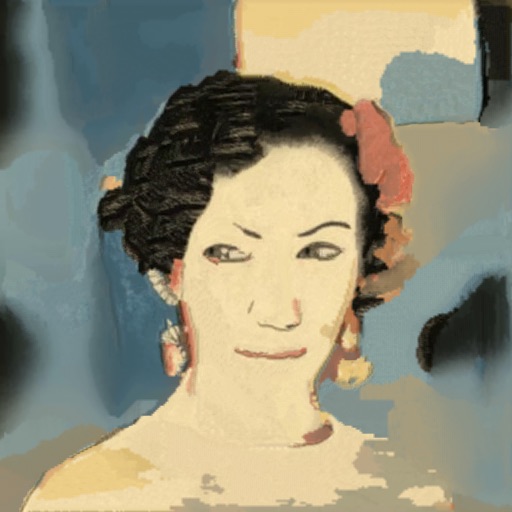} &
      \includegraphics[width=0.1235\linewidth]{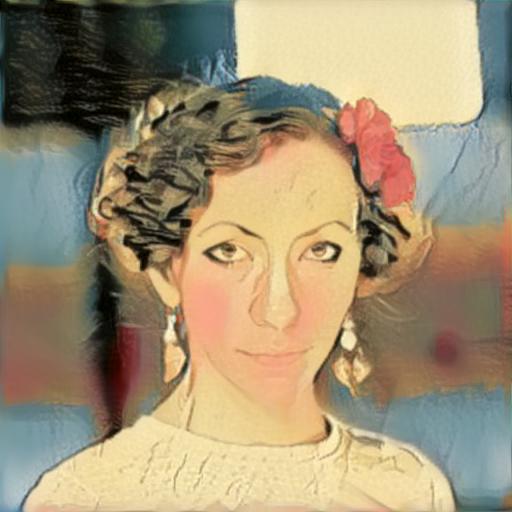} &
      \includegraphics[width=0.1235\linewidth]{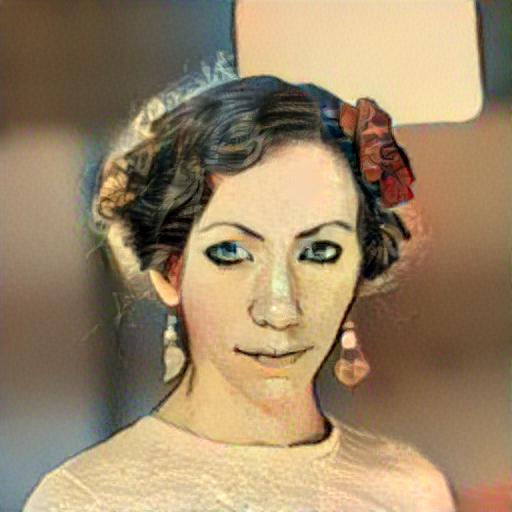} &
      \includegraphics[width=0.1235\linewidth]{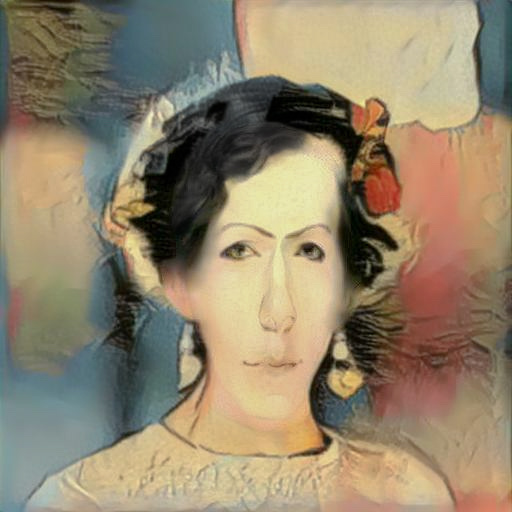} &
      \includegraphics[width=0.1235\linewidth]{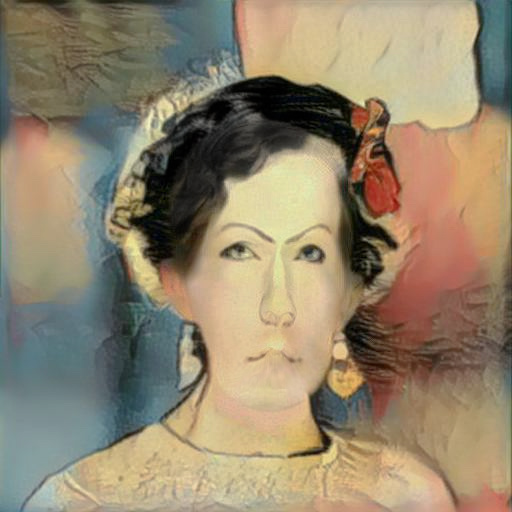} \\
      \includegraphics[width=0.1235\linewidth]{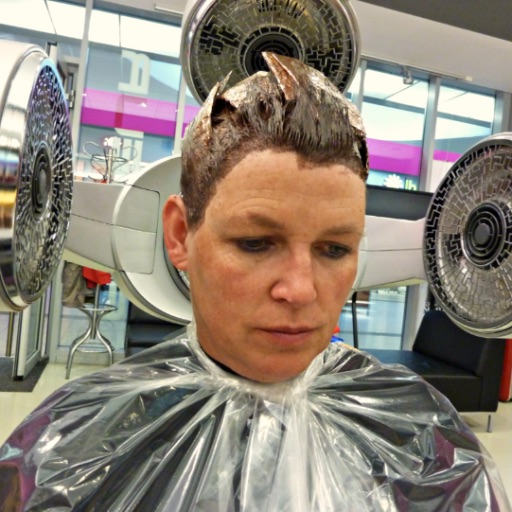} &
      \includegraphics[width=0.1235\linewidth]{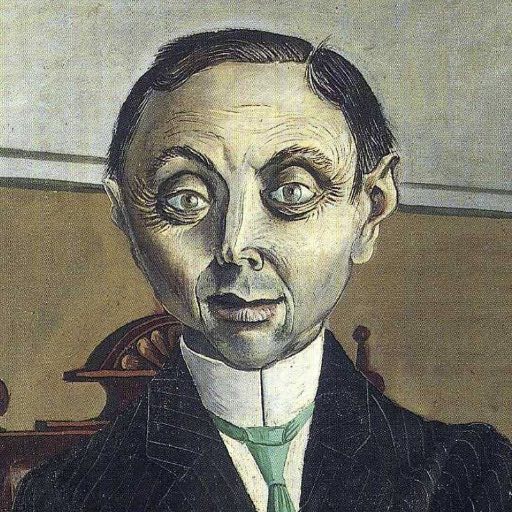} &
      \includegraphics[width=0.1235\linewidth]{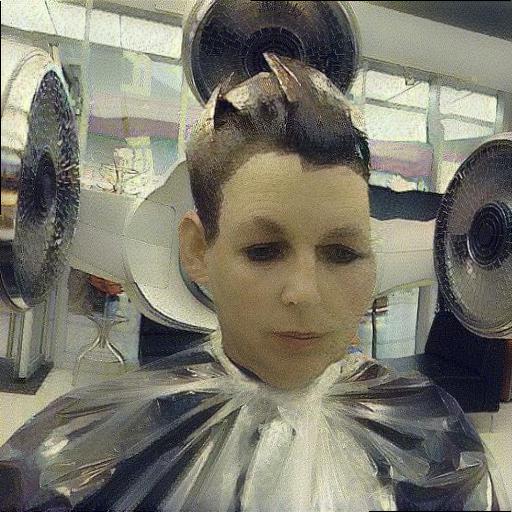} &
      \includegraphics[width=0.1235\linewidth]{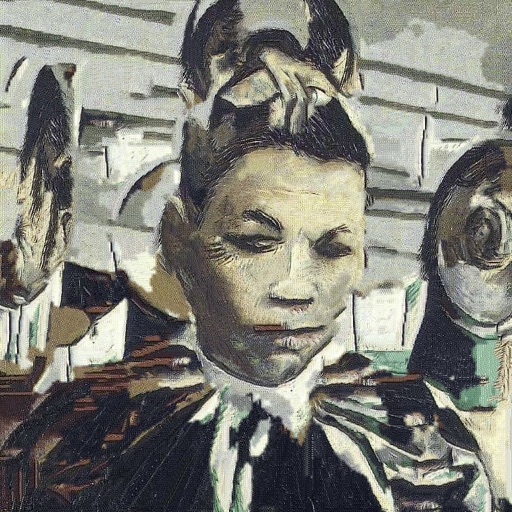} &
      \includegraphics[width=0.1235\linewidth]{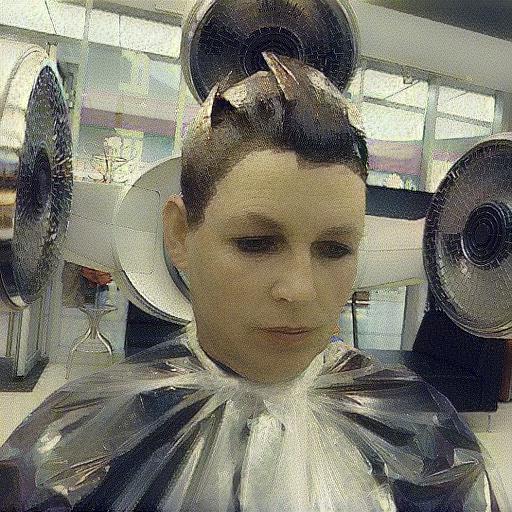} &
      \includegraphics[width=0.1235\linewidth]{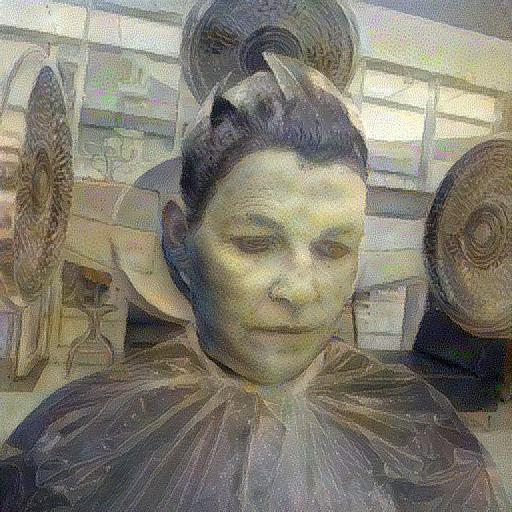} &
      \includegraphics[width=0.1235\linewidth]{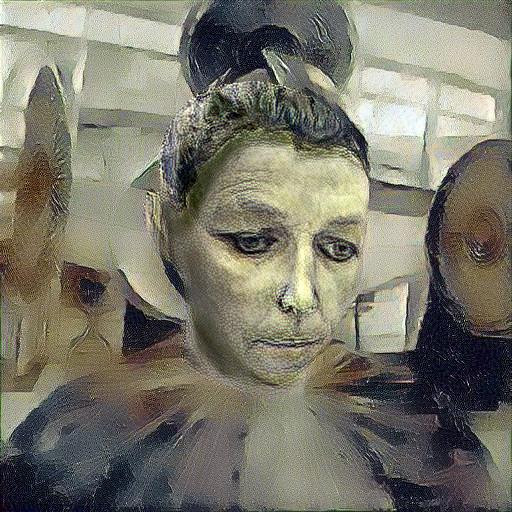} &
      \includegraphics[width=0.1235\linewidth]{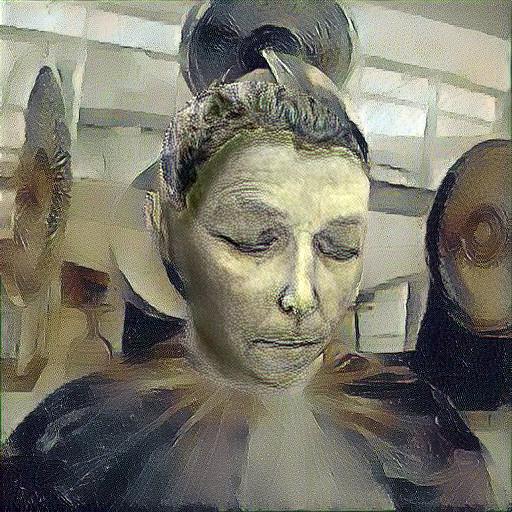} \\
      Content & Style & \small{\cite{yaniv2019face}} & \small{\cite{liao2017visual}} & \small{\cite{gatys2015neural}} & \small{\cite{kolkin2019style}} & Ours & Ours mod.\\
      \multicolumn{8}{c}{(a) Comparison on artistic style transfer of dataset from~\cite{yaniv2019face}.} \\
    \end{tabular}
    \vspace*{2mm}
    \begin{tabular}{cccccccc}
      \includegraphics[width=0.1235\linewidth]{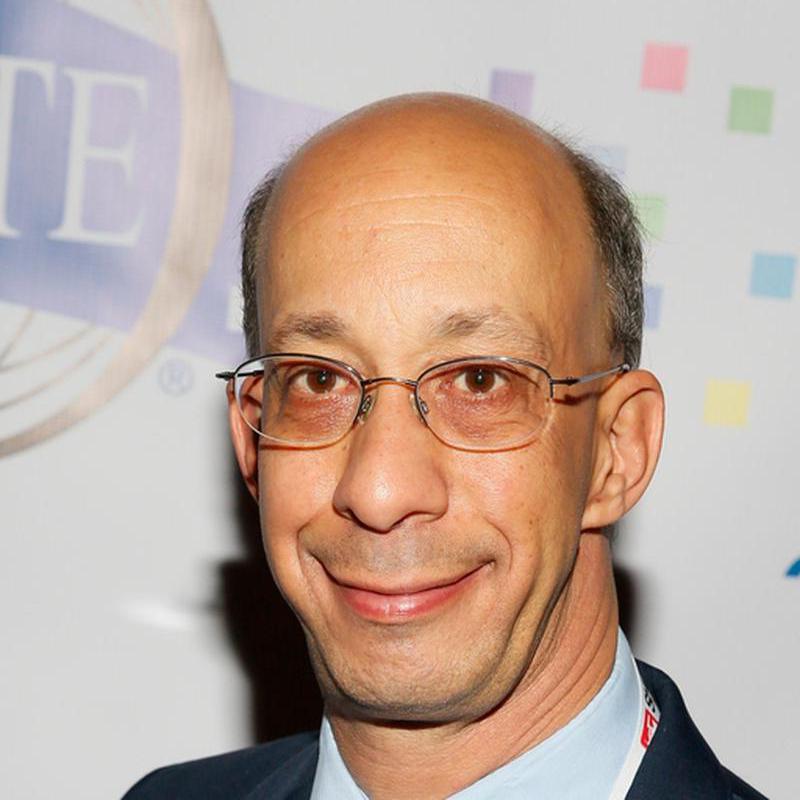} &
      \includegraphics[width=0.1235\linewidth]{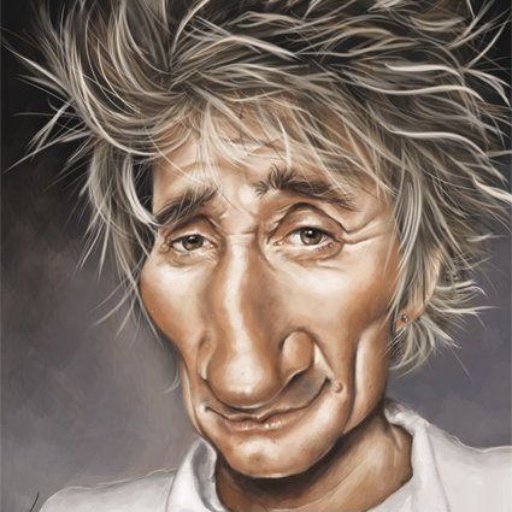} &
      \includegraphics[width=0.1235\linewidth]{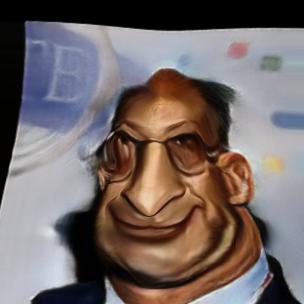} &
      \includegraphics[width=0.1235\linewidth]{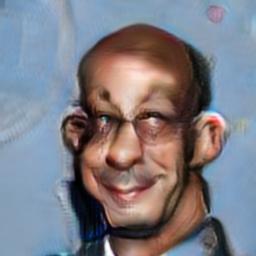} &
      \includegraphics[width=0.1235\linewidth]{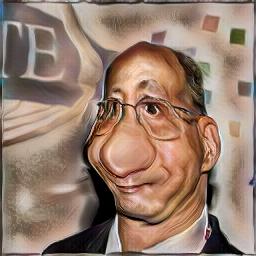} &
      \includegraphics[width=0.1235\linewidth]{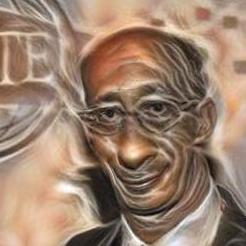} &
      \includegraphics[width=0.1235\linewidth]{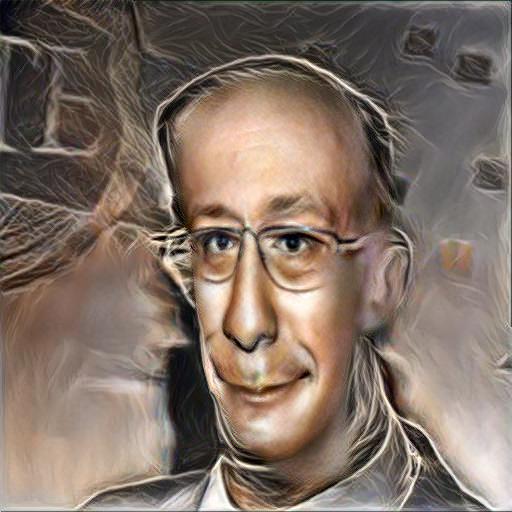} &
      \includegraphics[width=0.1235\linewidth]{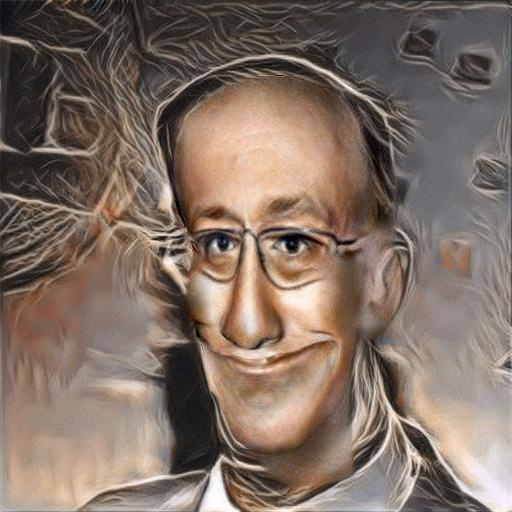}\\
      \includegraphics[width=0.1235\linewidth]{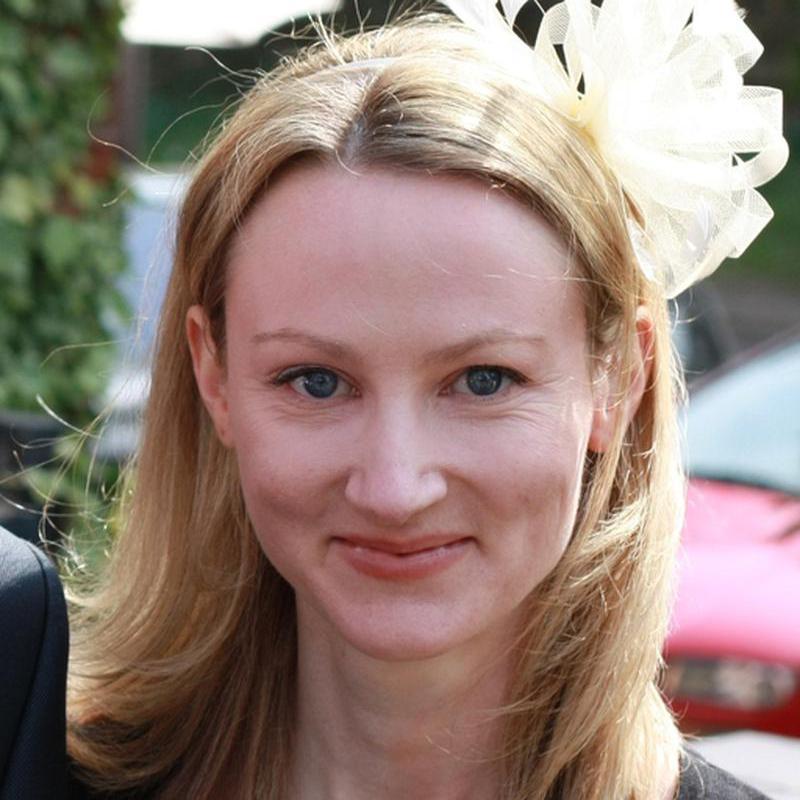} &
      \includegraphics[width=0.1235\linewidth]{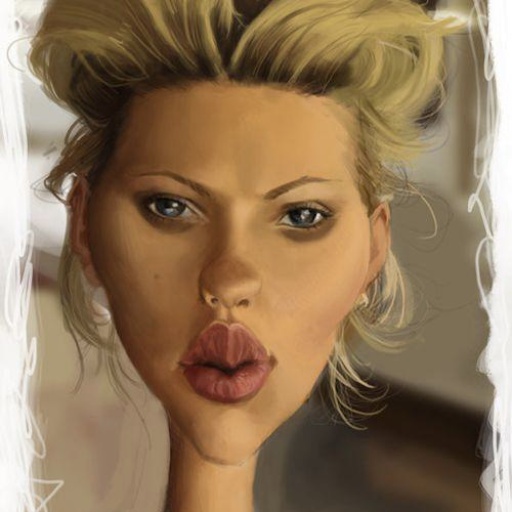} &
      \includegraphics[width=0.1235\linewidth]{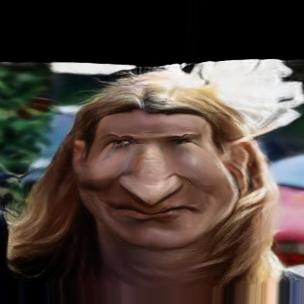} &
      \includegraphics[width=0.1235\linewidth]{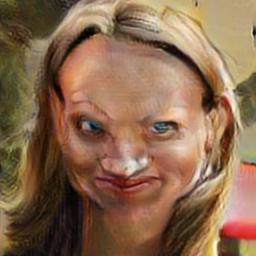} &
      \includegraphics[width=0.1235\linewidth]{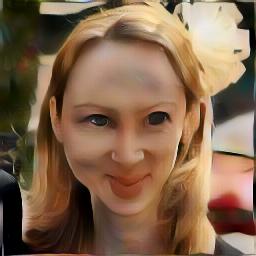} &
      \includegraphics[width=0.1235\linewidth]{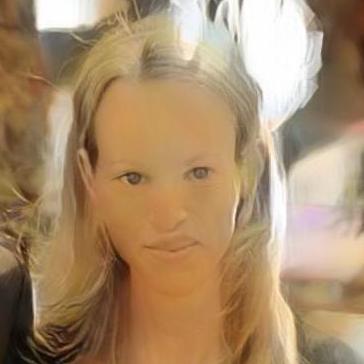} &
      \includegraphics[width=0.1235\linewidth]{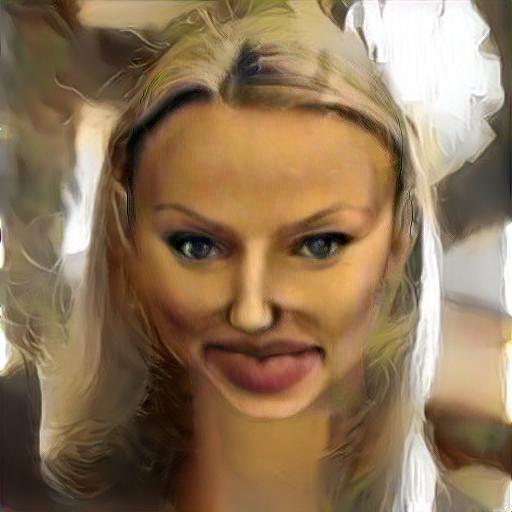} &
      \includegraphics[width=0.1235\linewidth]{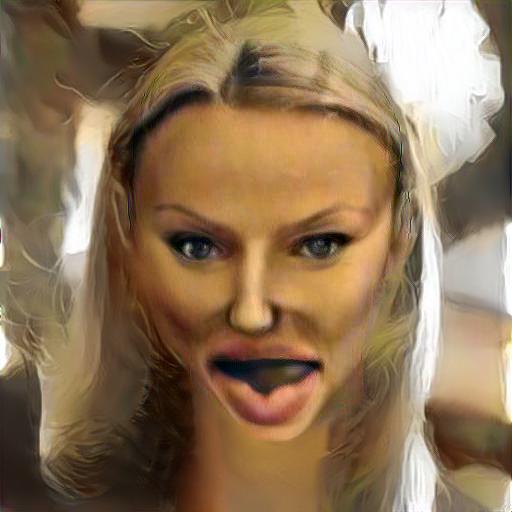}\\
      Content & Style & \small{\cite{cao2018carigans}} & \small{\cite{shi2019warpgan}} & \small{\cite{chu2020learning}} & \small{\cite{yaniv2019face}} & Ours & Ours mod.\\
      \multicolumn{8}{c}{(b) Comparison on caricature style transfer.} \\
    \end{tabular}}
 \caption{\textbf{Application of geometric-aware image style transfer.} We show our results and comparisons to the state-of-art image style transfer methods, on both artistic styles (a) and caricature styles (b). In the last column we generate image style transfer results with expression changed naturally by altering the 3DMM parameters of the models.}
 \label{fig:cmp_stoa}
\end{figure*}

\subsection{Stylized Portrait Reenactment}
3D animated avatars bring much fun to video meetings and social media. Currently, most avatars are predefined 3D models that cannot be customized. Different from them, our stylized models serving as avatars can capture users' facial characteristics and be reenacted by real-time video. Given a video sequence, we first select a neutral frame to go through our framework and obtain the deformed mesh and stylized texture. Meshes of other frames with continuously varying expressions and poses are then reconstructed by a face tracking algorithm \cite{cao2018stabilized}. We calculate the displacements between the meshes of the neutral frame and other frames. The displacements are scaled according to the local deformation and then added to the deformed mesh to reveal corresponding expressions. For texture, as meshes in all frames share identical topology and texture coordinates, we simply replace their original textures with the neutral frame's stylized texture. As shown in Fig. \ref{fig:app_video}, our method can generate and animate recognizable avatars with different artistic styles.

\subsection{Cartoon 3D Portrait Modeling}
Another application of our method is to model the complete 3D stylized portrait, including not just the face but also non-face regions containing the hair, the shoulders, and other foreground components. To achieve so, we adopt the 3D portrait depth estimation method~\cite{chai2015high} to estimate the depth of these non-face regions and merge it with the face geometry generated by our method to get the final portrait 3D shape. As for the texture, we adopt the optimization-based neural image style transfer method~\cite{kolkin2019style} to stylize the non-face regions in the image domain guided by the foreground segmentation mask and then project the stylized image back to the texture space to seamlessly blend with our face texture. As shown in Fig.~\ref{fig:app_3d_style}, by using this complete portrait model, we can produce convincing upper-body stylization results, especially for subjects with long hairstyles.

\subsection{Personalized 3D Character Animation}
In addition to 3D portrait modeling, our stylized face model can also be merged with a full-body character model for building a personalized 3D character, which can be further animated. To achieve this, a stylized portrait model is first generated with a style image being the rendered face region of the original characters, and then our stylized face model is stitched with the body of the characters to replace its original face. As shown in Fig. \ref{fig:app_3d_cartoon}, the identity of the original model are replaced by given contents. With this method, 3D character animations can be personalized. Although the results may not be as delicate as those produced manually by professional artists, our method can provide a good initialization for further refinements.

\subsection{Geometry-Aware Image Style Transfer}

Besides 3D graphics applications, our method also has 2D image applications. {For example, it can be applied to geometry-aware image style transfer, which is previously challenging for image-level methods. After obtaining a 3D stylized model, we first render it to be aligned with the warped image content guided by translated landmarks. This rendered image serves as the stylization guidance for the face region by adding an extra perceptual loss in the existing image style transfer framework, \textit{e.g.} \cite{kolkin2019style}. Joint optimization of the whole image guarantees the stylization consistency between background and face.} Fig. \ref{fig:cmp_stoa} shows example results of our proposed application. Compared to most style transfer methods that deal only with texture style transfer \cite{gatys2015neural,liao2017visual,kolkin2019style}, our geometry-combining-texture style transfer results are more visually appealing, while maintaining higher artistic credibility and presenting higher variation. Compared to other style transfer methods also consider geometry  \cite{yaniv2019face,cao2018carigans,shi2019warpgan,chu2020learning}, our methods win on the flexibility. CariGAN \cite{cao2018carigans}, WarpGAN \cite{shi2019warpgan} and Sematic-CariGANs \cite{chu2020learning} can only work on caricatures and fails to generate other results in other styles, like styles in Fig. \ref{fig:cmp_stoa} (a). FOA \cite{yaniv2019face} is more flexible than these three caricature generation methods because it can support styles from several artists in its bank, but it is still artist-specific and cannot extract geometry style from a single reference. Compared to existing geometry-aware style transfer methods, ours shows the largest flexibility, as it can not only support arbitrary styles but also easily control the result with the given reference. Moreover, by altering the expression coefficients in 3DMM reconstruction, we can modify the expression of the result models, therefore reasonably manipulate the expressions on the image transfer results. Some results are given in the last column of Fig. \ref{fig:cmp_stoa}, which has not been shown in previous image style transfer methods.

\begin{figure}[t]
\setlength{\tabcolsep}{0.2mm}{
  \begin{tabular}{cccc}
      \includegraphics[width=0.24\linewidth]{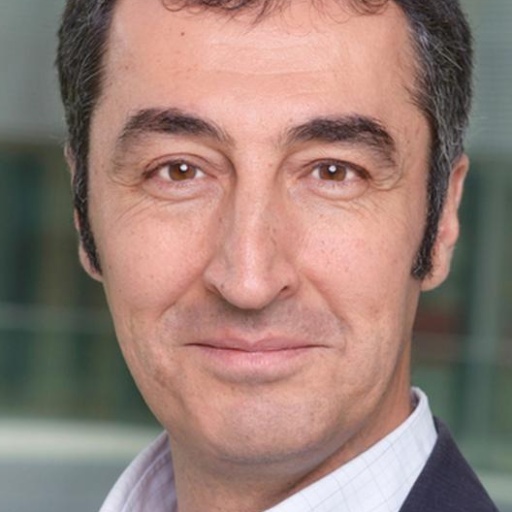} &
      \includegraphics[width=0.24\linewidth]{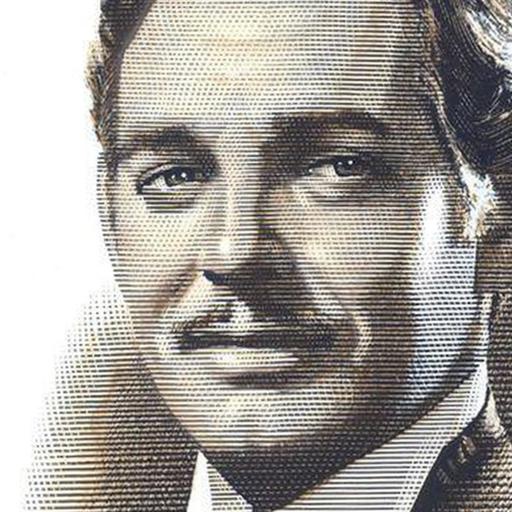} &
      \includegraphics[width=0.24\linewidth]{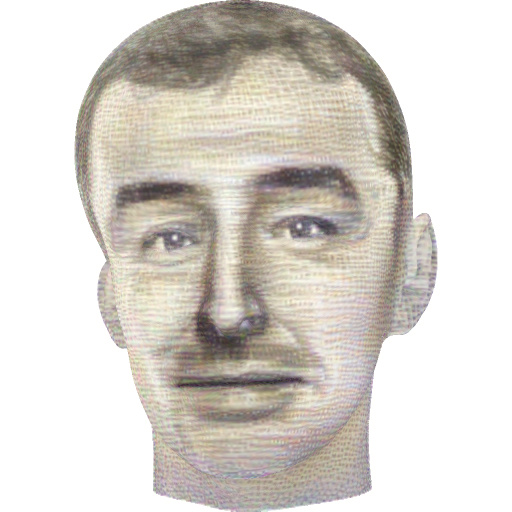} &
      \includegraphics[width=0.24\linewidth]{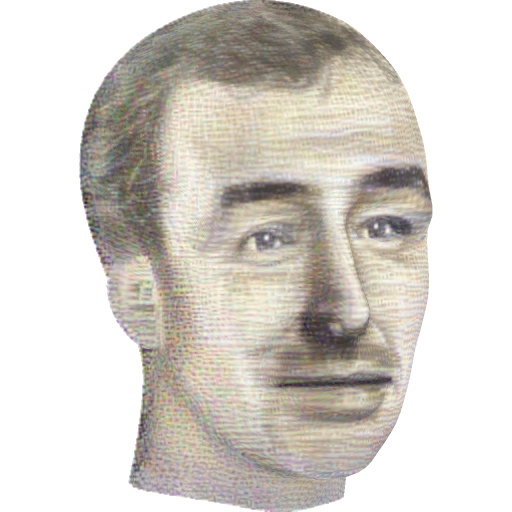}\\ 
      \includegraphics[width=0.24\linewidth]{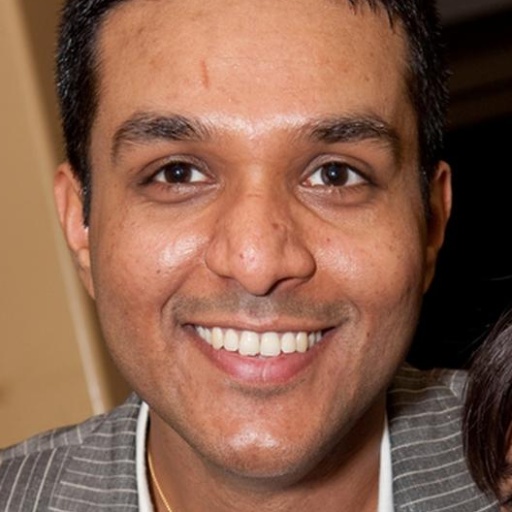} &
      \includegraphics[width=0.24\linewidth]{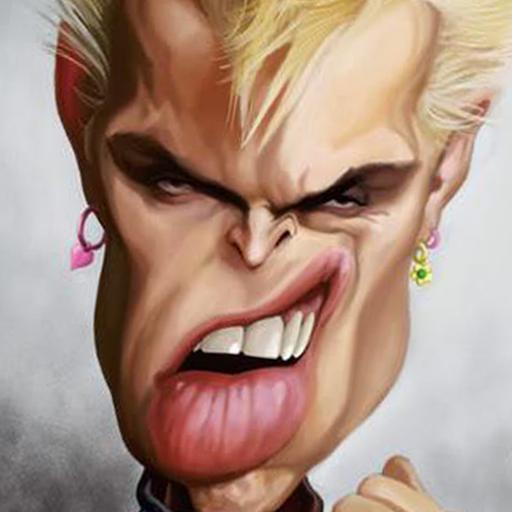} &
      \includegraphics[width=0.24\linewidth]{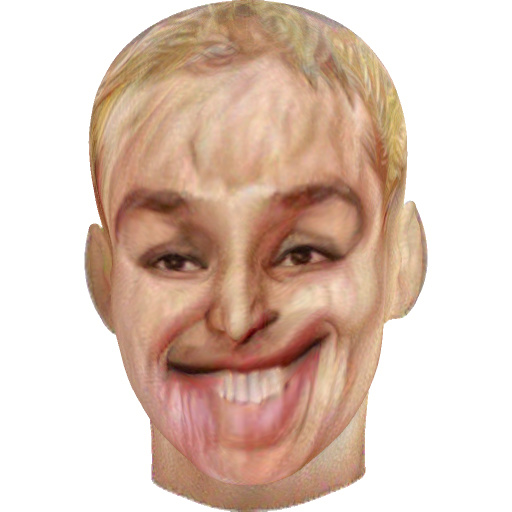} & 
      \includegraphics[width=0.24\linewidth]{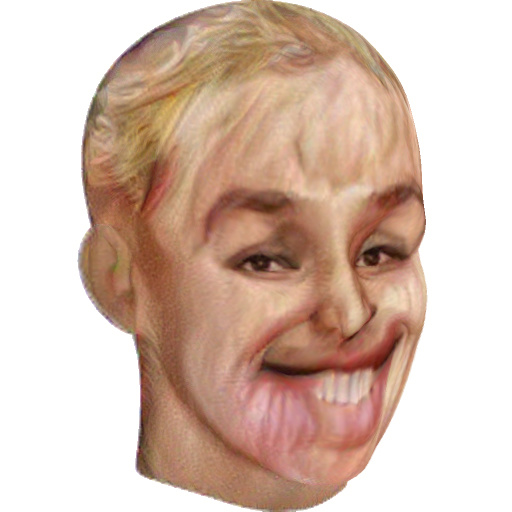}\\
      Content & Style & \multicolumn{2}{c}{Multi-View Results} \\
  \end{tabular}
  }
 \caption{Limitations of our method, including subtle deformation for styles with normal geometry and possible mismatches in the largely deformed regions. }
 \label{fig:limit}
\end{figure}

\section{Conclusion}
\label{sec:conclusion}

In this paper, we present the first automatic framework for 2D-to-3D portrait style transfer. The proposed method takes two face images as input, one portrait photo as the content, and one piece of portrait art as the style. Our method is able to generate a stylized 3D face model with both the geometry exaggerated and the texture transferred from the artistic reference while preserving the identity in the photo. Visually appealing quality and flexibility in control show the potential of our method to be applied in the areas of the art design, game development, film industry, for 3D modeling.

Our method still has limitations. First, in landmark translation, because we define the geometric style to be the shape exaggeration part beyond the normal face distribution. If the shape of a style is close to a normal face, the geometry learned from it will be subtle, as shown in the upper row of Fig.\ref{fig:limit}. Another limitation is inherited from the image style transfer. There may be mismatches between the texture and reference style, especially in the highly deformed regions, like lips in the lower row of Fig. \ref{fig:limit}. How to solve these limitations and how to extend our work to broad categories of 3D models, like the human body, would be directions worth exploring in the future.

\bibliographystyle{IEEEtran}
\bibliography{main}

\begin{thebibliography}{10}
\providecommand{\url}[1]{#1}
\csname url@samestyle\endcsname
\providecommand{\newblock}{\relax}
\providecommand{\bibinfo}[2]{#2}
\providecommand{\BIBentrySTDinterwordspacing}{\spaceskip=0pt\relax}
\providecommand{\BIBentryALTinterwordstretchfactor}{4}
\providecommand{\BIBentryALTinterwordspacing}{\spaceskip=\fontdimen2\font plus
\BIBentryALTinterwordstretchfactor\fontdimen3\font minus
  \fontdimen4\font\relax}
\providecommand{\BIBforeignlanguage}[2]{{%
\expandafter\ifx\csname l@#1\endcsname\relax
\typeout{** WARNING: IEEEtran.bst: No hyphenation pattern has been}%
\typeout{** loaded for the language `#1'. Using the pattern for}%
\typeout{** the default language instead.}%
\else
\language=\csname l@#1\endcsname
\fi
#2}}
\providecommand{\BIBdecl}{\relax}
\BIBdecl

\bibitem{gatys2015neural}
L.~A. Gatys, A.~S. Ecker, and M.~Bethge, ``A neural algorithm of artistic
  style,'' \emph{CoRR}, vol. abs/1508.06576, 2015.

\bibitem{liao2017visual}
J.~Liao, Y.~Yao, L.~Yuan, G.~Hua, and S.~B. Kang, ``Visual attribute transfer
  through deep image analogy,'' \emph{{ACM} Trans. Graph.}, vol.~36, no.~4, pp.
  120:1--120:15, 2017.

\bibitem{kolkin2019style}
N.~I. Kolkin, J.~Salavon, and G.~Shakhnarovich, ``Style transfer by relaxed
  optimal transport and self-similarity,'' in \emph{CVPR 2019}, 2019, pp.
  10\,051--10\,060.

\bibitem{yaniv2019face}
J.~Yaniv, Y.~Newman, and A.~Shamir, ``The face of art: landmark detection and
  geometric style in portraits,'' \emph{{ACM} Trans. Graph.}, vol.~38, no.~4,
  pp. 60:1--60:15, 2019.

\bibitem{cao2018carigans}
K.~Cao, J.~Liao, and L.~Yuan, ``Carigans: unpaired photo-to-caricature
  translation,'' \emph{{ACM} Trans. Graph.}, vol.~37, no.~6, pp. 244:1--244:14,
  2018.

\bibitem{shi2019warpgan}
Y.~Shi, D.~Deb, and A.~K. Jain, ``Warpgan: Automatic caricature generation,''
  in \emph{CVPR 2019}, 2019, pp. 10\,762--10\,771.

\bibitem{hertzmann2001image}
A.~Hertzmann, C.~E. Jacobs, N.~Oliver, B.~Curless, and D.~Salesin, ``Image
  analogies,'' in \emph{SIGGRAPH 2001}, 2001, pp. 327--340.

\bibitem{efros2001image}
A.~A. Efros and W.~T. Freeman, ``Image quilting for texture synthesis and
  transfer,'' in \emph{SIGGRAPH 2001}, 2001, pp. 341--346.

\bibitem{gatys2016image}
L.~A. Gatys, A.~S. Ecker, and M.~Bethge, ``Image style transfer using
  convolutional neural networks,'' in \emph{CVPR 2016}, 2016, pp. 2414--2423.

\bibitem{gu2018arbitrary}
S.~Gu, C.~Chen, J.~Liao, and L.~Yuan, ``Arbitrary style transfer with deep
  feature reshuffle,'' in \emph{CVPR 2018}, 2018, pp. 8222--8231.

\bibitem{johnson2016perceptual}
J.~Johnson, A.~Alahi, and L.~Fei{-}Fei, ``Perceptual losses for real-time style
  transfer and super-resolution,'' in \emph{ECCV 2016}, vol. 9906, 2016, pp.
  694--711.

\bibitem{ulyanov2016texture}
D.~Ulyanov, V.~Lebedev, A.~Vedaldi, and V.~S. Lempitsky, ``Texture networks:
  Feed-forward synthesis of textures and stylized images.'' in \emph{ICML},
  vol.~1, no.~2, 2016, p.~4.

\bibitem{chen2017stylebank}
D.~Chen, L.~Yuan, J.~Liao, N.~Yu, and G.~Hua, ``Stylebank: An explicit
  representation for neural image style transfer,'' in \emph{Proceedings of the
  IEEE conference on computer vision and pattern recognition}, 2017, pp.
  1897--1906.

\bibitem{huang2017arbitrary}
X.~Huang and S.~J. Belongie, ``Arbitrary style transfer in real-time with
  adaptive instance normalization,'' in \emph{ICCV 2017}, 2017, pp. 1510--1519.

\bibitem{li2017universal}
Y.~Li, C.~Fang, J.~Yang, Z.~Wang, X.~Lu, and M.-H. Yang, ``Universal style
  transfer via feature transforms,'' in \emph{Advances in neural information
  processing systems}, 2017, pp. 386--396.

\bibitem{sheng2018avatar}
L.~Sheng, Z.~Lin, J.~Shao, and X.~Wang, ``Avatar-net: Multi-scale zero-shot
  style transfer by feature decoration,'' in \emph{Proceedings of the IEEE
  Conference on Computer Vision and Pattern Recognition}, 2018, pp. 8242--8250.

\bibitem{zhu2017unpaired}
J.~Zhu, T.~Park, P.~Isola, and A.~A. Efros, ``Unpaired image-to-image
  translation using cycle-consistent adversarial networks,'' in \emph{ICCV
  2017}, 2017, pp. 2242--2251.

\bibitem{liu2017unsupervised}
M.~Liu, T.~Breuel, and J.~Kautz, ``Unsupervised image-to-image translation
  networks,'' in \emph{NeurIPS 2017}, 2017, pp. 700--708.

\bibitem{huang2018multimodal}
X.~Huang, M.~Liu, S.~J. Belongie, and J.~Kautz, ``Multimodal unsupervised
  image-to-image translation,'' in \emph{ECCV 2018}, vol. 11207, 2018, pp.
  179--196.

\bibitem{lee2019drit}
H.~Lee, H.~Tseng, Q.~Mao, J.~Huang, Y.~Lu, M.~Singh, and M.~Yang, ``{DRIT++:}
  diverse image-to-image translation via disentangled representations,''
  \emph{CoRR}, vol. abs/1905.01270, 2019.

\bibitem{selim2016painting}
A.~Selim, M.~A. Elgharib, and L.~Doyle, ``Painting style transfer for head
  portraits using convolutional neural networks,'' \emph{{ACM} Trans. Graph.},
  vol.~35, no.~4, pp. 129:1--129:18, 2016.

\bibitem{fiser2017example}
J.~Fiser, O.~Jamriska, D.~Simons, E.~Shechtman, J.~Lu, P.~Asente,
  M.~Luk{\'{a}}c, and D.~S{\'{y}}kora, ``Example-based synthesis of stylized
  facial animations,'' \emph{{ACM} Trans. Graph.}, vol.~36, no.~4, pp.
  155:1--155:11, 2017.

\bibitem{futschik2019real}
D.~Futschik, M.~Chai, C.~Cao, C.~Ma, A.~Stoliar, S.~Korolev, S.~Tulyakov,
  M.~Kucera, and D.~S{\'{y}}kora, ``Real-time patch-based stylization of
  portraits using generative adversarial network,'' in \emph{Expressive 2019},
  2019, pp. 33--42.

\bibitem{brennan1985dynamic}
S.~E. Brennan, ``The dynamic exaggeration of faces by computer,''
  \emph{Leonardo}, vol.~18, no.~3, pp. 170--178, 1985.

\bibitem{akleman1997making}
E.~Akleman, ``Making caricatures with morphing,'' in \emph{SIGGRAPH 1997},
  1997, p. 145.

\bibitem{liang2002example}
L.~Liang, H.~Chen, Y.~Xu, and H.~Shum, ``Example-based caricature generation
  with exaggeration,'' in \emph{PG 2002}, 2002, pp. 386--393.

\bibitem{li2018carigan}
W.~Li, W.~Xiong, H.~Liao, J.~Huo, Y.~Gao, and J.~Luo, ``Carigan: Caricature
  generation through weakly paired adversarial learning,'' \emph{CoRR}, vol.
  abs/1811.00445, 2018.

\bibitem{kato2018neural}
H.~Kato, Y.~Ushiku, and T.~Harada, ``Neural 3d mesh renderer,'' in \emph{CVPR
  2018}, 2018, pp. 3907--3916.

\bibitem{liu2018paparazzi}
H.~D. Liu, M.~Tao, and A.~Jacobson, ``Paparazzi: surface editing by way of
  multi-view image processing,'' \emph{{ACM} Trans. Graph.}, vol.~37, no.~6,
  pp. 221:1--221:11, 2018.

\bibitem{mordvintsev2018differentiable}
A.~Mordvintsev, N.~Pezzotti, L.~Schubert, and C.~Olah, ``Differentiable image
  parameterizations,'' \emph{Distill}, vol.~3, no.~7, p. e12, 2018.

\bibitem{lewiner2011interactive}
T.~Lewiner, T.~Vieira, D.~M. Morera, A.~Peixoto, V.~Mello, and L.~Velho,
  ``Interactive 3d caricature from harmonic exaggeration,'' \emph{Comput.
  Graph.}, vol.~35, no.~3, pp. 586--595, 2011.

\bibitem{vieira2013three}
R.~C.~C. Vieira, C.~A. Vidal, and J.~B.~C. Neto, ``Three-dimensional face
  caricaturing by anthropometric distortions,'' in \emph{SIBGRAPI 2013}, 2013,
  pp. 163--170.

\bibitem{liu2009semi}
J.~Liu, Y.~Chen, C.~Miao, J.~Xie, C.~X. Ling, X.~Gao, and W.~Gao,
  ``Semi-supervised learning in reconstructed manifold space for 3d caricature
  generation,'' \emph{Comput. Graph. Forum}, vol.~28, no.~8, pp. 2104--2116,
  2009.

\bibitem{wu2018alive}
Q.~Wu, J.~Zhang, Y.~Lai, J.~Zheng, and J.~Cai, ``Alive caricature from 2d to
  3d,'' in \emph{CVPR 2018}, 2018, pp. 7336--7345.

\bibitem{han2017deepsketch}
X.~Han, C.~Gao, and Y.~Yu, ``Deepsketch2face: a deep learning based sketching
  system for 3d face and caricature modeling,'' \emph{{ACM} Trans. Graph.},
  vol.~36, no.~4, pp. 126:1--126:12, 2017.

\bibitem{han2018caricatureshop}
X.~Han, K.~Hou, D.~Du, Y.~Qiu, Y.~Yu, K.~Zhou, and S.~Cui, ``Caricatureshop:
  Personalized and photorealistic caricature sketching,'' \emph{CoRR}, vol.
  abs/1807.09064, 2018.

\bibitem{ye2020carigan}
Z.~Ye, R.~Yi, M.~Yu, J.~Zhang, Y.~Lai, and Y.~Liu, ``3d-carigan: An end-to-end
  solution to 3d caricature generation from face photos,'' \emph{CoRR}, vol.
  abs/2003.06841, 2020.

\bibitem{dlib}
\BIBentryALTinterwordspacing
DLib. (2019) Dlib c++ library. [Online]. Available: \url{http://dlib.net/}
\BIBentrySTDinterwordspacing

\bibitem{cao2014facewarehouse}
C.~Cao, Y.~Weng, S.~Zhou, Y.~Tong, and K.~Zhou, ``Facewarehouse: {A} 3d facial
  expression database for visual computing,'' \emph{{IEEE} Trans. Vis. Comput.
  Graph.}, vol.~20, no.~3, pp. 413--425, 2014.

\bibitem{li2016combining}
C.~Li and M.~Wand, ``Combining markov random fields and convolutional neural
  networks for image synthesis,'' in \emph{Proceedings of the IEEE Conference
  on Computer Vision and Pattern Recognition}, 2016, pp. 2479--2486.

\bibitem{chai2015high}
M.~Chai, L.~Luo, K.~Sunkavalli, N.~Carr, S.~Hadap, and K.~Zhou, ``High-quality
  hair modeling from a single portrait photo,'' \emph{{ACM} Trans. Graph.},
  vol.~34, no.~6, pp. 204:1--204:10, 2015.

\bibitem{chu2020learning}
W.~Chu, W.-C. Hung, Y.-H. Tsai, Y.-T. Chang, Y.~Li, D.~Cai, and M.-H. Yang,
  ``Learning to caricature via semantic shape transform,'' \emph{arXiv preprint
  arXiv:2008.05090}, 2020.

\bibitem{cao2018stabilized}
C.~Cao, M.~Chai, O.~Woodford, and L.~Luo, ``Stabilized real-time face tracking
  via a learned dynamic rigidity prior,'' \emph{ACM Transactions on Graphics
  (TOG)}, vol.~37, no.~6, pp. 1--11, 2018.

\end{thebibliography}

\end{document}